%% file: ijcai26.tex
\documentclass{article}
\usepackage{ijcai26}
\usepackage{times}
\usepackage{soul}
\usepackage{url}
\usepackage[hidelinks]{hyperref}
\usepackage[utf8]{inputenc}
\usepackage[small]{caption}
\usepackage{graphicx}
\usepackage{amsmath}
\usepackage{amssymb}
\usepackage{amsthm}
\usepackage{booktabs}
\usepackage{algorithm}
\usepackage{algorithmic}
\usepackage{multirow, multicol}
\usepackage{listings}
\usepackage{amssymb}
\usepackage{minitoc}
\usepackage{pdflscape}
\usepackage{comment}
\usepackage{xcolor}
\usepackage{bm}
\usepackage{tabularx}
\usepackage{array}
\usepackage{makecell}
\usepackage{rotating}
\usepackage{tikz}
\usetikzlibrary{matrix, positioning, arrows.meta}
\usepackage{bibunits}
\usepackage{etoc}

\newif\ifanonymized
\anonymizedfalse  

\ifanonymized
    \usepackage[switch]{lineno}
    \linenumbers
\else
\fi

\ifanonymized
    \linenumbers
\else
\fi

\title{Out-of-Distribution Detection via Channelwise Feature Aggregation \\
in Neural Network--Based Receivers}

\ifanonymized
    \author{
        Author Name
        \affiliations
        Affiliation
        \emails
        email@example.com
    }
\else
    \author{
    Marko Tuononen$^{1,3}$\and
    Heikki Penttinen$^1$\and
    Duy Vu$^1$\and
    Dani Korpi$^2$\and \\
    Vesa Starck$^1$\And
    Ville Hautamäki$^3$\\
    \affiliations
    $^1$Nokia Networks, Karakaari 7, 02610 Espoo, Finland\\
    $^2$Nokia Bell Labs, Karakaari 7, 02610 Espoo, Finland\\
    $^3$University of Eastern Finland, P.O. Box 111, 80101 Joensuu, Finland\\
    \emails
    Corresponding author: marko.1.tuononen@nokia.com}
\fi

\begin{document}
\begin{bibunit}[named]

\maketitle

\begin{abstract}
Neural network--based radio receivers are expected to play a key role in future wireless systems, making reliable Out-Of-Distribution (OOD) detection essential. We propose a post-hoc, layerwise OOD framework based on channelwise feature aggregation that avoids classwise statistics--critical for multi-label soft-bit outputs with astronomically many classes. Receiver activations exhibit no discrete clusters but a smooth Signal-to-Noise-Ratio (SNR)--aligned manifold, consistent with classical receiver behavior and motivating manifold-aware OOD detection. We evaluate multiple OOD feature types, distance metrics, and methods across layers. Gaussian Mahalanobis with mean activations is the strongest single detector, earlier layers outperform later, and SNR/classifier fusions offer small, inconsistent AUROC gains. High-delay OOD is detected reliably, while high-speed remains challenging.
\end{abstract}

\section{Introduction}
Deep neural network--based radio receivers are emerging as a key component of future 6G systems~\cite{hoydis2021toward}, offering improved performance and reduced design complexity compared to conventional signal-processing pipelines~\cite{farhadi2025}. However, \emph{neural receivers can be brittle under distribution shift}: when propagation conditions at inference deviate from those seen during training, performance may degrade sharply without any internal indication of failure~\cite{luostari2025otavalidation}. As emphasized in recent work on AI safety~\cite{bengio2024extremerisks}, monitoring and understanding the internal behavior of neural networks is essential for reliable deployment. In wireless systems, which operate under highly variable and unpredictable channel conditions, this requirement is fundamental.

\begin{figure}[t!]
\centering
\includegraphics[width=0.98\columnwidth]{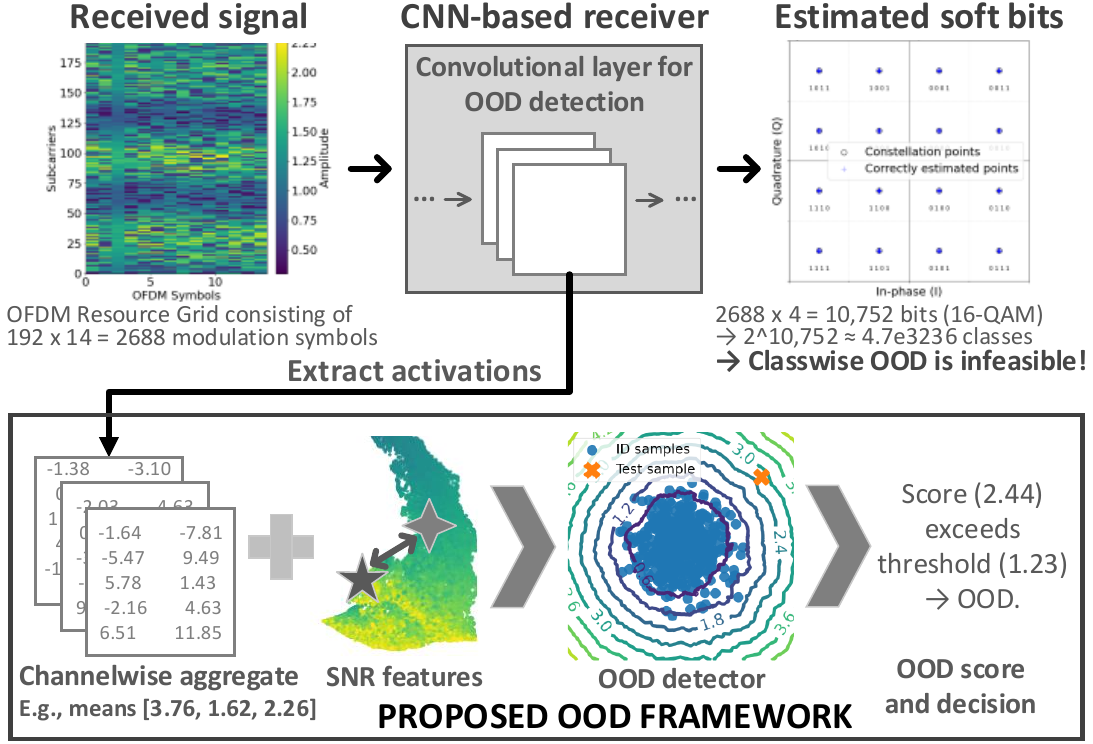}
\caption{Visual overview of OOD detection for neural receivers. OOD detection operates on aggregated activation features, rather than classwise statistics, forming an SNR-aligned manifold.}
\label{fig:high_level}
\end{figure}

Out-Of-Distribution (OOD) detection is therefore becoming indispensable in wireless machine learning~\cite{liu2022oodwireless}. We focus on Type-2 OOD (covariate shift), where the input distribution changes, while the prediction task remains unchanged, as in deployment scenarios with larger delay spreads, higher user velocities, or unforeseen interference conditions. Neural receivers such as DeepRx~\cite{honkala2021deeprx} produce thousands of soft-bit outputs per sample, forming a \emph{multi-label prediction problem}~\cite{tarekegn2024mll}--implying an astronomical output space (Fig.~\ref{fig:high_level}, top). This makes classwise OOD detection methods fundamentally ill-posed, as they rely on a limited number of discrete classes.

Practical deployment further constrains OOD detection: the receiver architecture is typically frozen and tightly optimized for latency, power consumption, and hardware compatibility~\cite{tuononen2025interpreting}. Many powerful OOD approaches--such as Bayesian neural networks~\cite{blundell2015}, deep ensembles~\cite{lakshminarayanan2017}, evidential networks \cite{aguilar2025evidential}, or auxiliary OOD heads~\cite{hendrycks2019}--require architectural modification, retraining, or additional inference passes, making them unsuitable for production-grade wireless receivers. This motivates post-hoc OOD detection methods that operate solely on internal activations of a frozen model.

In this paper, we propose a post-hoc OOD detection framework for neural network--based radio receivers (Fig.~\ref{fig:high_level}, bottom) that avoids classwise modeling and leaves the receiver architecture unchanged. We show that receiver activations do not form discrete clusters but instead lie on a continuous Signal-to-Noise Ratio (SNR)--aligned manifold, a physically plausible structure that motivates manifold-aware OOD detection. Building on this insight, we develop a layerwise, channelwise feature aggregation framework that supports multiple OOD models and metrics, incorporates SNR-aware conditioning, and enables both intra- and inter-layer scoring. The approach enables both offline debugging and online monitoring, including anomaly detection~\cite{yang2024oodsurvey}, drift awareness~\cite{uzlaner2025}, and network-assisted reporting~\cite{ren2021ood}, contributing toward reliable and trustworthy neural receivers for future 6G systems.

\section{Related Work}
According to the manifold hypothesis, high-dimensional data governed by continuous generative factors tends to concentrate near low-dimensional manifolds embedded in the input space~\cite{nilsson2007regressionmanifolds,bengio2013representation}. Such continuous structures have been observed in domains ranging from images~\cite{tenenbaum2000} to biomedical data~\cite{islam2023}, whereas discrete cluster structures are more typical of single-label classification tasks~\cite{bengio2013representation}.

Understanding internal neural representations is the goal of \emph{mechanistic interpretability}, which studies how networks encode and manipulate information~\cite{rauker2023survey,sharkey2025openproblems}. Prior work spans visualization~\cite{zeiler2014visualizing}, concept-based probes~\cite{kim2018}, representation-similarity~\cite{kornblith2019cka}, and internal circuits or neuron communities~\cite{wang2023circuit,lan2025inner}. In wireless receivers, DeepRx has been analyzed through representation-level encoding of channel parameters such as SNR~\cite{tuononen2025interpreting}, underscoring the practical value of interpretability for deployed systems~\cite{jeffares2025position}.

OOD detection aims to identify inputs that deviate from the training distribution due to covariate or semantic shift~\cite{yang2024oodsurvey}. Many post-hoc methods operate on hidden representations, including $k$-Nearest Neighbors~\cite{cover1967,papernot2018deepknn}, class-conditional Mahalanobis~\cite{lee2018mahalanobis,muller2025mahalanobis}, and channel-selection for semantic shifts~\cite{yuan2024discriminability}. Extensions to multi-label settings include energy-based aggregation~\cite{wang2021energy}, which assumes semantic shift and conditional label independence~\cite{zhang2023theoretical}, and latent-variable models such as $\beta$-Variational Auto Encoder ($\beta$-VAE) for discrete generative factors~\cite{sundar2020betavae}. OOD score fusion methods include conformal p-value aggregation~\cite{magesh2023conformalpvalues} and one-class classifiers such as One-Class Support Vector Machine (OC-SVM~\cite{scholkopf2001ocsvm,bounsiar2014ocsvm}). 

In wireless communications, OOD detection has mainly focused on transmitter identification and modulation recognition~\cite{liu2022oodwireless}. Related anomaly-detection work includes Deep Convolutional Autoencoder (DCAE)--based detection of physical tampering at Orthogonal Frequency-Division Multiplexing (OFDM) receivers~\cite{dehmollaian2021dcae}, where natural channel variation is excluded. Despite this extensive literature, post-hoc OOD detection for multi-label, Type-2 OOD neural physical-layer receivers--where classwise modeling is infeasible and the architecture must remain frozen--has not been previously addressed.

\section{Methodology}
\label{sec:methodology}
Let $X \subset \mathbb{C}^{N_f \times N_t}$ denote the space of OFDM resource-grid~\cite{3gpp38211} inputs consisting of $N_f$ subcarriers and $N_t$ OFDM symbols (e.g., $192 \times 14$). For a given modulation order $M$, the receiver outputs $Y \subset [0,1]^{N_f \times N_t \times B}$, where $B = \log_2 M$ is the number of bits per symbol; for example, $B=4$ for 16-Quadrature Amplitude Modulation (16-QAM).
We consider a frozen neural receiver $F_\theta$, parameterized by $\theta$, as a mapping $F_\theta : X \to Y$. During receiver training, data came from the joint distribution $P_{\text{train}}(X,Y)$, while deployment samples follow $P_{\text{test}}(X,Y)$. In Type-2 OOD detection (covariate shift~\cite{quinonero2008datasetshift}),
\begin{equation}
    P_{\text{train}}(Y \mid X) = P_{\text{test}}(Y \mid X) 
    \text{, but}\,
    P_{\text{train}}(X) \neq P_{\text{test}}(X),
\end{equation}
as occurs when propagation characteristics--SNR, Doppler spread, multipath profile, interference--differ between training and deployment. Our task is to construct a detector $D$ that maps each received signal to a real-valued OOD score.

\begin{figure}
    \centering
    \resizebox{\columnwidth}{!}{\input{figure_block_diagram.tikz}}
    \caption{Post‑hoc OOD detection operating on a frozen model.}
    \label{fig:block_diagram}
\end{figure}

\textbf{Layerwise Activation Extraction.} Let the neural receiver contain $L$ convolutional layers. For a chosen layer  $\ell \in \{1,\dots,L\}$, the activation tensor is
\begin{equation}
    A_\ell(x) \in \mathbb{R}^{C_\ell \times H_\ell \times W_\ell},
\end{equation}
where $C_\ell$ is the number of convolutional channels and ($H_\ell$,$W_\ell$) are the spatial dimensions (i.e., $(N_f,N_t)$ for DeepRx).
For a dataset $X = \{x_i\}_{i=1}^N$, we denote activations
\begin{equation}
    A_\ell(X) = \bigl\{A_\ell(x_i)\bigr\}_{i=1}^N.
\end{equation}

\textbf{Channelwise Feature Extraction.} Features used for OOD detection are derived from per-channel activation vectors:
\begin{equation}
\mathbf{a}_{\ell,i,c} = \mathrm{vec}\bigl(A_\ell(x_i)[c,:,:]\bigr) \in \mathbb{R}^{H_\ell W_\ell}, \, c=1,\dots,C_\ell.
\end{equation}
Each feature family applies a normalization scheme appropriate to the statistic being computed, and all resulting per-channel features are concatenated across channels. Thus, for each layer~$\ell$ and input sample~$x_i$, the activation-derived feature vector used for downstream modeling is one of
\begin{equation}
    \mathbf{g}_{\ell,i}
    \in
    \bigl\{
        \mathbf{m}_{\ell,i},\;
        \mathbf{k}_{\ell,i},\;
        \mathbf{e}_{\ell,i}
    \bigr\}.
\end{equation}

\paragraph{(i) Mean-activation features.} The MaxAbs scaled mean activation for channel $c$ is computed as ~\cite{bauerle2022nap}
\begin{equation}
    m_{\ell,i,c}
    =
    \frac{1}{H_\ell W_\ell}
    \sum_{q=1}^{H_\ell W_\ell}
        \frac{\mathbf{a}_{\ell,i,c}}{\max_j |\mathbf{a}_{\ell,j,c}|}[q],
\end{equation}
where $q$ indexes the flattened spatial locations.
Concatenating across channels gives the mean feature vector
\begin{equation}
\label{eq:mean_feature}
    \mathbf{m}_{\ell,i}
    =
    [m_{\ell,i,1},\dots,m_{\ell,i,C_\ell}] .
\end{equation}

\paragraph{(ii) Activation-distribution features.}
The activation-distribution shape for channel $c$ is computed by first applying Z-score normalization~\cite{han2012datamining},
\begin{equation}
    \tilde{\mathbf{a}}_{\ell,i,c}^{\mathrm{Zscore}}
    =
    \frac{\mathbf{a}_{\ell,i,c}-\mu_{\ell,c}}
         {\sigma_{\ell,c}},
\end{equation}
and then computing a one-dimensional Kernel Density Estimate (KDE)~\cite{parzen1962},
\begin{equation}
\label{eq:kde}
    \hat{p}_{\ell,i,c}(t)
    =
    \frac{1}{H_\ell W_\ell \, h_{\ell,c}}
    \sum_{q=1}^{H_\ell W_\ell}
    K\!\left(
        \frac{t-\tilde{\mathbf{a}}_{\ell,i,c}^{\mathrm{Zscore}}[q]}
             {h_{\ell,c}}
      \right),
\end{equation}
where $\mu_{\ell,c}$ and $\sigma_{\ell,c}$ are computed over all spatial locations and training samples and $K$ is a Gaussian kernel with bandwidth $h_{\ell,c}$. We use Scott's Rule~\cite{scott1992multivariate} for selecting the bandwidth, providing a standard bias-variance tradeoff:
\begin{equation}
\label{eq:scottsrule}
    \phantom{.}h_{\ell,c} = 1/\sqrt[5]{H_\ell W_\ell}.
\end{equation}
To obtain fixed-length features, we sample the KDE over a Z-score interval covering $P\%$ of activation mass,
\begin{equation}
[z_{\min},z_{\max}] =
\left[
    \Phi^{-1}\!\Bigl(\tfrac{1-P/100}{2}\Bigr),
    \Phi^{-1}\!\Bigl(\tfrac{1+P/100}{2}\Bigr)
\right],
\end{equation}
 where $\Phi^{-1}(\cdot)$ is inverse cumulative distribution function of the standard normal distribution~\cite{bliss1934probit}, using $R$ uniformly spaced points,
\begin{equation}
    \hat{p}_{\ell,i,c}^{(r)}
    =
    \hat{p}_{\ell,i,c}
    \!\left(
        z_{\min}
        + r\,\frac{z_{\max}-z_{\min}}{R-1}
    \right),
\end{equation}
where $r \in [0, R-1]$. The per-channel KDE vector is
\begin{equation}
    \mathbf{k}_{\ell,i,c}
    =[\hat{p}_{\ell,i,c}^{(0)},\dots,\hat{p}_{\ell,i,c}^{(R-1)}] ,
\end{equation}
and concatenation across channels yields
\begin{equation}
\label{eq:kde_feature}
    \mathbf{k}_{\ell,i} =[\mathbf{k}_{\ell,i,1},\dots,\mathbf{k}_{\ell,i,C_\ell}] .
\end{equation}

\paragraph{(iii) Free-energy features.} The absolute energy for channel $c$ is computed as~\cite{liu2020energy}
\begin{equation}
    e_{\ell,i,c} = - \log \sum_{q=1}^{H_\ell W_\ell} \exp\!\bigl(\mathbf{a}_{\ell,i,c}[q]\bigr),
\end{equation}
and concatenation across channels yields 
\begin{equation}
    \mathbf{e}_{\ell,i}
    =
    [e_{\ell,i,1},\dots,e_{\ell,i,C_\ell}] .
\end{equation}

\textbf{SNR-Based Complementary Features.} Let $p_i \in \mathbb{R}$ denote the estimated Signal-to-Noise-Ratio (SNR) of input $x_i$. SNR estimate $p_i$ can be used directly as an auxilary feature or through the SNR distance 
\begin{equation}
    d_{\mathrm{SNR}}(i,j) = |p_i - p_j|.
\end{equation}
To better separate samples that are close in activation space but correspond to different channel conditions, we incorporate SNR into OOD detectors. In particular, we use conditional k-NN schemes, where neighbor selection is either restricted to samples with similar SNR (hard conditioning) or smoothly weighted by SNR mismatch (soft conditioning), analogous to stratified k-NN~\cite{katila2002} but with a continuous auxiliary variable. Full definitions are provided in Supplemental~\ref{supplemental:conditional_knn}.

\textbf{OOD Modeling.}
Let $\mathbf{g}_{\ell,i}$ denote the feature for layer~$\ell$ and sample~$i$. OOD detectors are fitted on the In-Distribution (ID) training set and produce a scalar OOD score. HDBSCAN and k-NN rely on an explicit distance metric $d(\cdot,\cdot)$, whereas OC-SVM and Mahalanobis induce distances implicitly via a learned decision boundary or covariance structure.

\paragraph{(i) Density-based (HDBSCAN).}
We cluster ID features using HDBSCAN*~\cite{campello2013hdbscan,mcinnes2017accelerate} and use the cluster-membership strength
\begin{equation}
    s_{\ell,i}^{\mathrm{HDBSCAN}}
    =
    1 - \mathrm{membership}\bigl(\mathbf{g}_{\ell,i}\bigr),
\end{equation}
since low membership indicates noise and is treated as OOD.

\paragraph{(ii) Neighbourhood-based (k-NN).} We train on ID features and use the mean distance to the $k$-nearest neighbours~\cite{cover1967},
\begin{equation}
    s_{\ell,i}^{\mathrm{kNN}}
    =
    \frac{1}{k}
    \sum_{j \in \mathcal{N}_k(i)}
        d(\mathbf{g}_{\ell,i}, \mathbf{g}_{\ell,j}),
\end{equation}
where $\mathcal{N}_k(i)$ denotes the $k$-nearest neighbours in training set for sample $i$. We do not apply Deep $k$-Nearest Neighbors~\cite{papernot2018deepknn}, as it relies on class-label agreement, which is undefined for multi-label soft-bit outputs.

\paragraph{(iii) Boundary-based (One-Class SVM).}
We train on ID features and use the distance to decision boundary~\cite{scholkopf2001ocsvm,bounsiar2014ocsvm},
\begin{equation}
    s_{\ell,i}^{\mathrm{OCSVM}} = -b_\ell(\mathbf{g}_{\ell,i}),
\end{equation}
where $b_\ell(\mathbf{g}_{\ell,i})=\sum_j \alpha_{\ell,j} K(\mathbf{g}_{\ell,i},\mathbf{g}_{\ell,j})-\rho$
is the learned decision function (positive inside the ID region) with kernel $K$.

\paragraph{(iv) Covariance-based (Gaussian Mahalanobis).}
Assuming Gaussian ID features,
\begin{equation}
    s_{\ell,i}^{\mathrm{Mahalanobis}}
    =
    \bigl(\mathbf{g}_{\ell,i}-\mu_\ell\bigr)^\top
    \Sigma_\ell^{-1}
    \bigl(\mathbf{g}_{\ell,i}-\mu_\ell\bigr),
\end{equation}
where $(\mu_\ell, \Sigma_\ell)$ are the empirical mean and covariance~\cite{mahalanobis1936generalised}.
We do not adopt the class-conditional Mahalanobis variants of~\cite{lee2018mahalanobis,muller2025mahalanobis}, as they rely on per-class distributions, which are ill-defined for multi-label receiver outputs.

\textbf{Fusion Across Detectors and Layers (Optional).}
Each layer $\ell$ yields a set of scores
\begin{equation}
    \mathbf{s}_{\ell,i}
    =
    \bigl[
        s_{\ell,i}^{\text{kNN}},
        s_{\ell,i}^{\text{Mahalanobis}},
        s_{\ell,i}^{\text{HDBSCAN}},
        s_{\ell,i}^{\text{OCSVM}},
        p_i
    \bigr].
\end{equation}

\emph{Within-layer fusion} combines (two or more) detector outputs at a single layer, either by fitting an OC-SVM on scores or by converting the scores to conformal $p$-values and applying a Benjamini--Hochberg (BH)~\cite{benjamini1995bh} style combination rule~\cite{magesh2023conformalpvalues}. Within-layer fusion leverages classifier diversity.

\emph{Across-layer fusion} combines layer-level scores (of one or more detectors) either by fitting OC-SVM or applying BH style combining, yielding a continuous OOD score via Supplemental~\ref{supplemental:extension_to_conformal_p}. Layer fusion leverages across-layer diversity.

\textbf{OOD Decision.} Let $S_i$ denote the final OOD score for sample $x_i$. A binary decision is obtained by thresholding,
\begin{equation}
    x_i \text{ is OOD} \quad \Longleftrightarrow \quad S_i \ge \tau ,
\end{equation}
where the threshold $\tau$ is chosen to satisfy a desired operating point
(e.g., FPR@95\%TPR). The continuous score $S_i$ can also be used to generate Detection Error Tradeoff (DET) curves~\cite{martin1997det}.

Our framework is post-hoc: the receiver parameters $\theta$ are kept frozen. It is not model-agnostic, as feature extraction assumes convolutional layers with channelwise structure, as in ResNet-like neural receivers such as DeepRx~\cite{honkala2021deeprx}.
The block diagram in Fig.~\ref{fig:block_diagram} depicts the full dataflow.

\input{table_simulation_parameters}

\begin{figure}[h]
\centering
\includegraphics[width=0.98\columnwidth]{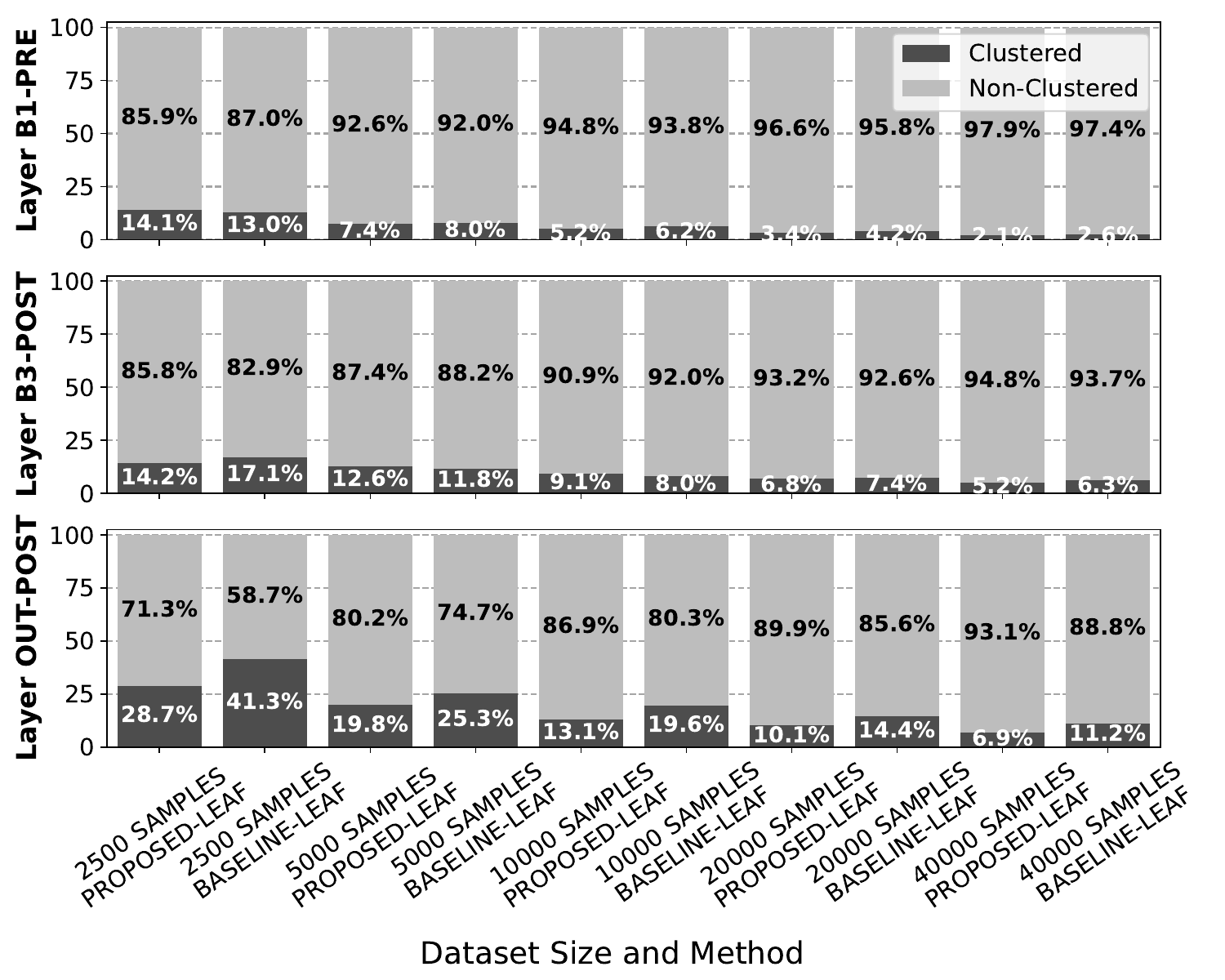}
\caption{Share of clustered and non‑clustered samples versus sample size and aggregation method (minimum cluster size = 5).}
\label{fig:clustered_non_clustered_vs_sample_size}
\end{figure}

\begin{figure}[h]
\centering
\includegraphics[width=0.98\columnwidth]{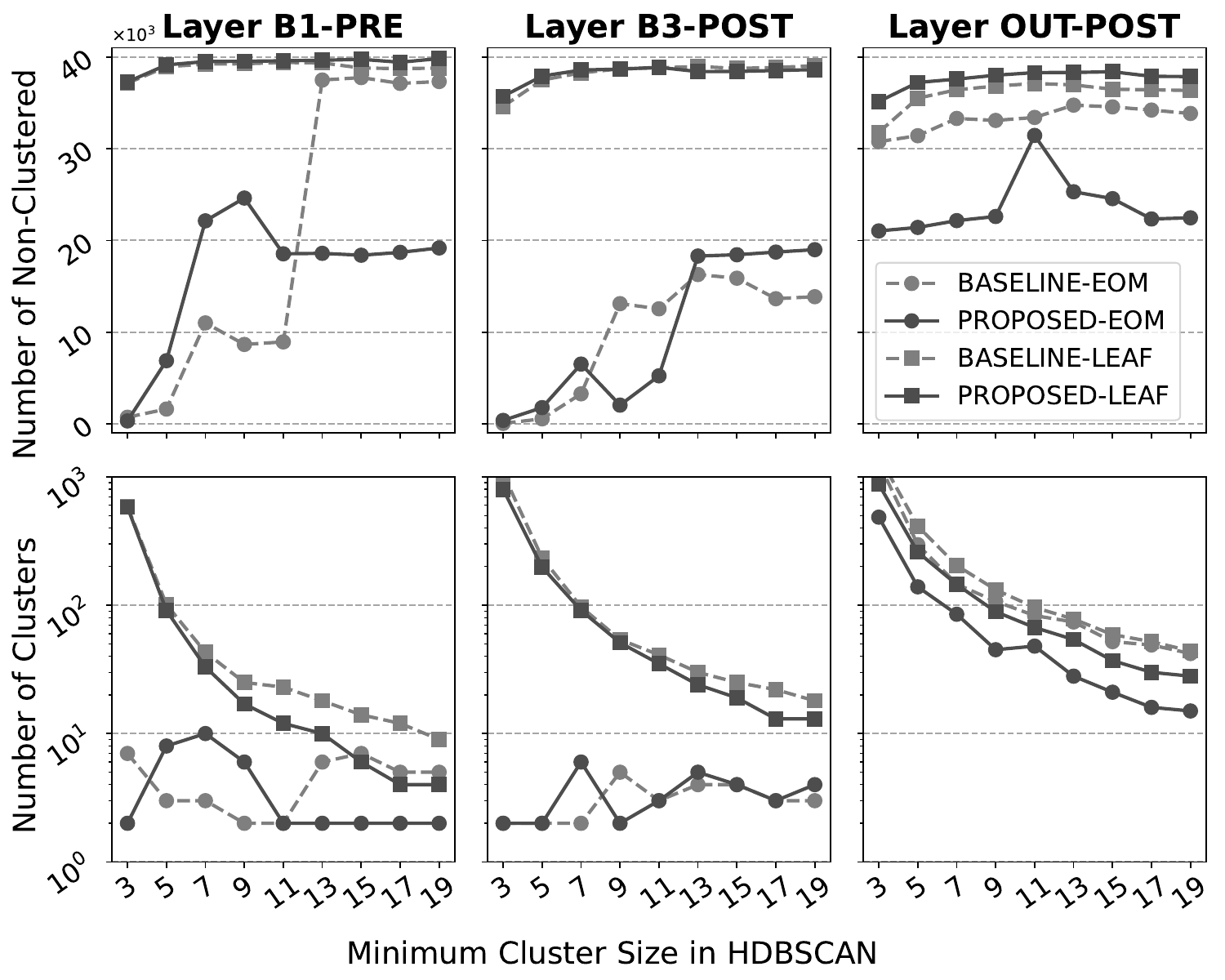}
\caption{Counts of non‑clustered samples and clusters versus minimum cluster size, by aggregation and cluster‑selection method.}
\label{fig:hdbscan_parameters}
\end{figure}

\section{Cluster and Manifold Analysis}
\label{sec:cluster_and_manifold_analysis_experiments}
We study whether channelwise activation features from the neural receiver (Supplemental~\ref{supplemental:radio_receiver_model}) form coherent clusters reflecting channel conditions, and how these features are organized on lower-dimensional manifolds. This analysis (i) provides insight into the receiver’s learned internal representations and (ii) informs whether clustering-based or manifold-aware OOD detectors are appropriate.

\begin{figure*}[h]
\centering
\includegraphics[trim={0 0 0 0cm},clip,width=0.99\textwidth]{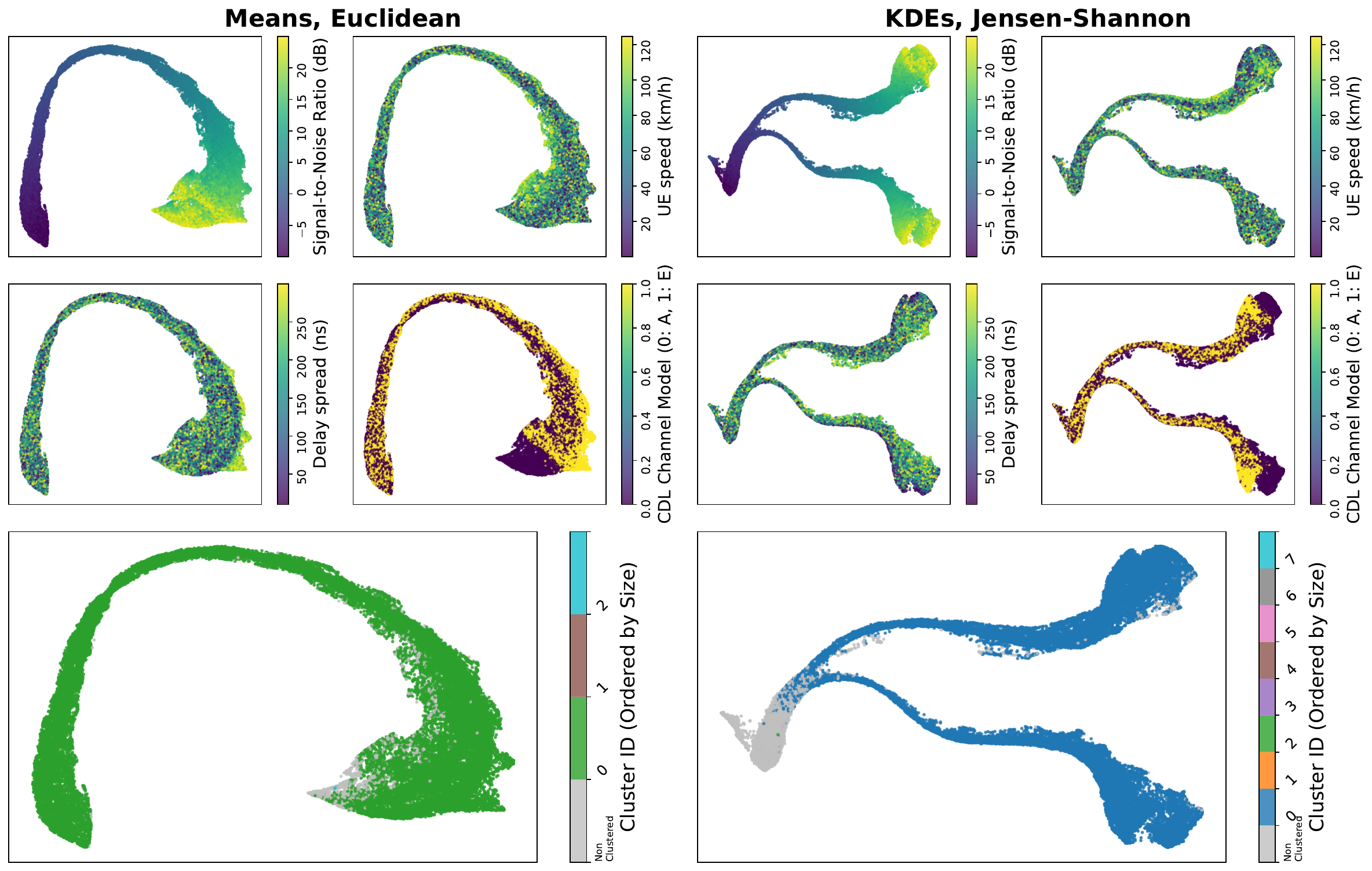}
\caption{Uniform Manifold Approximation and Projection (UMAP) embeddings of DeepRx activations from layer B1-PRE using the baseline (left) and proposed (right) methodology, colored by channel parameters and partitions with HDBSCAN (EoM, min. cluster size 5).}
\label{fig:data_structure}
\end{figure*}

\subsection{Experimental Setup}
\label{sec:experimentalsetup_cluster_analysis}
We evaluate two feature families introduced in Section~\ref{sec:methodology}: (i) mean-activation features (Eq.~\ref{eq:mean_feature}) with Euclidean distance~\cite{euclid1956}, as in~\cite{bauerle2022nap}; and (ii) KDE-based activation-distribution features (Eq.~\ref{eq:kde_feature}) with Jensen--Shannon distance~\cite{endres2003jsmetric}, representing one of our methodological extensions.

Data are generated with a 5G-compliant PUSCH link-level simulator (NVIDIA Sionna~\cite{hoydis2023sionna}) using the parameters in Table~\ref{tab:simulation_parameters}. We reimplemented both the baseline NAP and our extensions in PyTorch~\cite{paszke2019pytorch} using standard scientific Python libraries (Supplementals~\ref{supplemental:baseline_nap_methodology}--\ref{supplemental:nap_implementation_details}). For KDE features, each per-channel distribution is sampled at $R=10$ Z-scores covering $99\%$ of the standard-normal mass, producing a compact ten-dimensional descriptor per channel.

To explore activation-space structure, we apply HDBSCAN*~\cite{campello2013hdbscan,mcinnes2017accelerate} with both Leaf and Excess-of-Mass (EoM) selection criteria over a sweep of minimum cluster sizes. UMAP~\cite{mcinnes2018umap} is used \emph{only} for visualization, with default Scikit-learn~\cite{scikit2011} settings.

\subsection{Results}
\label{sec:results_cluster_analysis}
\textbf{Absence of discrete, stable clusters.} Across layers, feature types, and HDBSCAN configurations, we observe no stable or semantically meaningful clustering structure. Under the Leaf criterion used by~\cite{bauerle2022nap}, both mean and KDE-based features yield very high proportions of unclustered samples (often exceeding 90\%; Fig.~\ref{fig:clustered_non_clustered_vs_sample_size}). Switching to the Excess-of-Mass (EoM) criterion reduces the number of unclustered samples but produces degenerate solutions when the minimum cluster size is reduced (most samples collapsing into one diffuse cluster; Fig.~\ref{fig:hdbscan_parameters}), and internal validation metrics (DBCV, persistence; Supplemental~\ref{supplemental:example_case_cluster_statistics}) remain near zero. These findings are consistent with our visual object recognition experiments (Supplemental~\ref{supplemental:nap_analysis_visual}), suggesting that the lack of cluster structure reflects fundamental properties of activation geometry rather than domain specifics.

\textbf{Discovery of an SNR-aligned manifold.} While clustering fails, a clear low-dimensional manifold emerges when visualized with UMAP (Fig.~\ref{fig:data_structure}, Supplemental~\ref{supplemental:umap_repetition_radio}). Activations form a smooth continuum primarily parameterized by SNR, with the channel model contributing only small secondary variations in the high-SNR regime. This behavior is consistent with classical physical-layer receiver design, where internal processing adapts smoothly as a function of estimated SNR~\cite{goldsmith2005}. This observation also aligns with our independent replication using the methodology of~\cite{tuononen2025interpreting}, where user speed and delay spread exhibited similarly limited influence (Supplemental~\ref{supplemental:nucpi_experiments_on_uespeed_and_delayspread}). KDE-based aggregation sometimes reveals two SNR-aligned branches at high SNR that merge at low SNR, suggesting a richer representation than the mean-activation baseline, though the global geometry remains dominated by a single SNR-driven trajectory. This regression-like latent structure is consistent with prior observations that deep regression models tend to produce continuous high-dimensional manifolds rather than discrete clusters~\cite{nilsson2007regressionmanifolds,bengio2013representation,islam2023}.

\textbf{Implications for OOD detection.} The manifold structure has direct consequences for OOD modeling: cluster-based OOD detection (requiring multiple coherent clusters) is unsuitable because ID data do not form compact, well-separated regions. Distance- and distribution-based OOD methods are better aligned with the activation geometry, especially when augmented with SNR information. The clear ordering of activations by SNR directly motivates the SNR-aware and conditional distance metrics introduced in Section~\ref{sec:methodology}.

\section{Out-Of-Distribution (OOD) Detection}
\label{sec:ood_experiments}
Next we evaluate the proposed OOD detection framework, quantifying the performance of feature–model combinations, the utility of SNR as an auxiliary signal, and the benefits of within‑ and across‑layer fusion.

\subsection{Experimental Setup}
The experiments use the same dataset as in Section~\ref{sec:cluster_and_manifold_analysis_experiments} (Table~\ref{tab:simulation_parameters}). The In-Distribution (ID) set spans a wide range of SNRs, delay spreads, and user speeds. Two OOD scenarios--high delay and high speed--are evaluated, introducing substantial channel mismatch while keeping SNR within the ID range. This prevents trivial SNR-based detection and isolates SNR as an auxiliary signal.

We first establish single-model baselines on the earliest convolutional layer (B1-PRE), motivated by the hypothesis--later confirmed--that earlier layers retain richer information for OOD detection. Using these baselines, we perform two ablations: incorporating SNR to exploit feature diversity, and within-layer fusion to exploit classifier diversity. The same baselines are then evaluated on B3-POST and OUT-POST to enable across-layer fusion.

\textbf{Feature extraction.} We evaluate mean features, KDE features, and free-energy features (Section~\ref{sec:methodology}). Free-energy features are computed with temperature $T=1$. KDE features are as in Section~\ref{sec:cluster_and_manifold_analysis_experiments}. SNR is incorporated for k-NN via (i) raw absolute distance for neighboring points, (ii) hard conditioning, or (iii) soft conditioning (Supplemental~\ref{supplemental:conditional_knn}). SNR is incorporated for OC-SVM and Gaussian Mahalanobis by concatenating the activation features and raw SNR values.

\textbf{Modeling methods and metrics.} We evaluate four OOD detector families: density-based (HDBSCAN*, EoM, minimum cluster size 5; Euclidean~\cite{euclid1956}, cosine~\cite{salton1975}, or Jensen--Shannon distances~\cite{endres2003jsmetric}), neighbourhood-based (k-NN with $K=5$ using the same distances), boundary-based (OC-SVM with RBF kernel), and covariance-based (Gaussian Mahalanobis with Ledoit--Wolf covariance~\cite{ledoit2003lw}).

\textbf{Classifier fusion.} Performed within-layer and across-layer using (i) OC‑SVM on concatenated detector scores and raw SNR, or (ii) BH‑style conformal p‑value fusion (Section~\ref{sec:methodology}).

\textbf{Evaluation metrics.} Followed common OOD practice: Area Under Receiver Operating Characteristic (AUROC), False Positive Rate at 95\% True Positive Rate (FPR95), and OOD detection latency. Results also visualized as the Detection Error Tradeoff (DET) curve~\cite{martin1997det}.

\subsection{Results}
\textbf{Single-model baselines on B1-PRE.}
The best average AUROC across both OOD scenarios selects the following baselines (Table~\ref{tab:single_model_baselines_on_b1-pre_average_over_scenarios}):
(A) HDBSCAN: mean features + Euclidean distance;
(B) k-NN: mean features + cosine distance;
(C) OC-SVM: mean features; and
(D) Mahalanobis: mean features.
Mean features consistently outperform free-energy and KDE features, despite the latter carrying richer distributional information. Mahalanobis achieves the highest AUROC overall, with k-NN close behind; OC-SVM performs clearly worse. HDBSCAN yields reasonable AUROC in the high-delay scenario (e.g., 0.92) but \emph{always} fails in FPR95 (equal to 1.0), providing no usable operating point. Scenariowise\footnote{\label{detailedresultsnote}See Supplemental~\ref{supplemental:results} for the detailed results.}, high delay is essentially solved ($\text{AUROC}\!>\!0.99$, $\text{FPR95}\!\approx\!2\%$), whereas high speed remains difficult ($\text{AUROC}\!\approx\!0.74$, $\text{FPR95}\!\approx\!79\%$).

\textbf{Effect of SNR on B1-PRE.}
We evaluate SNR integration via conditional k-NN and via concatenation for Mahalanobis and OC-SVM\footref{detailedresultsnote}.
In the high-speed scenario, conditional k-NN yields a small improvement (AUROC~0.7358 vs.~0.7339).  
In the high-delay scenario, adding SNR slightly degrades AUROC (0.9922 vs.~0.9931).  
Soft conditioning helps in high speed but hurts in high delay, with hard conditioning showing the opposite trend.  
Concatenating SNR for OC-SVM and Mahalanobis brings little to no benefit.  
Overall, SNR yields only marginal AUROC gains (typically fourth decimal place).

\textbf{Within-layer fusion.}
We evaluate 27 pairwise and multi-way fusions among the four baselines and raw SNR for each layer\footref{detailedresultsnote}.
Earlier layers are more informative for OOD detection:
$
    \text{AUROC}_{\text{B1-PRE}} > \text{AUROC}_{\text{B3-POST}} > \text{AUROC}_{\text{OUT-POST}}.
$
OC-SVM applied to detector outputs is the most effective fusion method, outperforming BH-style conformal fusion.  
Typical best combinations include Mahalanobis~+~SNR for high delay, while the optimal combination for high speed is layer-dependent.  
On B1-PRE, for example:  
high delay--Mahalanobis~+~SNR increases AUROC from 0.9965 to 0.9966; 
high speed--k-NN baseline~+~OC-SVM baseline~+~SNR increases AUROC from 0.7339 to 0.7550.  
Overall, within-layer fusion yields modest AUROC gains, primarily in challenging cases such as the OUT‑POST layer and high‑speed scenarios (Tables~\ref{tab:main_results_high_delay}--\ref{tab:main_results_high_speed} with OC-SVM fusion).

\textbf{Across-layer fusion.}\footref{detailedresultsnote}
We extend fusion from within a single layer to jointly combine detector outputs from the B1-PRE, B3-POST, and OUT-POST layers using OC-SVM (Tables~\ref{tab:main_results_high_delay}--\ref{tab:main_results_high_speed}). Across-layer fusion yields complementary AUROC gains, most notably in challenging high-speed scenarios, where both k-NN and Gaussian Mahalanobis benefit.

\begin{figure}[h!]
\centering
\includegraphics[width=\columnwidth]{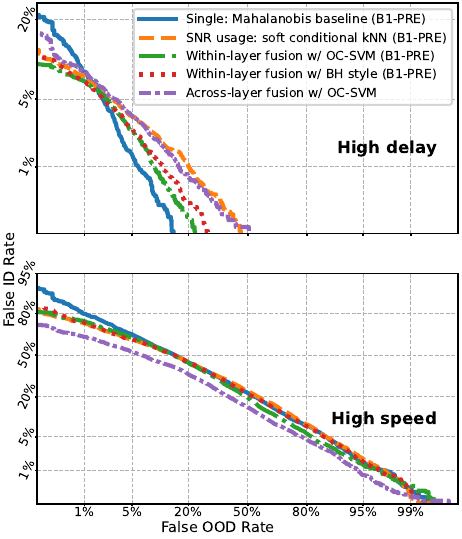}
\caption{DET curves (single repeat; negligible SD) across scenarios, with selected detectors from Tables~\ref{tab:main_results_high_delay}–\ref{tab:main_results_high_speed} and Supplemental~\ref{supplemental:results}.}
\label{fig:main_det_curves}
\end{figure}

\textbf{DET curve summary.} Mahalanobis has a steady performance (Figure~\ref{fig:main_det_curves})--especially under high delay scenario--and has the lowest latency (Tables~\ref{tab:main_results_high_delay}–\ref{tab:main_results_high_speed}). When latency is less critical and lower false OOD is prioritized, use conditional k‑NN with SNR or within‑layer fusion on B1‑PRE; for harder cases (e.g., high speed), fuse across layers for added robustness.

\begin{table*}[h!]
    \centering
    \input{table_single_model_baselines_on_b1-pre_average_over_scenarios}
    \caption{OOD detection on layer B1‑PRE averaged over scenarios (mean of five runs; negligible SD). Best feature–distance pairs per metric are \textbf{bold}; AUROC‑best pairs were chosen as baselines for subsequent experiments. Scenario‑wise results provided in the Supplemental~\ref{supplemental:results}.}
\label{tab:single_model_baselines_on_b1-pre_average_over_scenarios}
\end{table*}

\begin{table}
    \centering
    \input{table_main_results_high_delay}
    \caption{OOD detection across layers under high delay (mean of five runs; negligible SD); best method–layer per metric are \textbf{bold}.}
    \label{tab:main_results_high_delay}
\end{table}

\begin{table}
    \centering
    \input{table_main_results_high_speed}
    \caption{OOD detection across layers under high speed (mean of five runs; negligible SD); best method–layer per metric are \textbf{bold}.}
    \label{tab:main_results_high_speed}
\end{table}

\section{Conclusions}
We introduced a post-hoc OOD detection framework for neural radio receivers and analyzed the geometry of their activation spaces. DeepRx does not form discrete or semantically meaningful activation clusters--most samples remain unclustered even with improved feature extraction and modeling settings--mirroring our visual object recognition results and suggesting that such models inherently lack cluster structure. Instead, activations lie on a low-dimensional manifold dominated by SNR, with other channel parameters contributing only minor variation. This smooth, regression-like geometry aligns with classical physical-layer behavior and offers a natural basis for OOD modeling.

In OOD experiments, high‑delay conditions are detected reliably, whereas high‑speed remains challenging. Mahalanobis distance on mean activations is the strongest single detector; OC‑SVM provides the most stable fusion; earlier layers outperform later ones; and SNR integration and classifier fusions yield only small, inconsistent AUROC gains--useful mainly when latency is acceptable and extremely low false OOD rate is preferred.

Overall, OOD detectors for neural receivers should be manifold-aware and leverage early-layer representations rather than clustering. Future work includes latent-variable models (e.g., DCAE, $\beta$-VAE) to better capture activation structure and improved integration of SNR or other channel-state information. These advances support reliable, trustworthy AI-native communication in 6G and beyond.

\appendix

\ifanonymized
\else
    \section*{Acknowledgments}
    Marko Tuononen was partially supported by the Nokia Foundation. Ville Hautamäki was partially supported by Jane and Aatos Erkko Foundation. We acknowledge the creators of the DeepRx framework for their foundational contributions.
\fi

\putbib[ijcai26]
\end{bibunit}

\begin{bibunit}[named]
\input{supplemental/main}
\putbib[ijcai26]
\end{bibunit}

\end{document}

%% file: figure_block_diagram.tikz
\begin{tikzpicture}[>=Stealth,
  box/.style={
    draw,
    minimum width=22mm,
    minimum height=15mm,
    align=center,
    font=\sffamily\scriptsize,
    anchor=center
  },
  row sep=5mm, column sep=5mm
]
  \matrix (m) [matrix of nodes, nodes=box] {
    \shortstack{Raw physical-\\layer signal\\$X=\{x_i\}$} &
    \shortstack{Frozen neural\\receiver\\$F_{\theta}(x_i)$} &
    \shortstack{Layerwise\\activations\\$A_l(x_i)$} \\
    \shortstack{OOD score\\and decision\\$S_i\geq\tau$} &
    \shortstack{OOD detector\\(e.g. k-NN)\\$S_i=D(g_{l,i})$} &
    \shortstack{Channelwise\\aggregated +\\SNR features\\$g_{l,i}$ ; $p_i$} \\
  };

  \draw[->] (m-1-1) -- (m-1-2);
  \draw[->] (m-1-2) -- (m-1-3);

  \draw[->] (m-1-3) -- (m-2-3);

  \draw[->] (m-2-3) -- (m-2-2);
  \draw[->] (m-2-2) -- (m-2-1);
\end{tikzpicture}

%% file: table_simulation_parameters.tex
\begin{table*}
    \centering
    \caption{Simulation parameters for DeepRx and Out‑Of‑Distribution (OOD) detector training/testing. In‑Distribution (ID) testing mirrors the OOD‑training parameters. OOD scenarios: (i) fixed 1000 ns delay spread (very long multipath; Table 7.7.3‑1 of \protect\cite{3gpp38901}), (ii) fixed 866 Hz Doppler (approximately 100 m/s at 2.6 GHz; high‑speed rail). These OOD settings exceed DeepRx’s training range yet remain realistic. OOD testing uses the Clustered Delay Line (CDL)‑C model to induce a channel‑model mismatch; training uses CDL‑A and CDL‑E.}
    \label{tab:simulation_parameters}
    \begin{tabular}{l c c c c}
        \toprule
        \textbf{Parameter} & \textbf{DeepRx train} & \textbf{Detector train} & \textbf{Detector OOD test} & \textbf{Detector OOD test} \\
        &  & \textbf{and ID test} & \textbf{High DS} & \textbf{High Speed}\\
        \midrule
        \multicolumn{3}{l}{\textit{Uniformly Distributed Channel Parameters}} \\
        Signal-to-Noise Ratio (SNR) & -10--25~dB & -10--25~dB & -10--25~dB & -10--25~dB \\
        Doppler Shift & 0--330~Hz & 0--300~Hz & 0--300~Hz & 866~Hz \\
        Delay Spread & 10--300~ns & 10--300~ns & 1000~ns & 10--300~ns \\
        Channel Models & UMi, UMa & CDL-A, E (train) & CDL-C & CDL-C \\
        & & CDL-C (test) & & \\
        \midrule
        \multicolumn{5}{l}{\textit{Other Parameters}} \\
        Number of PRBs & \multicolumn{4}{c}{16 (192 subcarriers)} \\
        Subcarrier Spacing & \multicolumn{4}{c}{30 kHz} \\
        TTI Length & \multicolumn{4}{c}{14 OFDM symbols / 0.5 ms} \\
        Modulation Scheme & \multicolumn{4}{c}{16-QAM} \\
        RX / TX Antennas & \multicolumn{4}{c}{2 / 1} \\
        Pilot Symbols per Slot & \multicolumn{4}{c}{1 or 2 OFDM symbols} \\
        Pilot Symbol Distribution & \multicolumn{4}{c}{Uniformly distributed, fixed locations} \\
        \bottomrule
    \end{tabular}
\end{table*}

%% file: table_single_model_baselines_on_b1-pre_average_over_scenarios.tex
\renewcommand{\arraystretch}{0.95}
\begin{tabular}{lllllllll}
\toprule
\multirow{2}{*}{Feature + Distance} & \multicolumn{2}{c}{HDBSCAN} & \multicolumn{2}{c}{k-NN} & \multicolumn{2}{c}{OC-SVM} & \multicolumn{2}{c}{Mahalanobis} \\
{} &            AUROC &            FPR95 &            AUROC &            FPR95 &                             AUROC &                             FPR95 &                             AUROC &                             FPR95 \\
\midrule
Mean + Euclidean     &  \textbf{0.7868} &  \textbf{1.0000} &           0.8205 &           0.5557 &  \multirow{2}{*}{\textbf{0.7694}} &  \multirow{2}{*}{\textbf{0.5669}} &  \multirow{2}{*}{\textbf{0.8658}} &           \multirow{2}{*}{0.4384} \\
Mean + Cosine        &           0.6823 &  1.0000 &  \textbf{0.8635} &           0.4419 &                                   &                                   &                                   &                                   \\
KDE + Euclidean      &           0.7806 &  1.0000 &           0.8554 &           0.4327 &           \multirow{3}{*}{0.7385} &           \multirow{3}{*}{0.6572} &           \multirow{3}{*}{0.8605} &  \multirow{3}{*}{\textbf{0.4223}} \\
KDE + Cosine         &           0.7855 &  1.0000 &           0.8542 &           0.4409 &                                   &                                   &                                   &                                   \\
KDE + Jensen-Shannon &           0.7806 &  1.0000 &           0.8603 &  \textbf{0.4222} &                                   &                                   &                                   &                                   \\
Energy + Euclidean   &           0.7478 &  1.0000 &           0.8156 &           0.5871 &           \multirow{2}{*}{0.7432} &           \multirow{2}{*}{0.6252} &           \multirow{2}{*}{0.8542} &           \multirow{2}{*}{0.4587} \\
Energy + Cosine      &           0.7387 &  1.0000 &           0.8208 &           0.5617 &                                   &                                   &                                   &                                   \\
\bottomrule
\end{tabular}

%% file: table_main_results_high_delay.tex
\renewcommand{\arraystretch}{0.95}
\begin{tabular}{clrrr|r}
\toprule
\multicolumn{1}{c}{Method} & \multicolumn{1}{c}{Metric} & \multicolumn{1}{c}{\makecell{B1-\\PRE}} & \multicolumn{1}{c}{\makecell{B3-\\POST}} & \multicolumn{1}{c}{\makecell{OUT-\\POST}} & \multicolumn{1}{|c}{\makecell{Across\\layers\footref{detailedresultsnote}}} \\
\midrule
\multirow{3}{*}{\rotatebox{90}{\makecell{HDB-\\SCAN\\baseline}}} & AUROC          & 0.9209           & 0.8370           & 0.5208           & \textbf{0.9536} \\
                         & FPR95          & 1.0000           & 0.3543           & 1.0000           & \textbf{0.1483} \\
                         & Lat. ($\mu$s)  & 436.40         & 569.23         & \textbf{263.80} & 1892.3 \\
\midrule
\multirow{3}{*}{\rotatebox{90}{\makecell{\\k-NN\\baseline}}}    & AUROC          & \textbf{0.9931}  & 0.9174           & 0.6929           & 0.9926 \\
                         & FPR95          & \textbf{0.0218}  & 0.6440           & 0.8847           & 0.0312 \\
                         & Lat. ($\mu$s)  & 1926.2        & \textbf{1116.2} & 1417.0      & 5130.5 \\
\midrule
\multirow{3}{*}{\rotatebox{90}{\makecell{OC-\\SVM\\baseline}}}  & AUROC          & \textbf{0.9394}  & 0.7372           & 0.5509           & 0.9198 \\
                         & FPR95          & \textbf{0.2084}  & 0.8597           & 0.9486           & 0.2698 \\
                         & Lat. ($\mu$s)  & 2771.5        & 1722.0        & \textbf{716.51} & 5115.5 \\
\midrule
\multirow{3}{*}{\rotatebox{90}{\makecell{Mahala-\\nobis\\baseline}}} & AUROC      & \textbf{0.9965}  & 0.9410           & 0.5677           & 0.9941 \\
                            & FPR95       & \textbf{0.0200}  & 0.3327           & 0.9359           & 0.0367 \\
                            & Lat. ($\mu$s) & 4.6913         & 0.8825           & \textbf{0.1678}  & 665.10 \\
\midrule
\multirow{3}{*}{\rotatebox{90}{\makecell{Within-\\layer\footref{detailedresultsnote}}}} & AUROC & \textbf{0.9966}  & 0.9263           & 0.7753           & 0.9865 \\
                                  & FPR95 & \textbf{0.0222}  & 0.4924           & 0.7666           & 0.0359 \\
                                  & Lat. ($\mu$s) & 5448.4   & 3601.3        & \textbf{956.96} & 9663.3 \\
\bottomrule
\end{tabular}

%% file: table_main_results_high_speed.tex
\renewcommand{\arraystretch}{0.95}
\begin{tabular}{clrrr|r}
\toprule
\multicolumn{1}{c}{Method} & \multicolumn{1}{c}{Metric} & \multicolumn{1}{c}{\makecell{B1-\\PRE}} & \multicolumn{1}{c}{\makecell{B3-\\POST}} & \multicolumn{1}{c}{\makecell{OUT-\\POST}} & \multicolumn{1}{|c}{\makecell{Across\\layers\footref{detailedresultsnote}}} \\
\midrule
\multirow{3}{*}{\rotatebox{90}{\makecell{HDB-\\SCAN\\baseline}}} & AUROC          & 0.6528           & 0.6955           & 0.5096           & \textbf{0.7198} \\
                         & FPR95          & 1.0000           & \textbf{0.5966}  & 1.0000           & 0.9051 \\
                         & Lat. ($\mu$s)  & 546.26         & \textbf{309.72} & 315.09        & 1790.6 \\
\midrule
\multirow{3}{*}{\rotatebox{90}{\makecell{\\k-NN\\baseline}}}    & AUROC          & 0.7339           & 0.7522           & 0.6620           & \textbf{0.8054} \\
                         & FPR95          & 0.8620           & 0.9138           & 0.9235           & \textbf{0.7762} \\
                         & Lat. ($\mu$s)  & 4068.9        & \textbf{1202.9} & 1255.9      & 7170.3 \\
\midrule
\multirow{3}{*}{\rotatebox{90}{\makecell{OC-\\SVM\\baseline}}}  & AUROC          & 0.5994           & \textbf{0.7387}  & 0.4458           & 0.7068 \\
                         & FPR95          & 0.9253           & \textbf{0.8957}  & 0.9919           & 0.9232 \\
                         & Lat. ($\mu$s)  & 2084.1        & 1912.5        & \textbf{842.64} & 4610.1 \\
\midrule
\multirow{3}{*}{\rotatebox{90}{\makecell{Mahala-\\nobis\\baseline}}} & AUROC      & 0.7351           & 0.7472           & 0.5204           & \textbf{0.7941} \\
                            & FPR95       & 0.8569           & 0.8615           & 0.9465           & \textbf{0.7659} \\
                            & Lat. ($\mu$s) & 2.3000         & 1.0221           & \textbf{0.1878}  & 639.36 \\
\midrule
\multirow{3}{*}{\rotatebox{90}{\makecell{Within-\\layer\footref{detailedresultsnote}}}} & AUROC & 0.7550           & 0.7757           & 0.7042           & \textbf{0.8073} \\
                                  & FPR95 & 0.8184           & 0.9241           & 0.8742           & \textbf{0.7922} \\
                                  & Lat. ($\mu$s) & 6873.0   & 3785.7        & \textbf{1010.2} & 11271 \\
\bottomrule
\end{tabular}

%% file: supplemental/main.tex
\appendix
\setcounter{page}{0}
\renewcommand{\thepage}{A-\arabic{page}}

\part*{Supplemental Material}
\addcontentsline{toc}{part}{Supplemental Material}
\etocsetnexttocdepth{2}
\localtableofcontents

\newpage
\input{supplemental/nap}
\clearpage
\input{supplemental/deeprx}
\newpage
\input{supplemental/nucpi}
\clearpage
\input{supplemental/umap}
\input{supplemental/conditional_knn}
\input{supplemental/conformal_p_extension}
\newpage
\input{supplemental/ood}
\clearpage

%% file: supplemental/nap.tex
\section{Neural Activation Pattern (NAP) Methodology for Concept Discovery}

\subsection{Introduction and Motivation}
Neural networks are increasingly used in high-stakes applications (e.g., \cite{jumper2021alphafold})---yet their internal representations remain difficult to interpret (\cite{molnar2022}). This opacity complicates debugging, trust, and accountability, especially in domains where model behavior must align with physical or safety constraints (\cite{huyen2022,adebayo2020debugging}). With regulatory frameworks like the EU AI Act mandating transparency in AI systems (\cite{euaiact2024,euaiethics2019}), interpretable machine learning has become both a technical and societal necessity.

6G is a high-stakes application, projected to reach a \$68.7B global market by 2035 with a CAGR of 76.9\% (\cite{marketsandmarkets2023}), with the first commercial deployments expected by 2030 (\cite{nokia2025vision}). It promises extreme data rates, ultra-low latency, and dense connectivity, driving the need for advances in spectrum use, ultra-massive MIMO, and adaptive signal processing (\cite{nokia2024envisioning6g}). At the core is the physical layer, which sets the fundamental limits for capacity, efficiency, and reliability. Errors here propagate upward, making its performance critical. Traditional signal processing (\cite{goldsmith2005,tse2005}) is reaching its limits---motivating a shift to data-driven models with better adaptability (\cite{hoydis2021toward}).

We explore this shift in the context of data-driven physical-layer receivers, where emerging 6G architectures are expected to replace hand-crafted pipelines with end-to-end learned models (\cite{hoydis2021toward}). Architectures like DeepRx (\cite{honkala2021deeprx}) process raw physical-layer radio signals to decode transmitted information under dynamic, noisy channel conditions. Consequently, these models introduce a new challenge: interpretability (\cite{tuononen2025interpreting}). Physical-layer inputs are non-semantic, combining transmitted data with unpredictable environmental effects (e.g., noise, fading, interference), making most standard interpretability methods ineffective (\cite{chergui2025,fiandrino2022}).

\begin{figure*}
\centering
\includegraphics[width=0.60\textwidth]{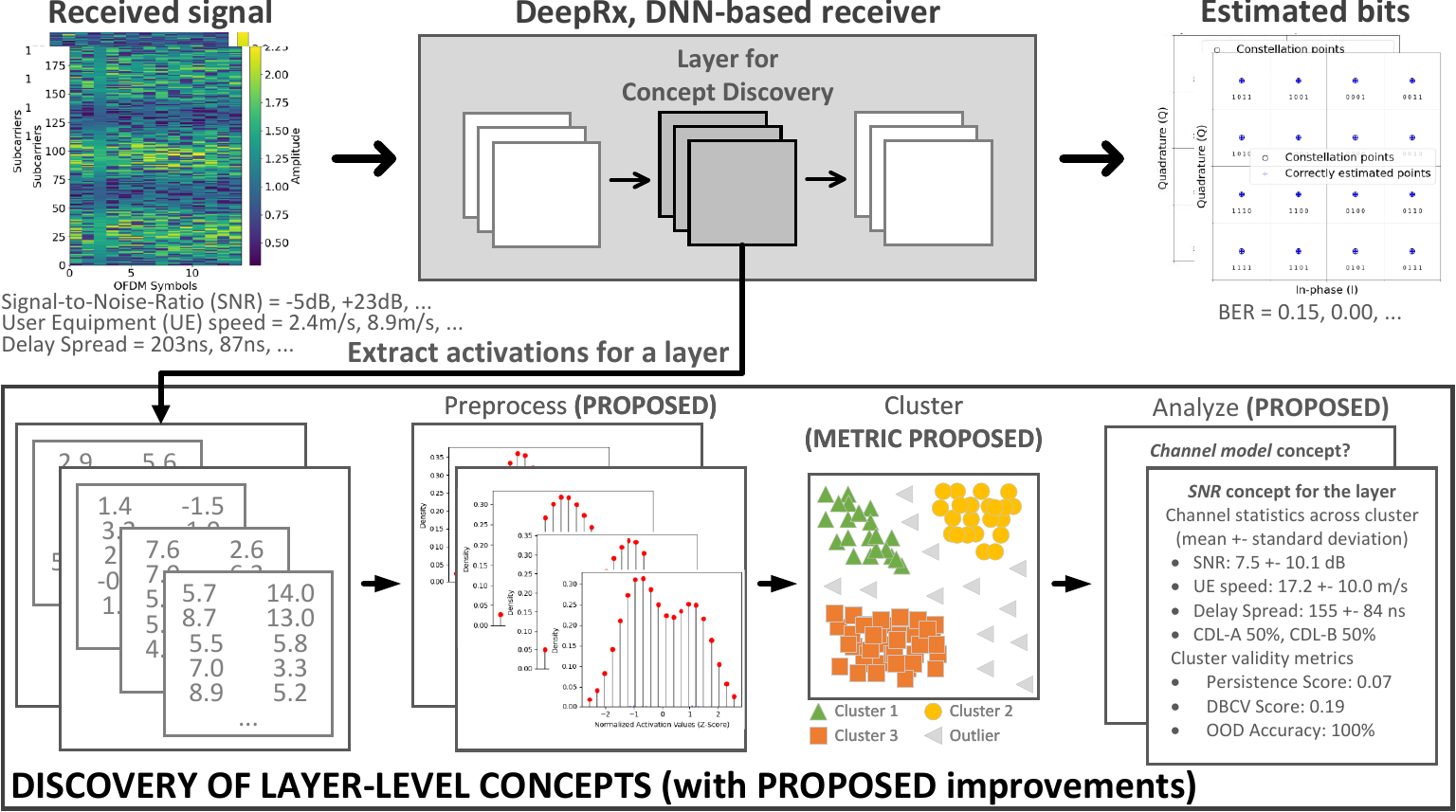}
\caption{Layer-level concept discovery with proposed improvements highlighted in \textbf{PROPOSED}. The top part illustrates the DeepRx model processing received signals to estimate transmitted bits. The bottom part outlines the concept discovery pipeline operating at the layer level.}
\label{fig:nap_high_level}
\end{figure*}

To address this, we build on the Neural Activation Pattern (NAP) methodology (\cite{bauerle2022nap}), which clusters full-layer activation distributions to identify distributed, layer-level concepts. We extend NAP with improved normalization, distribution estimation, distance metrics, and varied cluster selection---as outlined in Figure~\ref{fig:nap_high_level}---resulting in fewer unclustered samples and more stable clustering, and better out-of-distribution detection accuracy. To interpret the discovered clusters, we compare them with known physical parameters such as Signal-to-Noise Ratio (SNR). While no discrete cluster structure emerged, lower-dimensional embeddings revealed a continuous activation manifold aligned with SNR---suggesting neural receivers implicitly learn representations structured by core physical properties, even without explicit supervision.

\subsection{Related Work}
\label{supplemental:nap_relatedwork}
Interpreting neural networks has led to methods for analyzing learned representations. This section reviews concept discovery approaches, including neuron-level and human-interpretable methods, clustering of distributed activation patterns, and clustering evaluation.

\subsubsection{Concept Discovery in Neural Networks}
Approaches to concept discovery often focus on either individual neurons or broader, human-interpretable concepts but may overlook layer-wide patterns emerging from distributed interactions.

\textbf{Neuron-centric methods} analyze individual neurons to identify features that excite specific units. Techniques include saliency maps (\cite{simonyan2014saliency}), which highlight input features relevant to activations; neuron maximization (\cite{erhan2009}), which generates inputs that strongly activate specific neurons; activation histograms (\cite{yosinski2015}), which summarize neuron behaviors; and input perturbation testing (\cite{fong2017}), which evaluates the effects of input alterations on activations or outputs. Grad-CAM (\cite{selvaraju2017}) produces class-specific visualizations by localizing important regions but focuses on localized activations. Network Dissection (\cite{bau2017}) aligns neuron behaviors with predefined semantic categories (e.g., textures, objects) but fails to account for distributed patterns across layers.

\textbf{Concept-based methods} link input features to human-interpretable concepts (\cite{yeh2022}). Testing with Concept Activation Vectors (TCAV) (\cite{kim2018}) measures how user-defined concepts (e.g., stripes, wheels) influence predictions, while automated approaches (\cite{ghorbani2019}) reduce reliance on manual definitions. These methods have been extended to tabular data (\cite{pendyala2022}), made more efficient (\cite{schmalwasser2025}), and improved on a more global level~\cite{zhang2025gcav}, but they still rely on predefined or discovered concept sets and may miss broader layer-wide patterns or interactions.

Modern deep networks often learn distributed and poly-semantic features, where neurons respond to multiple unrelated stimuli or encode several sub-features (\cite{olah2017feature,olah2020zoom,bau2017,bau2020}). This highlights the limitations of focusing on individual neurons or predefined concepts, as individual units may learn similar or no new concepts, or interact with each other to learn a concept (\cite{kim2018,ghorbani2019,bauerle2022nap}).

\textbf{Neural Activation Pattern (NAP) methodology} (\cite{bauerle2022nap}) addresses these limitations by identifying groups of inputs with similar activation profiles across a layer, capturing \textit{layer-level concepts} without retraining the model. It clusters activation patterns to reveal distributed features and visualizes them via the NAP Magnifying Glass, which presents representative images, activation statistics, and metadata filtering for comparison across layers and models, as demonstrated in a visual object recognition task by \cite{bauerle2022nap}. NAP complements other interpretability methods and supports use cases such as TCAV-based testing or \textit{ante-hoc training}, where interpretability is integrated during model development (\cite{gupta2023,ciravegna2025}). By leveraging predefined concepts, NAP has the potential to enhance both interpretability and model performance.

\subsubsection{Clustering Methods for Concept Discovery}
\textbf{Clustering} is an unsupervised learning task that partitions data such that instances within the same cluster are more similar to each other than to those in different clusters (\cite{jain1988clustering}). Selecting an effective clustering method is essential for concept discovery, as it plays a critical role in uncovering patterns within high-dimensional activation spaces.

\textbf{Hierarchical clustering} creates a tree-like partition hierarchy. Methods can be categorized into agglomerative (bottom-up; e.g., BIRCH \cite{zhang1996}) and divisive (top-down; e.g., PNN \cite{virmajoki2004}) approaches. Agglomerative methods iteratively merge clusters based on linkage criteria such as single linkage (shortest distance), complete linkage (farthest distance), average linkage (mean distance), or Ward’s method (minimizing variance increase). Divisive methods recursively split clusters and can be monothetic (splitting by a single attribute) or polythetic (multiple attributes).

\textbf{Partitional clustering} divides data into non-overlapping clusters and methods can be categorized based on the framework proposed by \cite{tuononen2006rls}. This categorization includes optimization-based methods (e.g., K-Means \cite{steinhaus1956division,macqueen1967,lloyd1982}) that minimize an objective function such as intra-cluster variance; graph-theoretic methods (e.g., spectral clustering \cite{ng2001spectral}) that partition similarity graphs; density-based methods (e.g., DBSCAN \cite{ester1996}) that identify dense regions; grid-based methods (e.g., CLIQUE \cite{agrawal1998}), which discretize the feature space; and model-based methods (e.g., Gaussian Mixture Models \cite{dempster1977em}, Self-Organizing Maps \cite{kohonen1990som}) that fit statistical or neural network models.

\textbf{HDBSCAN} (\cite{campello2013hdbscan}) combines density-based and hierarchical clustering by using a minimum spanning tree for single-linkage agglomeration, effectively handling noise and irregular boundaries---ideal for high-dimensional data like neural activations with unknown cluster counts. Its accelerated variant, \textbf{HDBSCAN*} (\cite{mcinnes2017accelerate}), improves efficiency through optimized neighbor searches and pruning, making it well-suited for large-scale clustering tasks. The NAP framework (\cite{bauerle2022nap}) adopts this approach to reveal complex activation structures.

\subsubsection{Evaluation of Cluster Validity}
Evaluating cluster validity is essential in clustering-based concept discovery, as it determines whether the identified groupings reflect meaningful structure in the data. While many validation methods exist (\cite{hassan2024clustervalidation}), choosing the right one is nontrivial: the utility of a clustering often depends highly on the specific task and domain. As \cite{jain2010dataclustering} noted, \textit{clustering is in the eye of the beholder}.

\textbf{Cluster validation criteria} fall into three categories (\cite{jain1988clustering}): external (compare to ground truth labels, e.g., Rand index~\cite{rand1971}, Jaccard index~\cite{jaccard1901}); internal (assess intrinsic structure, e.g., Silhouette score~\cite{rousseeuw1987}, Davies–Bouldin score~\cite{davies1979}); and relative (compare multiple clustering results, e.g. \cite{halkidi2001clustering}). Key indicators of cluster quality are (\cite{jain1988clustering}) \textit{compactness} (minimizing distances within clusters) and \textit{separation} (maximizing distances between clusters). For density-based clustering algorithms, internal validation criteria should favor high-density clusters with low-density separation. Two widely used internal indices for density-based clustering are CDbw (\cite{halkidi2008density}), which uses boundary points to assess structure, and DBCV (\cite{moulavi2014density}), preferred for its more accurate density estimation and boundary-free formulation.

\subsection{Baseline Methodology}
\label{supplemental:baseline_nap_methodology}
The Neural Activation Pattern (NAP) methodology (\cite{bauerle2022nap}) addresses the limitations of focusing on individual neurons or predefined concepts---as described in Supplemental~\ref{supplemental:nap_relatedwork}---by identifying groups of inputs with similar activation profiles across a neural network layer. As a model-specific method, it relies on inner model details and provides post-hoc layer-level concept-based interpretability. The key steps in the NAP methodology are as follows:
\begin{itemize}
    \item \textbf{Data Extraction:} Given the layer of interest $l$ for neural network model $n$ and the set of inputs $X = \{x_i\ \mid i \in [1, \ldots, N]\}$ to the model, the activation output $A^l = \{a^l_i \mid i \in [1, \ldots, N]\}$ of the sub-network $n^l$ is extracted, denoted as $A^l = n^l(X)$. For convolutional layers, each $a^l_i$ is a collection of matrices (one per convolutional channel). There are no formal requirements on the input $X$ beyond those set by the neural network model. It can be the training set, test set, subsets thereof, or entirely separate data sets.

    \item \textbf{Location Disentanglement:} For convolutional layers, spatial information in activation values can act as a confounding factor (\cite{bauerle2022nap}). To address this, activations of each convolutional channel are aggregated using statistical measurements, either maximum value (highest activation), minimum and maximum values (range of activations), mean value (average amount of activations), or mean and standard deviation (average amount and spread of activations). The disentangled activations are denoted as $\tilde{A}^l = \{\tilde{a}^l_i \mid i \in [1, \ldots, N]\}$. Aggregating activations removes spatial information and reduces computational cost, making the method scalable for larger models.

    \item \textbf{Normalization:} The aggregated activation outputs are normalized by their maximum absolute activation across all inputs $X$. This \textit{max-abs scaling} is given by Equation~\eqref{eq:maxabs}, where $\tilde{a}^l_{i,u}$ represents the disentangled activations for layer $l$ and convolutional channel $u$ with input data $x_i$. Channels that can only activate positively are normalized between 0 and 1, while channels with negative activations are normalized between -1 and 1.
    \begin{equation}
        \label{eq:maxabs}
        \hat{a}^l_{i,u} = \frac{ \tilde{a}^l_{i,u} }{ \max (\left|\tilde{a}^l_u(X)\right|) }
    \end{equation}

    \item \textbf{Clustering Analysis:} The aggregated and normalized activation outputs $\hat{A}^l$ are clustered, forming a clustering $C^l = \{c^l_j \mid j \in [1, \ldots, M]\}$ with $M$ clusters, using the HDBSCAN* (\cite{mcinnes2017accelerate}). The clustering process employs the Euclidean distance metric, selects clusters from the leaves of the cluster tree, and uses the minimum cluster size of five. Each cluster $c^l_j \in C^l$ groups together a set of inputs $x_i \in X$ which produce similar activations for layer $l$. Clusters are mutually exclusive but not collectively exhaustive, meaning that no input $x_i$ can belong to more than one cluster, but some inputs $x_i$ may not belong to any cluster (i.e., they are classified as noise).

    \item \textbf{Post-Clustering Analysis:} HDBSCAN* provides information about the persistence of the clusters. \textit{Persistence score} reflects the stability or strength of a cluster; higher persistence meaning cluster being more well-defined and robust. Clusters are sorted based on their persistence scores, which help identify the most stable and reliable clusters. Clusters are considered to be just "raw mathematical groupings," while NAPs in general are clusters that reflect features learned by the network. Concepts are interpretable NAPs that align with human understanding.
\end{itemize}

\subsection{Proposed Methodology}
\label{supplemental:proposed_nap_methodology}
We refine the original NAP methodology (\cite{bauerle2022nap}; detailed in Supplemental~\ref{supplemental:baseline_nap_methodology}) to improve the stability and generalizability of the clustering results. In addition, we propose a principled approach to interpreting and evaluating clustering in the context of neural network-based systems, with a focus on radio receivers. Our refinements are shown in Figure~\ref{fig:nap_proposed_block_diagram} and described below:

\begin{figure*}
\centering
\includegraphics[width=0.70\textwidth]{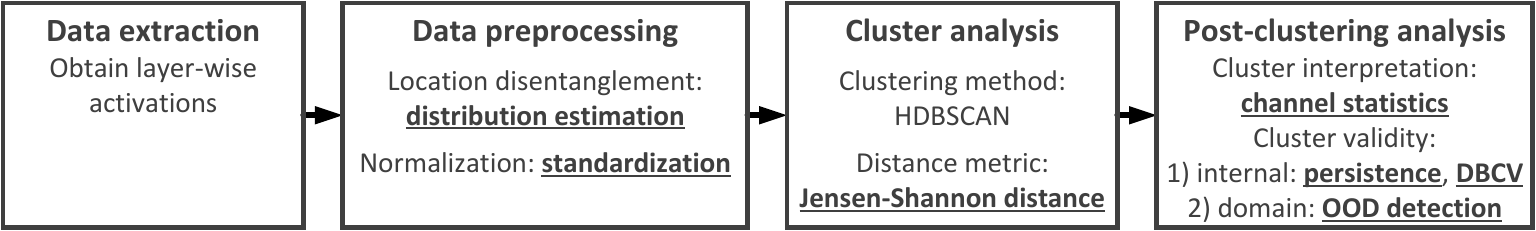}
\caption{Discovering layer-level concepts by applying NAP methodology with our proposed improvements (\underline{underlined}): extracting, preprocessing, and clustering activations---followed by interpreting and evaluating the clustering results.}
\label{fig:nap_proposed_block_diagram}
\end{figure*}

\textbf{Proposed Data Extraction:} Follows the same procedure as in the baseline NAP methodology, where we extract activation outputs from a specified layer $l$ of a neural network $n$ for a given input set $X = \{x_i \mid i \in [1, \ldots, N]\}$. This yields a set of activations  $A^l = \{a^l_i = n^l(x_i)\}$, where each $a^l_i$ contains the output feature maps from layer $l$. In convolutional layers, these are structured as per-channel activation matrices, used in subsequent processing.

\textbf{Proposed Normalization:} We rescale the activations of each convolutional channel to mean 0 and standard deviation 1 (known as \textit{Z-Score Normalization}), making them more suitable for subsequent distribution estimation by removing scale differences across channels and centering the data:

\begin{equation}
\label{eq:zscore}
    \phantom{,}\hat{a}^l_{i,u} = \frac{a^l_{i,u} - \mu^l_u}{\sigma^l_u},
\end{equation}
where $\hat{a}^l_{i,u}$ represents the normalized activation, $a^l_{i,u}$ is the original activation, $\mu^l_u$ is the mean, and $\sigma^l_u$ is the standard deviation of the activation output of channel $u$ in layer $l$ across all data samples in $X$.

\textbf{Proposed Location Disentanglement:} We employ a form of non-parametric density estimation technique, known as \textit{Kernel Density Estimator} (KDE; \cite{rosenblatt1956,parzen1962}), to aggregate the normalized activations per convolutional channel. For KDE, the estimated \textit{Probability Density Function} (PDF) for channel $u$ in layer $l$ is given by
\begin{equation}
\label{eq:kde2}
    \phantom{,}\hat{f}^{l}_{i,u}(a) = \frac{1}{Q H} \sum_{q=1}^{Q} K\left(\frac{a - \hat{a}^l_{i,u,q}}{H}\right),
\end{equation}
where $K()$ is the \textit{kernel function} (e.g., Gaussian), $H$ is a smoothing parameter called the \textit{bandwidth}, and $Q$ is the number of activation values on channel $u$ in layer $l$. We use \textit{Scott's Rule} (\cite{scott1992multivariate}) for selecting the bandwidth, providing a standard bias-variance tradeoff:
\begin{equation}
\label{eq:scottsrule2}
    \phantom{.}H = 1/\sqrt[5]{Q}.
\end{equation}
We sample the estimated PDF (Equation~\ref{eq:kde2}) over a Z-score range covering $P\%$ of the normalized activation mass to reduce outlier influence:
\begin{equation}
\label{eq:activation_range}
    \phantom{,}z_{\text{min}} = \Phi^{-1} \left(\frac{1 - \tfrac{P}{100}}{2} \right), \quad z_{\text{max}} = \Phi^{-1} \left(1 - \frac{1 - \tfrac{P}{100}}{2} \right),
\end{equation}
where $\Phi^{-1}(\cdot)$ is the \textit{Inverse Cumulative Distribution Function} (ICDF) of the standard normal distribution. The range is then discretized into $R$ uniformly spaced points for channel $u$ in layer $l$ to produce a fixed-length representation, as follows:
\begin{equation}
\label{eq:kde_sampled}
    \phantom{.}\mathbf{\hat{f}}^{l}_{i,u} = \left[ \hat{f}^{l}_{i,u}\left( z_{\text{min}} + r \cdot \frac{z_{\text{max}} - z_{\text{min}}}{R - 1}\right) \right]_{r=0}^{R-1}.
\end{equation}

\textbf{Proposed Clustering Analysis:} We cluster the sampled and concatenated estimated density values
\begin{equation}
\label{eq:kde_sampled_stacked}
    \phantom{,}\mathbf{\hat{g}}^{l}_{i} = \left[\hat{f}^{l}_{i,1,1}, \hat{f}^{l}_{i,1,2}, \dots, \hat{f}^{l}_{i,1,R}, \hat{f}^{l}_{i,2,1}, \hat{f}^{l}_{i,2,2}, \dots, \hat{f}^{l}_{i,U,R}, \right]_{u=1}^{U},
\end{equation}
where $U$ is the number of channels in layer $l$, using HDBSCAN* with the \textit{Jensen-Shannon distance metric} (\cite{lin1991jsd,endres2003jsmetric}), which is symmetric, bounded, and well-suited for comparing probability distributions such as those estimated by KDE in the previous step. The Jensen-Shannon distance $D^{l}_{\text{JS}}$ between activations for inputs $x_i$ and $x_j$ for all channels in layer $l$ is
\begin{align}
\label{eq:jsd_all_channels}
    &D^{l}_{\text{JS}}(x_i,x_j) = \nonumber \\
    &\frac{1}{\sqrt{2}}\sqrt{\sum^{U \cdot R}_{k = 1}{\left(\mathbf{\hat{g}}^{l}_{i,k} \log \frac{2 \cdot \mathbf{\hat{g}}^{l}_{i,k}}{\mathbf{\hat{g}}^{l}_{i,k} + \mathbf{\hat{g}}^{l}_{j,k}} + \mathbf{\hat{g}}^{l}_{j,k} \log \frac{2 \cdot \mathbf{\hat{g}}^{l}_{j,k}}{\mathbf{\hat{g}}^{l}_{i,k} + \mathbf{\hat{g}}^{l}_{j,k}} \right)}}
\end{align}
and we use this as a distance metric in clustering.

\textbf{Proposed Post-Clustering Analysis:} We evaluate clustering quality using two internal metrics and one application-specific metric. As internal metrics, we use the \textit{persistence score} (\cite{campello2013hdbscan}), which reflects the stability or robustness of a cluster (with higher values indicating more well-defined clusters), and the \textit{Density-Based Clustering Validation (DBCV) index} (\cite{moulavi2014density}), which is designed to assess density-based clustering algorithms like HDBSCAN by balancing cluster compactness and separation. As an application-specific metric, we use \textit{Out-of-Distribution (OOD) detection} (see, e.g. \cite{zhu2025robust}), assuming that OOD samples fall outside the learned cluster structure. This allows clustering to act as a proxy for distributional mismatch. In this setting clustering serves not just as an explanation method but as a model debugging tool---similar to how \cite{adebayo2020debugging} frame OOD behavior as test-time contamination. To interpret the resulting clusters, we use the channel parameters of the inputs (e.g., SNR) to summarize and explain structure.

\subsection{Implementation Details}
\label{supplemental:nap_implementation_details}
Although \cite{bauerle2022nap} mention the availability of their NAP methodology as a Python package, we were unable to locate a public implementation. Consequently, we reimplemented both their baseline approach (Supplemental~\ref{supplemental:baseline_nap_methodology}) and our proposed enhancements (Supplemental~\ref{supplemental:proposed_nap_methodology}) using various Python libraries from which the most essential can be found from Listing~\ref{lst:compute_software_requirements}.

In our methodology, the density is estimated using a Kernel Density Estimator (KDE) with a Gaussian kernel $K$. To obtain a compact yet informative representation, the estimated probability density function $\hat{f}^{l}_{i,u}$ (Equation~\ref{eq:kde}) is sampled at $R = 10$ evenly spaced Z-scores (Equation~\ref{eq:kde_sampled}) over the interval $[z_{\text{min}}, z_{\text{max}}] = [-2.576, 2.576]$, which contains $P = 99\%$ of the mass of a standard normal distribution (Equation~\ref{eq:activation_range}). This one-point-per-decile grid (step $\Delta \approx 0.515$) yields a ten-dimensional feature vector per unit---an order-of-magnitude increase in information over a single-mean summary in baseline methodology, while remaining compact enough for downstream modeling or tabular presentation.

\begin{lstlisting}[caption={Most essential Python requirements used in \texttt{requirements.txt} format.}, label={lst:compute_software_requirements}]
    h5py==3.13.0
    hdbscan==0.8.40
    joblib==1.4.2
    matplotlib==3.10.1
    numba==0.61.0
    numpy==2.1.3
    pandas==2.2.3
    scikit-learn==1.6.1
    scipy==1.15.2
    sionna==0.10.0
    torch==2.6.0
    torchvision==0.21.0
    umap-learn==0.5.7
    tensorflow==2.6.4
\end{lstlisting}

\subsection{Experiments on Visual Object Recognition}
\label{supplemental:nap_analysis_visual}
We applied the NAP methodology using the average amount of activations as the baseline, as suggested by \cite{bauerle2022nap} to be the most suitable choice for extracting a larger number of NAPs. We then compared this baseline with our proposed methodology---outlined in Figure~\ref{fig:nap_proposed_block_diagram} and detailed in Supplemental~\ref{supplemental:proposed_nap_methodology}---and additionally studied the effect of HDBSCAN parameters.

\subsubsection{Experimental Setup}
\label{supplemental:nap_analysis_experimentalsetup_visual}
We employed a ResNet-50~\cite{he2016resnet} convolutional neural network pre-trained on the ImageNet-1K dataset---namely IMAGENET1K\_V2 weights from TorchVision~\cite{marcel2010torchvision}---to perform visual object classification. To align with the experimental protocol of \cite{bauerle2022nap}, we used a randomly selected 1\% subset (12,832 images) of the ImageNet ILSVRC2012 training set (\cite{russakovsky2015imagenet}). As part of the setup---and following their methodology---we extracted feature activations from the Conv4 and Conv5 stages of the network---corresponding to the \texttt{layer3} and \texttt{layer4} modules in TorchVision’s implementation---which produce intermediate activations of size [1024 × 14 × 14] and [2048 × 7 × 7], respectively, when processing 224 × 224 resolution images.

We used our own PyTorch-based implementation of both the NAP baseline and our proposed methodology; implementation details are provided in Supplemental~\ref{supplemental:nap_implementation_details}.

\subsubsection{Results}
\label{supplemental:nap_analysis_results_visual}
\begin{figure}[h]
\centering
\includegraphics[width=\columnwidth]{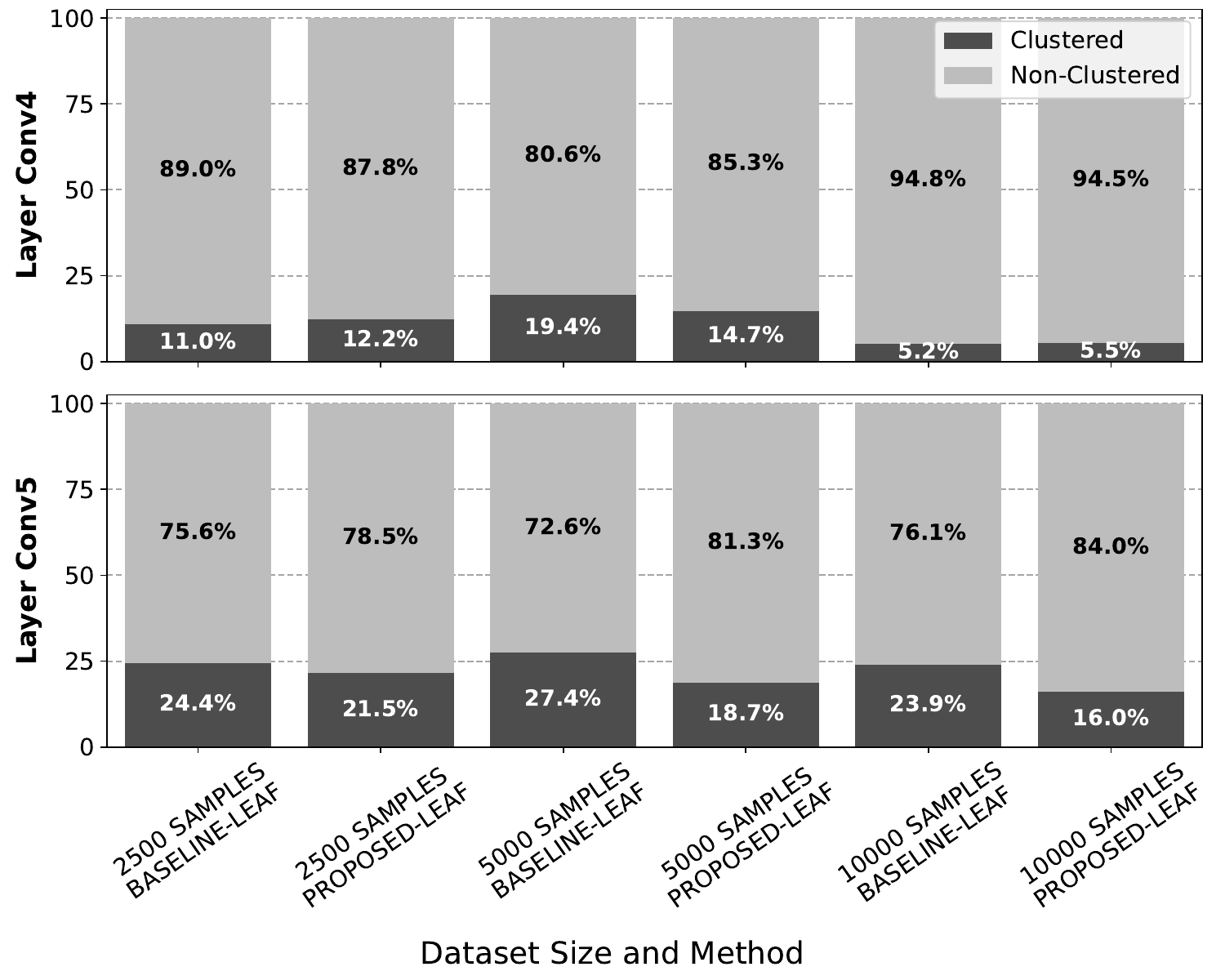}
\caption{Percentage of clustered and non-clustered samples as a function of sample size and aggregation method with minimum cluster size 5 using ImageNet subset with ResNet-50 model.}
\label{fig:imagenet_clustered_non_clustered_vs_sample_size}
\end{figure}

We first evaluated the impact of dataset size using the HDBSCAN parameters recommended by \cite{bauerle2022nap}---specifically, the Leaves of Tree (Leaf) as the cluster selection criteria with a minimum cluster size of 5. As shown in Figure~\ref{fig:imagenet_clustered_non_clustered_vs_sample_size}, the baseline method resulted in 73–95\% of samples remaining unclustered. Our proposed methodology did not mitigate this issue. A high proportion of non-clustered activations is problematic, as it hinders the discovery of meaningful layer-level concepts, reducing interpretability. It also undermines confidence in the clustering structure and impairs downstream tasks such as Out-of-Distribution (OOD) detection, which depend on well-formed, dense regions in activation space.

\begin{figure}[h]
\centering
\includegraphics[width=\columnwidth]{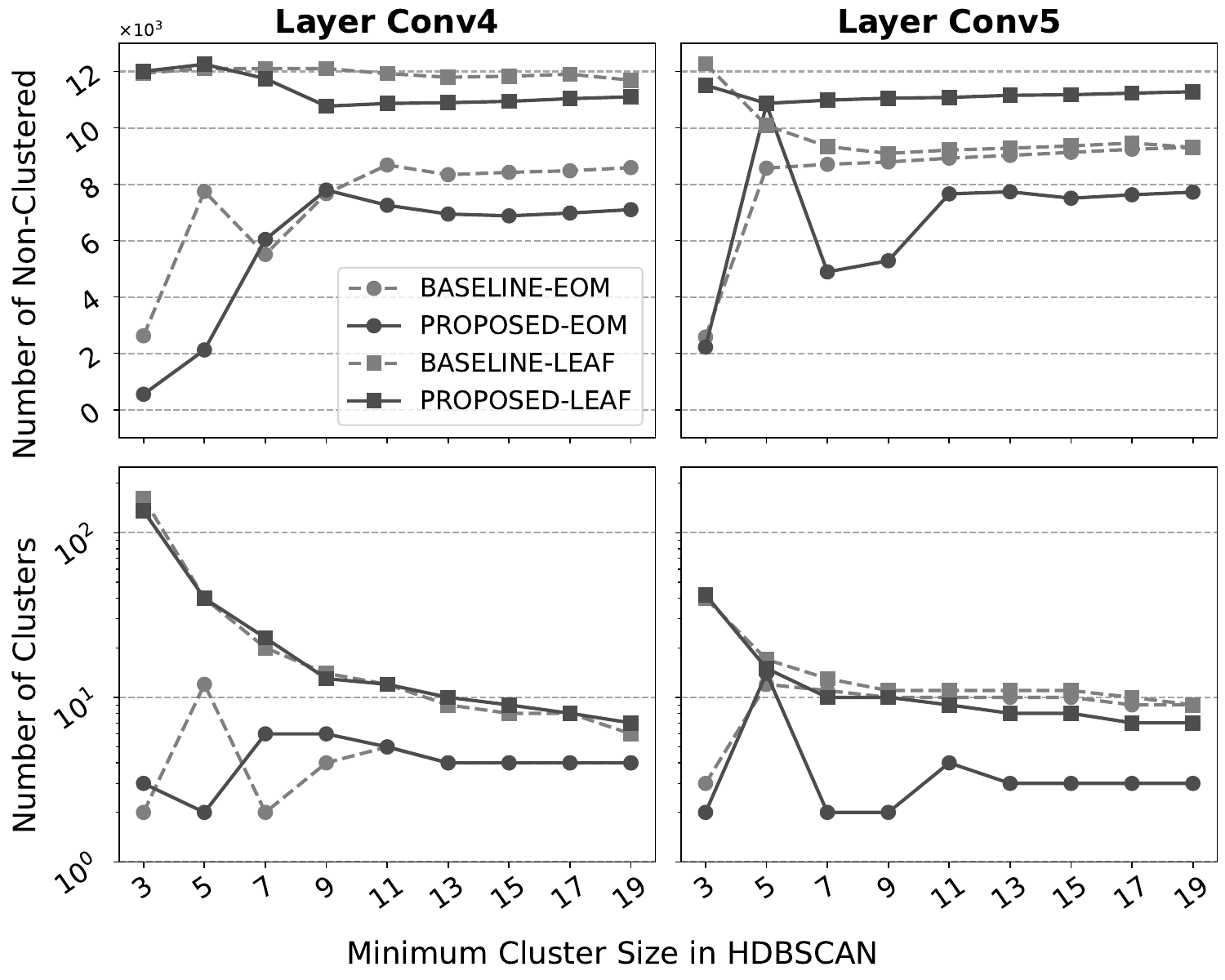}
\caption{Number of non-clustered samples and clusters by aggregation and cluster selection method as function of minimum cluster size using ImageNet subset with ResNet-50 model.}
\label{fig:imagenet_hdbscan_parameters}
\end{figure}

\begin{figure}[h]
\centering
\includegraphics[width=\columnwidth]{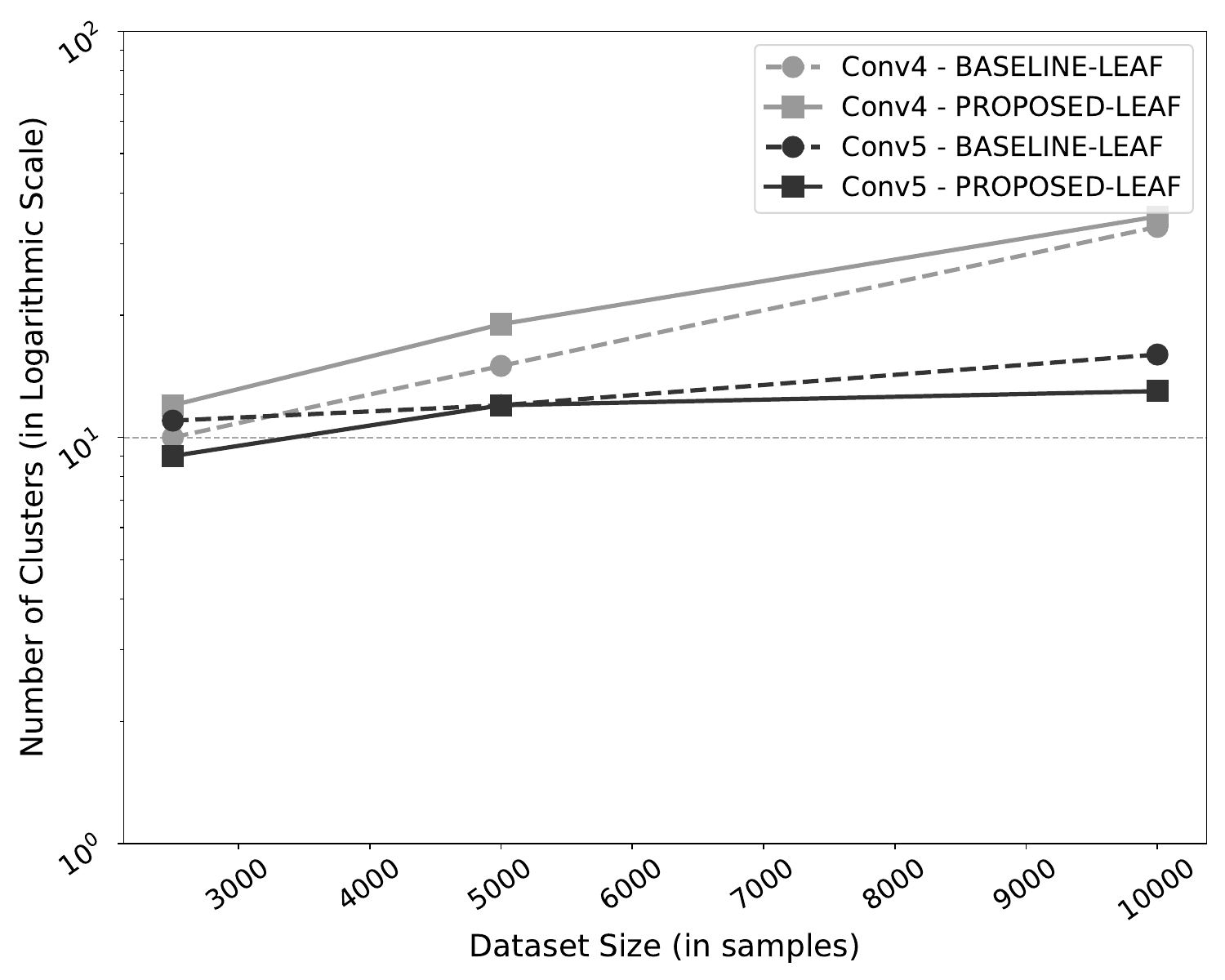}
\caption{Number of clusters per layer and methodology as function of sample size with minimum cluster size 5 using ImageNet subset with ResNet-50 model.}
\label{fig:imagenet_clusters_vs_sample_size}
\end{figure}

\begin{figure}[h]
\centering
\includegraphics[width=\columnwidth]{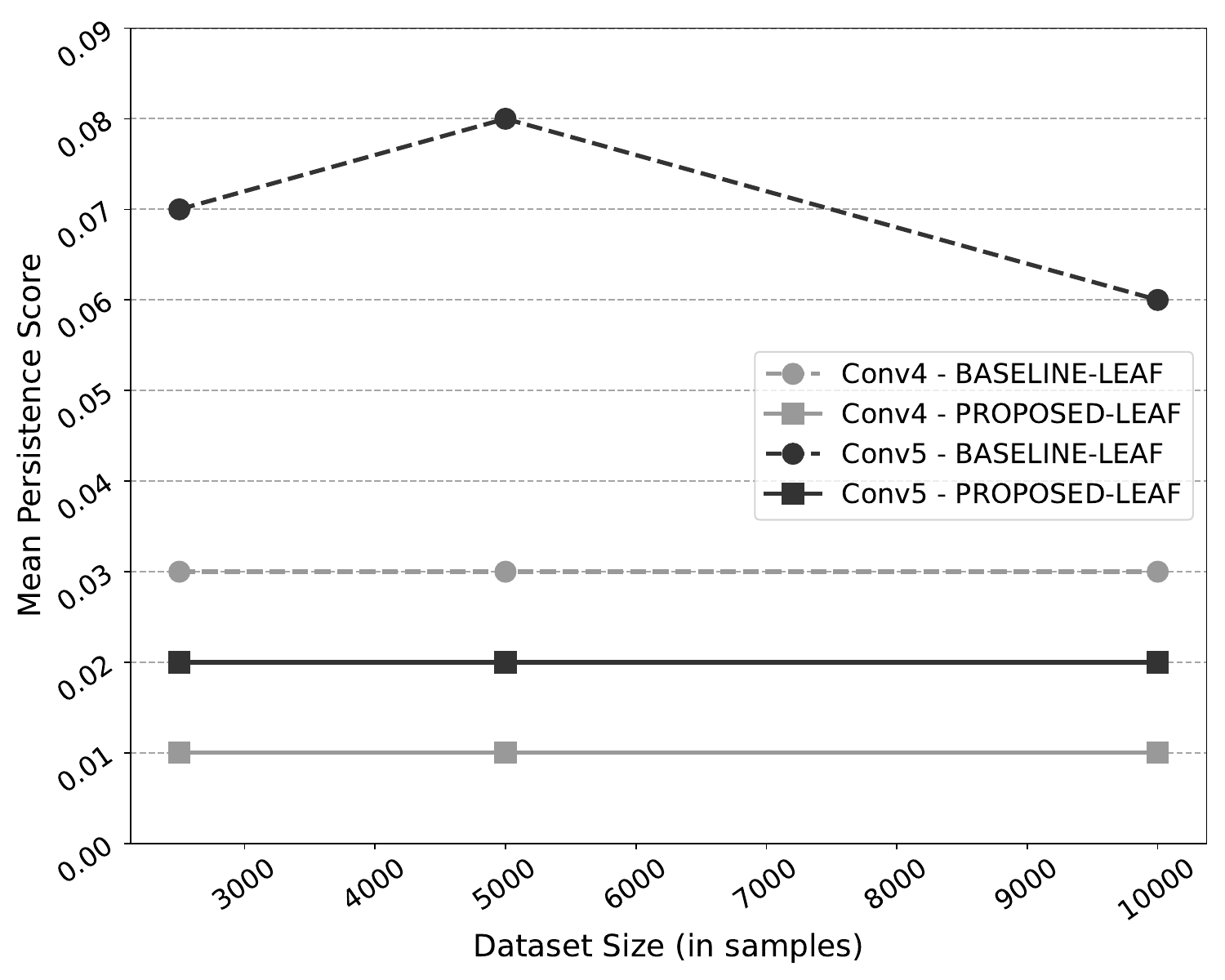}
\caption{Mean persistence score per layer and methodology as function of sample size with minimum cluster size 5 using ImageNet subset with ResNet-50 model.}
\label{fig:imagenet_mean_persistence_vs_sample_size}
\end{figure}

Notably, \cite{bauerle2022nap} do not report any analysis of unclustered or outlier samples, which we found to be a recurring issue in the visual object recognition setting as well. Our results show that increasing dataset size leads to more clusters being discovered—consistent with findings in \cite{bauerle2022nap}---but the mean persistence score remains relatively stable.

Next, we investigated the effect of HDBSCAN parameters on clustering outcomes. Figure~\ref{fig:imagenet_hdbscan_parameters} shows that using Excess of Mass (EoM) as the cluster selection criteria consistently reduces the number of unclustered samples across settings. Smaller minimum cluster sizes (e.g., 3) also lead to fewer unclustered samples. With the updated cluster selection criteria (EoM), our methodology generally clusters more samples---except for Conv5 with a minimum cluster size of 5. The number of clusters behaves differently depending on the cluster selection method: with Leaf, the number of clusters decreases as the minimum cluster size increases, likely due to aggressive pruning of small clusters; with EoM, the number remains more stable, as it prioritizes persistence over strict pruning.

We explore the impact of dataset size on visual object recognition using the ResNet50 architecture with pretrained weights and activations from the Conv4 and Conv5 stages of the network, as described in Supplemental~\ref{supplemental:nap_analysis_visual}. Figures \ref{fig:imagenet_clusters_vs_sample_size} and \ref{fig:imagenet_mean_persistence_vs_sample_size} illustrate how varying the dataset size influences the number of clusters and the mean persistence score across different layers and methodologies.

\subsection{Experiments on Radio Receiver}
\label{supplemental:nap_analysis_radio}
This section provides supplemental results for the radio receiver experiments. The experimental setup follows the main paper (Section~\ref{sec:experimentalsetup_cluster_analysis}). The central results are reported in the main paper (Section~\ref{sec:results_cluster_analysis}), while additional analyses are presented here.

\subsubsection{Impact of Dataset Size}
\begin{figure}[h!]
\centering
\includegraphics[width=\columnwidth]{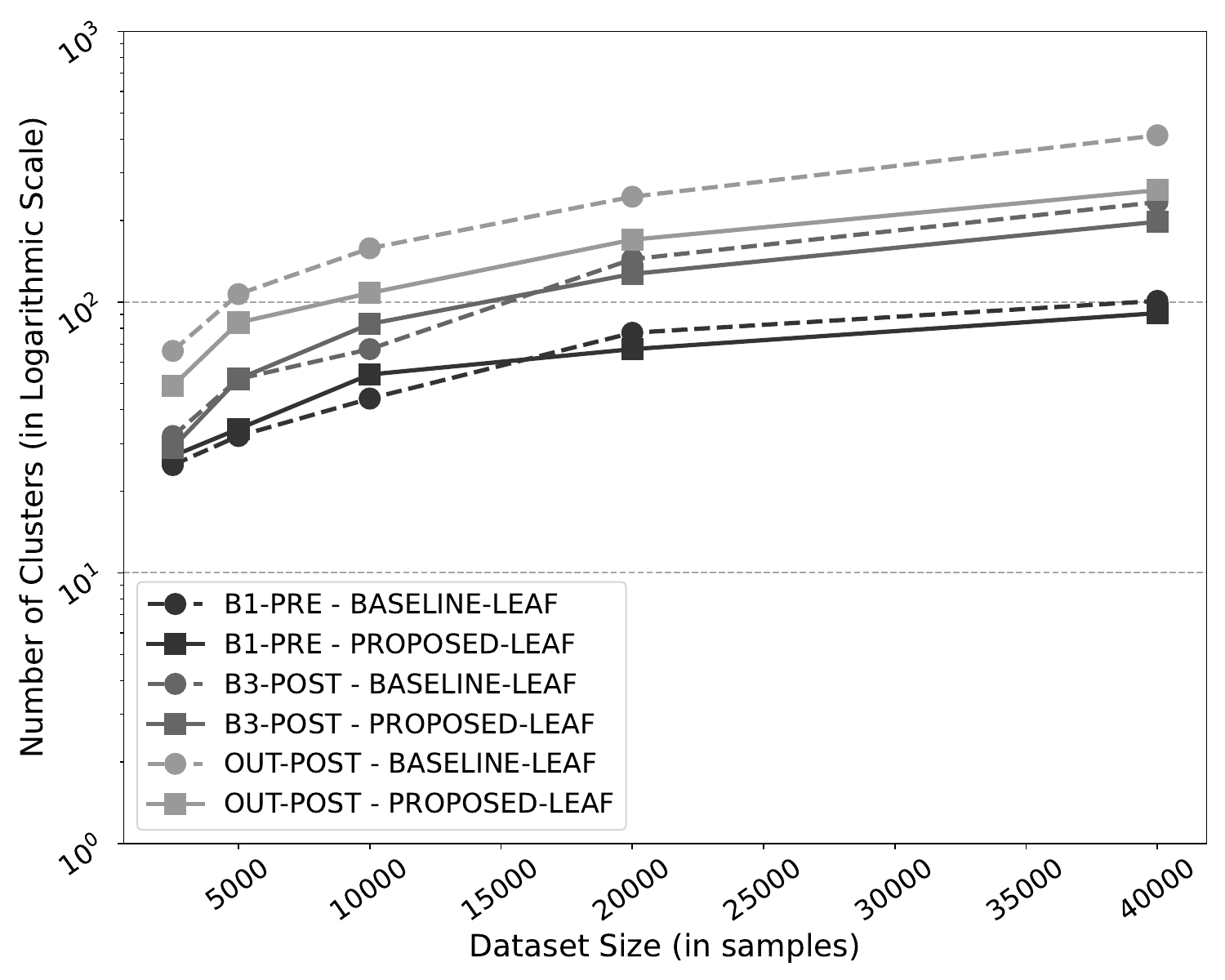}
\caption{Number of clusters per layer and methodology as function of sample size with minimum cluster size 5 using link-level simulations with DeepRx model.}
\label{fig:deeprx_clusters_vs_sample_size}
\end{figure}

\begin{figure}[h!]
\centering
\includegraphics[width=\columnwidth]{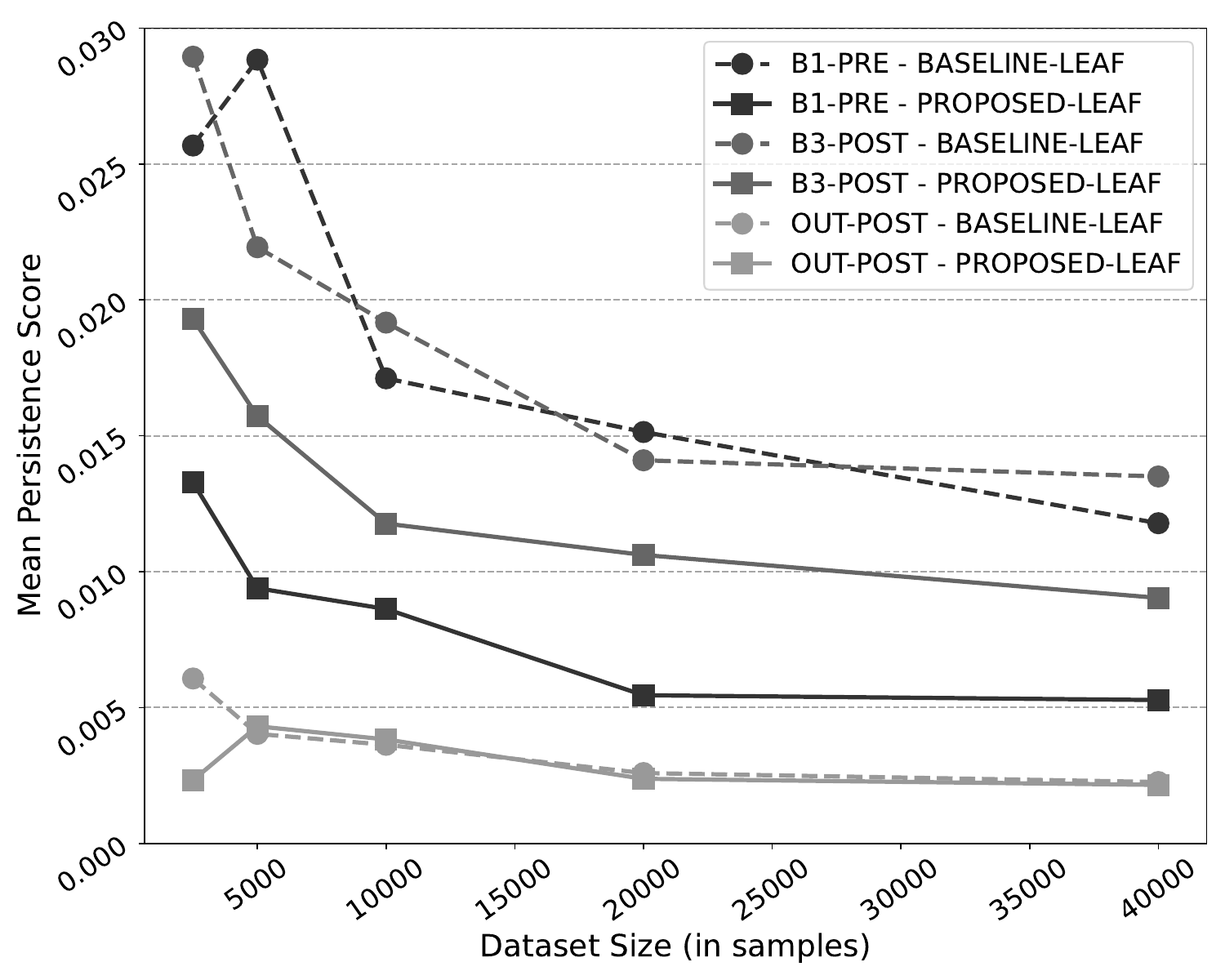}
\caption{Mean persistence score per layer and methodology as function of sample size with minimum cluster size 5 using link-level simulations with DeepRx model.}
\label{fig:deeprx_mean_persistence_vs_sample_size}
\end{figure}

We study the effect of dataset size on clustering behavior in the DeepRx architecture, using activations from multiple network stages as described in Supplemental~\ref{supplemental:radio_receiver_model}. Figures~\ref{fig:deeprx_clusters_vs_sample_size} and~\ref{fig:deeprx_mean_persistence_vs_sample_size} show how varying the sample size affects the number of clusters and the mean persistence score across layers and methods. 

\subsubsection{Cluster Statistics and DBCV Scores}
\label{supplemental:example_case_cluster_statistics}
This section reports detailed cluster statistics and DBCV scores for the example shown in Fig.~\ref{fig:data_structure}. Baseline results are summarized in Tables~\ref{tab:nap_baseline_cluster_stats} and~\ref{tab:nap_dbcv_scores}, while the corresponding results for the proposed method are given in Tables~\ref{tab:nap_proposed_cluster_stats} and~\ref{tab:nap_dbcv_scores}. The consistently low DBCV values indicate weak cluster separability for both methods, supporting the interpretation of a continuous activation manifold rather than discrete clusters.

\input{supplemental/tables/nap/baseline_cluster_stats}
\input{supplemental/tables/nap/proposed_cluster_stats}
\input{supplemental/tables/nap/dbcv_scores}

%% file: supplemental/tables/nap/baseline_cluster_stats.tex
\begin{table*}[h!]
\centering
\caption{Cluster statistics for the \textbf{baseline} methodology in the example shown in Fig.~\ref{fig:data_structure}. Values are mean $\pm$ standard deviation.}
\label{tab:nap_baseline_cluster_stats}
\begin{tabular}{crrrrrrr}
\toprule
Cluster & Size & Persist. & SNR [dB] & UE speed [km/h] & Delay [ns] & SER & Dominant CDL \\
\midrule
0 & 7     & 0.01 & $7.6 \pm 12.0$ & $71.8 \pm 31.6$ & $112 \pm 72$ & $0.33 \pm 0.34$ & A (100\%) \\
1 & 38373 & 0.12 & $7.5 \pm 10.1$ & $62.1 \pm 36.0$ & $156 \pm 84$ & $0.30 \pm 0.31$ & A/E (49/51) \\
2 & 6     & 0.00 & $10.1 \pm 9.2$ & $58.0 \pm 24.5$ & $156 \pm 89$ & $0.26 \pm 0.23$ & A (100\%) \\
\bottomrule
\end{tabular}
\end{table*}

%% file: supplemental/tables/nap/proposed_cluster_stats.tex
\begin{table*}[h!]
\centering
\caption{Cluster statistics for the \textbf{proposed} methodology in the example shown in Fig.~\ref{fig:data_structure}. Values are mean $\pm$ standard deviation.}
\label{tab:nap_proposed_cluster_stats}
\begin{tabular}{crrrrrrr}
\toprule
Cluster & Size & Persist. & SNR [dB] & UE speed [km/h] & Delay [ns] & SER & Dominant CDL \\
\midrule
0 & 14    & 0.00 & $6.5 \pm 11.3$ & $47.2 \pm 32.2$ & $145 \pm 75$ & $0.32 \pm 0.32$ & A/E (57/43) \\
1 & 5     & 0.00 & $10.0 \pm 10.0$ & $90.2 \pm 25.2$ & $222 \pm 59$ & $0.26 \pm 0.32$ & E (60\%) \\
2 & 5     & 0.00 & $4.5 \pm 9.6$   & $63.3 \pm 17.0$ & $198 \pm 31$ & $0.38 \pm 0.31$ & A (60\%) \\
3 & 6     & 0.00 & $10.4 \pm 10.4$ & $68.9 \pm 43.6$ & $119 \pm 80$ & $0.27 \pm 0.30$ & E (67\%) \\
4 & 10    & 0.00 & $11.2 \pm 8.9$  & $76.1 \pm 37.5$ & $157 \pm 72$ & $0.19 \pm 0.29$ & E (60\%) \\
5 & 6     & 0.01 & $8.9 \pm 7.4$   & $42.4 \pm 29.4$ & $186 \pm 98$ & $0.23 \pm 0.24$ & A (67\%) \\
6 & 33063 & 0.07 & $7.5 \pm 10.1$  & $62.0 \pm 36.0$ & $155 \pm 84$ & $0.30 \pm 0.31$ & A/E (50/50) \\
7 & 6     & 0.00 & $4.1 \pm 10.5$  & $55.4 \pm 27.4$ & $139 \pm 70$ & $0.42 \pm 0.32$ & E (67\%) \\
\bottomrule
\end{tabular}
\end{table*}

%% file: supplemental/tables/nap/dbcv_scores.tex
\begin{table*}[h!]
\centering
\caption{DBCV clustering quality scores for the example in Fig.~\ref{fig:data_structure}.}
\label{tab:nap_dbcv_scores}
\begin{tabular}{lccc}
\toprule
Method & Overall DBCV & Best cluster & Worst cluster \\
\midrule
Baseline & 0.055 & 0.259 & 0.058 \\
Proposed & 0.014 & 0.187 & 0.017 \\
\bottomrule
\end{tabular}
\end{table*}

%% file: supplemental/deeprx.tex
\section{Radio Receiver Model and Data}
\label{supplemental:radio_receiver_model}
\begin{figure*}[ht]
    \includegraphics[width=\textwidth]{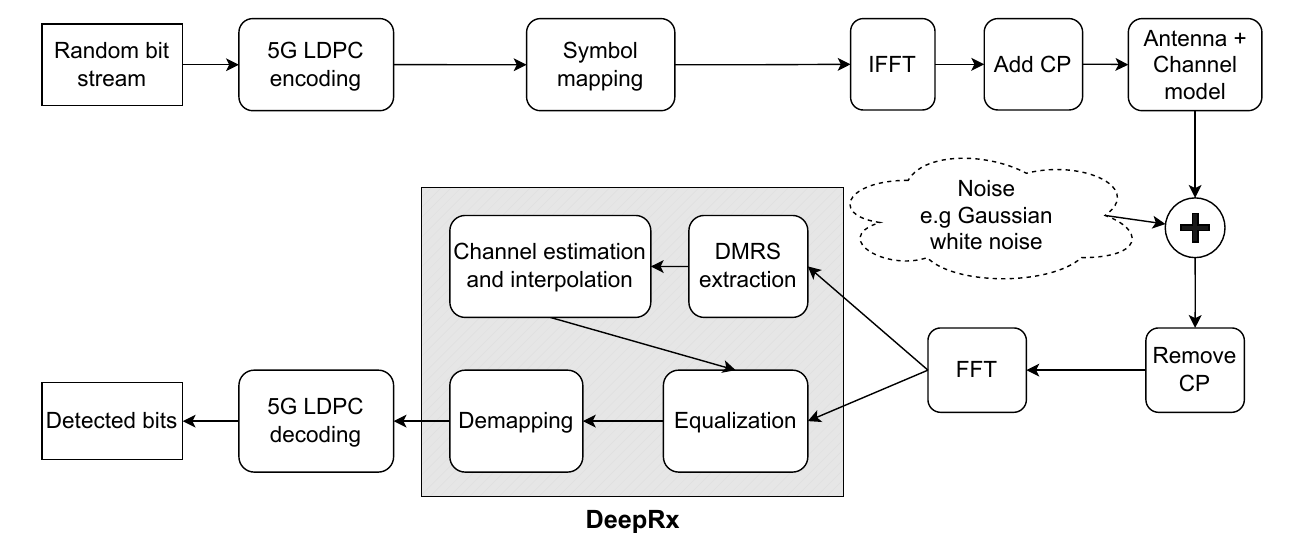}
    \centering
    \caption{Illustration of the PUSCH simulator (adapted from \protect\cite{honkala2021deeprx}), where DeepRx is used to replace channel estimation, equalization, and demapping blocks.}
    \label{fig:DeepRx}
\end{figure*}

DeepRx (\cite{honkala2021deeprx}) is a fully convolutional deep learning model that jointly performs the tasks of channel estimation, equalization, and demapping, as illustrated in Figure \ref{fig:DeepRx}. Moreover, the model is designed to be compliant with 5G NR specifications, capable of estimating the log-likelihood ratio (LLR), or the uncertainty, of each received bit so that it can be fed into an LDPC decoder. The model supports different demodulation reference signal (DMRS) or pilot configurations included in the 5G NR specifications (\cite{honkala2021deeprx}).

\subsection{Data Simulation} \label{sect:data_simulation}
The data are randomly generated using a 5G-compliant physical uplink shared channel (PUSCH) simulator, powered by NVIDIA Sionna (\cite{hoydis2023sionna}), and configured with the parameters listed in Table \ref{tab:simulation_parameters}. The simulation process is presented in Figure \ref{fig:DeepRx} and consists of three main stages:

\begin{itemize}
    \item \textbf{Transmitter}:
    \begin{enumerate}
        \item Generate random bits.
        \item Encode the bits using a 5G-compliant LDPC encoder.
        \item Map the encoded bits to symbols and distribute them across available physical resource blocks (PRBs) within the transmission time interval (TTI).
        \item Insert pilot symbols, or DMRS, into the specified subcarriers.
        \item Convert the PRBs to an OFDM waveform with 14 OFDM symbols per TTI by performing an inverse Fourier transform (IFFT).
        \item Add a CP to the beginning of each OFDM symbol to mitigate inter-symbol interference.
    \end{enumerate}
    \item \textbf{Channel}:
    \begin{enumerate}
        \item Pass the waveform through a channel model.
        \item Randomly select the maximum Doppler shift and root mean square (RMS) delay spread for each channel realization.
        \item Add noise to the signal.
    \end{enumerate}
    \item \textbf{Receiver}:
    \begin{enumerate}
        \item Receive the signal.
        \item Remove the CP.
        \item Perform FFT on each OFDM symbol to convert them into the frequency domain.
        \item Feed the frequency-domain signal of 14 OFDM symbols into the DeepRx model.
        \item Decode the LLR output using a 5G-compliant LDPC decoder.
    \end{enumerate}
\end{itemize}

\subsection{Model Input} 
To maximize the network's utilization of the input data, the model is provided with the received signal represented in the frequency domain over the entire transmission time interval (TTI). Denoting $S=14$ as the number of OFDM symbols, $F=192$ as the number of subcarriers, and $N_r=2$ as the number of receive antennas, the input consists of the following three arrays:

\begin{enumerate}
    \item Received signal after the FFT (data and received pilot symbols), denoted by $\mathbf{Y} \in \mathbb{C}^{S \times F \times N_r}$. 
    \item Pilot reference symbols within the received signal Y in both frequency and time, denoted by $\mathbf{X_p} \in \mathbb{C}^{S \times F \times 1}$.
    \item Raw channel estimate which is the element-wise product of the received signal $\mathbf{Y}$ and complex conjugate of the transmitted pilot symbols $\mathbf{X_p}$, denoted by $\mathbf{H_r} \in \mathbb{C}^{S \times F \times N_r}$.
\end{enumerate}

\subsection{Model Architecture} 
\begin{figure*}[h!]
    \includegraphics[width=\textwidth]{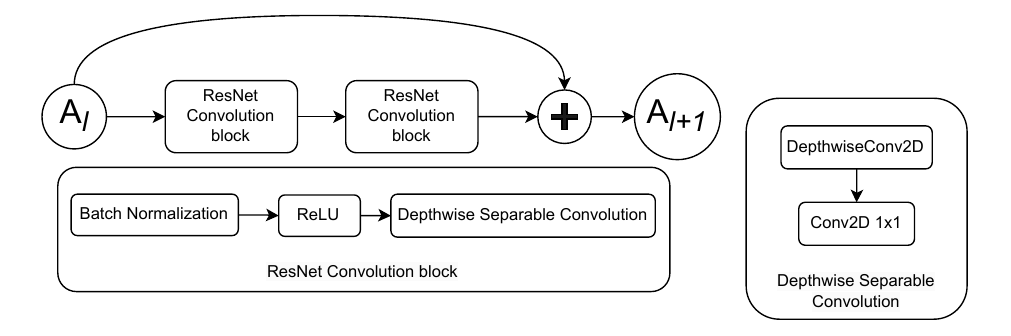}
    \centering
    \caption{DeepRx's pre-activation ResNet block (adapted from \protect\cite{honkala2021deeprx}).}
    \label{fig:deeprx_resnet}
\end{figure*}

\begin{figure}[h!]
    \includegraphics[width=0.95\columnwidth]{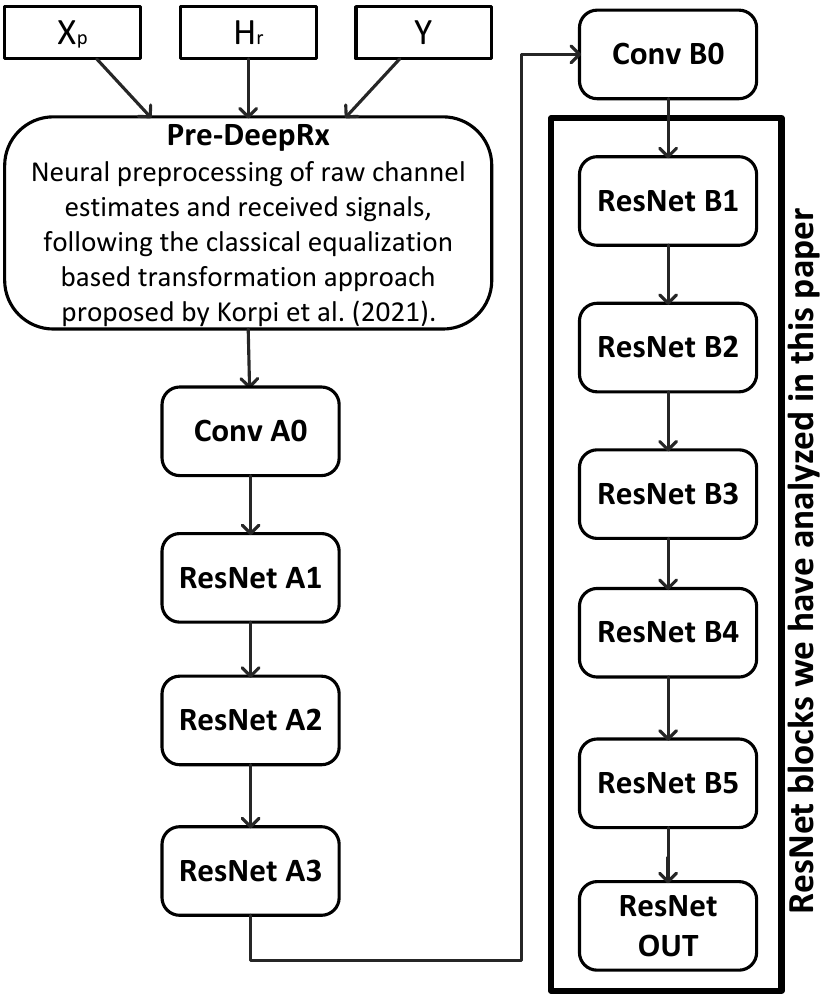}
    \centering
    \caption{Illustration of the data flow within the DeepRx model architecture.}
    \label{fig:deeprx_dataflow}
\end{figure}

The architecture of the DeepRx model, based on the work by \cite{honkala2021deeprx} and \cite{korpi2021}, is shown in Table \ref{tab:deeprx_resnet_architecture}. It consists of multiple pre-activation ResNet blocks---an extension of the ResNet architecture by \cite{he2016resnet}---and several convolutional layers. The pre-activation ResNet block, illustrated in Figure \ref{fig:deeprx_resnet}, is composed of two consecutive groups, each consisting of a batch normalization layer, a Rectified Linear Unit (ReLU) layer, and a depthwise separable convolution layer (\cite{Chollet2017}). In each depthwise separable convolution block, the depth multiplier value is set to 1, meaning the number of output channels in the depthwise convolution step is the same as the input. The data flow direction is depicted in Figure \ref{fig:deeprx_dataflow}. Certain architectural details have been omitted as they are not essential to the core contributions of our work.

The layers of interest in the paper are the ResNet blocks B1 through OUT, as illustrated in Figure~\ref{fig:deeprx_dataflow}. We focus particularly on the activations after their ReLU layers. Since each ResNet block contains two ReLU layers, we denote them as \textbf{PRE} and \textbf{POST} accordingly. For example, when referring to the second ReLU layer of ResNet block B3, we use the notation \textbf{B3-POST}.

\input{supplemental/tables/deeprx/deeprx_resnet_architecture}

\subsection{Model Output and Loss Function} \label{sect:deeprx_output_loss}
\begin{figure}[h]
    \centering
    \includegraphics[width=\columnwidth]{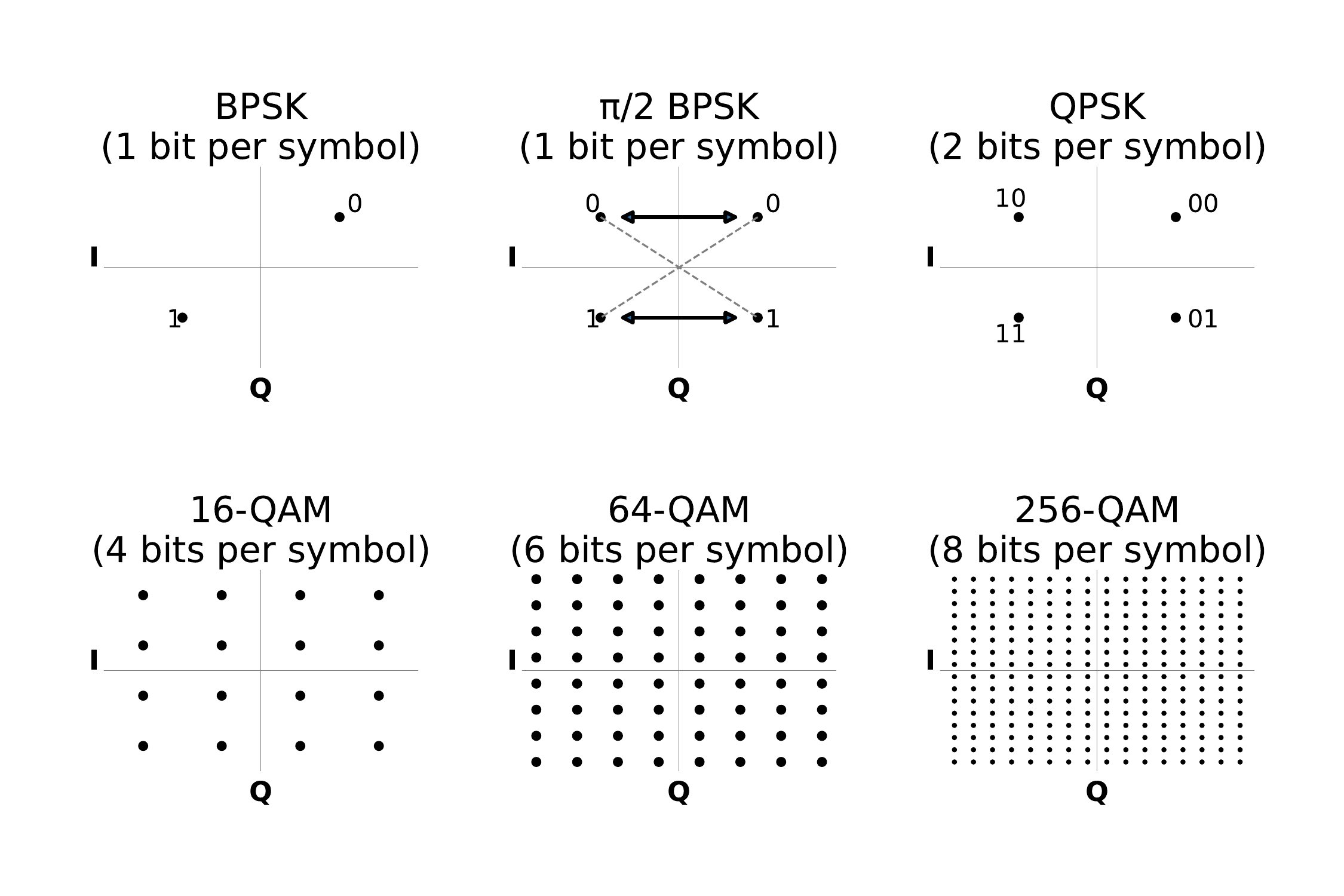}
    \caption{Modulation schemes used in 5G NR, including BPSK, Pi/2 BPSK, QPSK, 16-QAM, 64-QAM, and 256-QAM (redrawn from \protect\cite{Zaidi2018}).}
    \label{fig:modulation}
\end{figure}

Since DeepRx (\cite{honkala2021deeprx}) is designed to support multiple Quadrature Amplitude Modulation (QAM) schemes in 5G, including QPSK, 16-QAM, 64-QAM, and 256-QAM, which are hierarchically related (see Figure \ref{fig:modulation}).

\begin{figure}[h]
    \centering
    \includegraphics[width=\columnwidth]{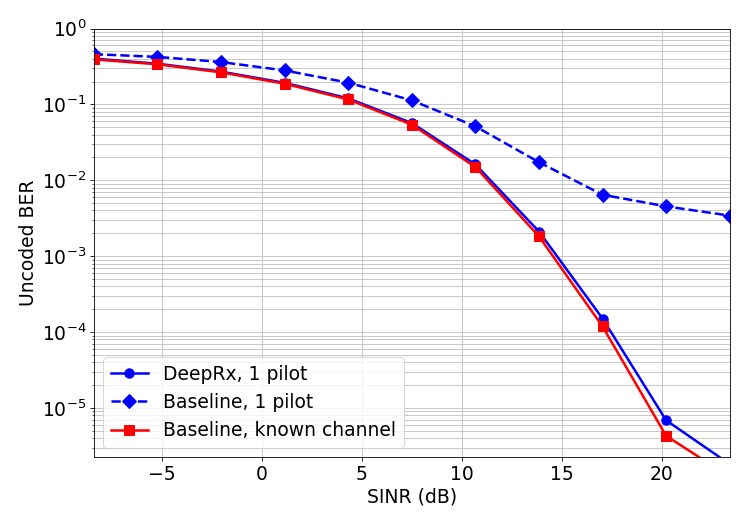}
    \caption{Performance comparison of the DeepRx model against baseline approaches across different SINR levels.}
    \label{fig:evaluation_curve}
\end{figure}

\begin{figure}[h]
    \centering
    \includegraphics[width=0.7\columnwidth]{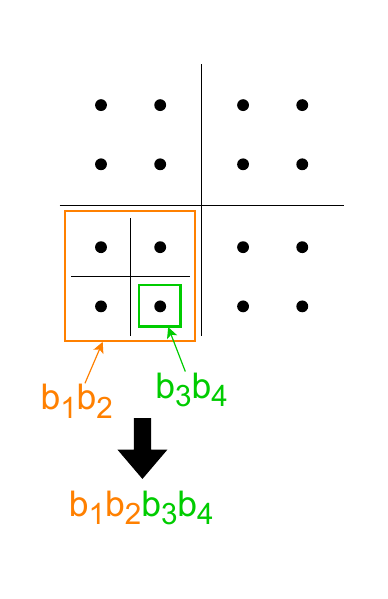}
    \caption{The relationship between 16-QAM and QPSK constellation points (redrawn from \protect\cite{honkala2021deeprx}). When QPSK is used, only 2 bits \protect\textcolor[HTML]{FF8000}{$b_1$} and \protect\textcolor[HTML]{FF8000}{$b_2$} are used to define the complex quadrant, but if 16-QAM is used, besides the first two bits, \protect\textcolor[HTML]{00CC00}{$b_3$} and \protect\textcolor[HTML]{00CC00}{$b_4$} are also used to define the quadrant within the quadrant.}
    \label{fig:qam_quadrant}
\end{figure}

The model output is designed so that the same output bits, or log-likelihood ratios (LLRs), correspond to the same part of the constellation space, regardless of the modulation scheme used. For example, for an individual symbol, the first two output bits describe the complex quadrant of the symbol, called the 1st-order point; the next two bits define the quadrant within the previously identified quadrant, called the 2nd-order point. The process continues until all the bits are utilized. The visualization of this hierarchical process is provided in Figure \ref{fig:qam_quadrant}.

\begin{figure}[h]
    \centering
    \includegraphics[width=\columnwidth]{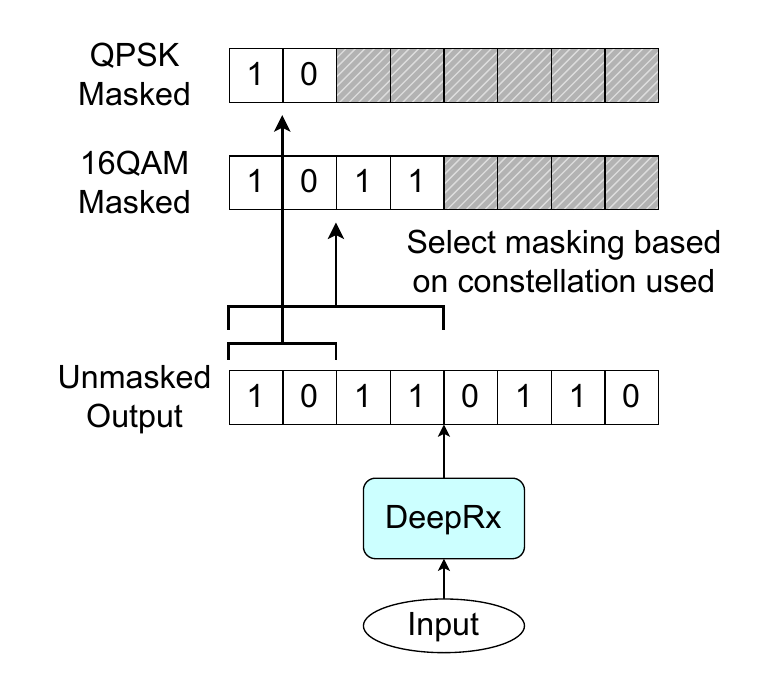}
    \caption{Masking bits based on the modulation, including both QPSK and 16QAM, where masking is selected based on the modulation. The output is masked accordingly so that unnecessary parts of the output array are not considered (redrawn from \protect\cite{honkala2021deeprx}).}
    \label{fig:deeprx_masking_modulation}
\end{figure}

In practice, the output length for each symbol is set according to the highest supported modulation order. Depending on the QAM scheme, the output is masked accordingly so that unnecessary parts of the output array are not considered. For example, if the highest modulation order is 256-QAM, each symbol in the model output is represented as an array with 8 elements for 8 bits. If a lower modulation order like 16-QAM, which requires 4 binary values, is used, the last 4 bits, or elements, are masked. The masking process is illustrated in Figure \ref{fig:deeprx_masking_modulation}.

Since the bit prediction can be framed as a binary classification problem, and the network output represents the LLR of the transmitted bits, the binary cross-entropy (CE) loss is used as the loss function for the DeepRx model:
\begin{align} \label{eqn:cross_entropy}
    \text{CE}(\bm{\theta}) =& -\frac{1}{\#DB} \sum_{(i,j) \in \mathcal{D}} \sum_{l=0}^{B-1} \nonumber \\
    &\left( b_{ijl} \log (\hat{b}_{ijl}) + (1 - b_{ijl}) \log (1 - \hat{b}_{ijl}) \right),
\end{align}
where \(\#\mathcal{D}\) is the number of resource elements carrying data; $B$ is the number of bits in the constellation used ($B=4$ for 16-QAM); \(b_{ijl}\) is the label bit; and \(\hat{b}_{ijl}\) is the probability that bit \(b_{ijl}\) equals one, calculated using the sigmoid function:
\begin{equation} \label{eqn:sigmoid}
    \hat{b}_{ijl} = \text{sigmoid} \left( L_{ijl} \right) = \frac{1}{1 + e^{-L_{ijl}}}.
\end{equation}

\subsection{Model's Predictive Performance} \label{sect:deeprx_performance}
The DeepRx model demonstrates competitive performance in terms of uncoded Bit Error Rate (BER) across varying Signal-to-Interference-plus-Noise-Ratio (SINR) levels, as seen in Figure \ref{fig:evaluation_curve}. It closely matches the baseline performance with a known channel, particularly at low SINR values. Notably, it outperforms the baseline with one pilot in higher SINR scenarios, which aligns with the findings reported in \cite{honkala2021deeprx}.

%% file: supplemental/tables/deeprx/deeprx_resnet_architecture.tex
\begin{table*}
    \centering
    \caption{The DeepRx CNN ResNet architecture, with all convolutional layers having a filter size of (3, 3). The ResNet block is illustrated in Figure~\ref{fig:deeprx_resnet}, and the overall data flow is depicted in Figure~\ref{fig:deeprx_dataflow}.}
    \label{tab:deeprx_resnet_architecture}
    \scriptsize
    \renewcommand{\arraystretch}{2}
    \newcolumntype{C}[1]{>{\centering\arraybackslash}p{#1}}
    \begin{tabularx}{\textwidth}{|X|X|C{1.1cm}|C{1.3cm}|c|}
        \hline
        \textbf{Layer}                          & \textbf{Type}                     & \textbf{Number of filters}  & \textbf{Dilation size (S, F)} & \textbf{Output Shape} \\ \hline
        Input 1 $\mathbf{X_p \in \mathbb{C}}$   & TX Pilot                          &                           &                           & $(14, 192, 1)$              \\ \hline
        Input 2 $\mathbf{H_r \in \mathbb{C}}$   & Raw channel estimate         	    &                           &                          	& $(14, 192, 2)$              \\ \hline
        Input 3 $\mathbf{Y \in \mathbb{C}}$	    & RX data                           &                          	&                           & $(14, 192, 2)$              \\ \hline
        Pre-DeepRx                              & Neural preprocessing              &                           &                           & $(14, 192, 20)$             \\ \hline
        Convolutive layer A0                    & 2D convolution                    & 40                        & (1,1)                     & $(14, 192, 40)$             \\ \hline
        ResNet block A1                         & Depthwise separable convolution   & 40              	        & (1,1)                    	& $(14, 192, 40)$             \\ \hline
        ResNet block A2                         & Depthwise separable convolution   & 40              	        & (2,3)                    	& $(14, 192, 40)$             \\ \hline
        ResNet block A3                         & Depthwise separable convolution   & 32                        & (3,6)                    	& $(14, 192, 32)$             \\ \hline
        Convolutive layer B0                    & 2D convolution                    & 64                  	    & (1,1)                    	& $(14, 192, 64)$             \\ \hline
        ResNet block B1                         & Depthwise separable convolution   & 64                  	    & (1,1)                    	& $(14, 192, 64)$             \\ \hline
        ResNet block B2                         & Depthwise separable convolution   & 64                  	    & (2,4)                    	& $(14, 192, 64)$             \\ \hline
        ResNet block B3                         & Depthwise separable convolution   & 32                  	    & (3,8)                    	& $(14, 192, 32)$             \\ \hline
        ResNet block B4                         & Depthwise separable convolution   & 32                  	    & (2,4)                    	& $(14, 192, 32)$             \\ \hline
        ResNet block B5                         & Depthwise separable convolution   & 32                  	    & (1,1)                    	& $(14, 192, 32)$             \\ \hline
        ResNet block OUT                        & Depthwise separable convolution   & 8                         & (1,1)                     & $(14, 192, 8)$              \\ \hline
        LLR Output $\mathbf{L}$, see Section \ref{sect:deeprx_output_loss} & Output &                           &                           & $(14, 192, 8)$              \\ \hline
    \end{tabularx}
\end{table*}

%% file: supplemental/nucpi.tex
\section{Neural Unit Channel Parameter Identification (NUCPI) Methodology for Mechanistic Interpretability}
\label{supplemental:nucpi_experiments_on_uespeed_and_delayspread}
\subsection{Introduction}
We present our replication of the experiments conducted by \cite{tuononen2025interpreting}, employing delay spread (see Appendix~\ref{appendix:nucpi_delay_spread}) and user equipment (UE) speed (see Appendix~\ref{appendix:nucpi_ue_speed}) as target parameters. This approach diverges from the original study, which utilized Signal-to-Noise Ratio (SNR) as the channel parameter of interest. Notably, \cite{tuononen2025interpreting} suggested that their method could be applicable to other channel parameters, which we explore here.

\subsection{Methodology}
We refer to the methodology proposed in \cite{tuononen2025interpreting} as \textit{Neural Unit Channel Parameter Identification (NUCPI)}, wherein an \textbf{explainer model} is trained to predict a given channel parameter based on intermediate activations from a \textbf{performer model}---in our case, the DeepRx model (see Supplemental~\ref{supplemental:radio_receiver_model}). We interpret the predictive performance of the explainer model---measured using Mean Squared Error (MSE), as suggested by \cite{tuononen2025interpreting}---as an indicator of how much information specific internal components (e.g., layers or channels) of the performer contain about the given channel parameter. This enables insights into the model’s internal behavior.

These insights could potentially be used to identify redundant or over-specialized components, guide model pruning \cite{han2016deepcompression}, or adapt architectures to better handle specific channel conditions \cite{cortes2017adanet}. While these applications are not experimentally explored in this work, the interpretability provided by NUCPI opens promising directions for future research in model compression and adaptive architecture design.

\begin{figure}[h]
\centering
\includegraphics[width=\columnwidth]{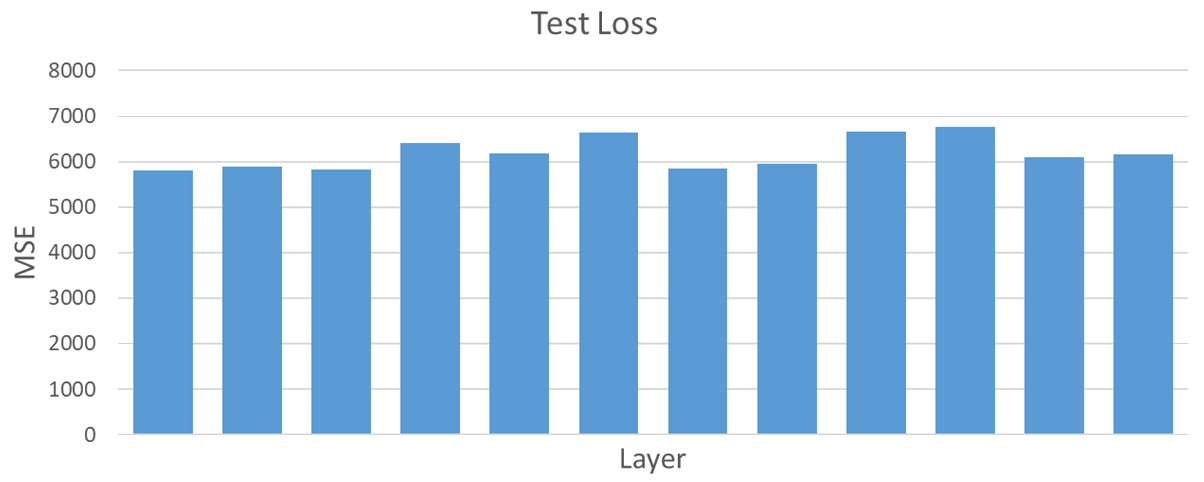}
\caption{Layerwise test losses for the explainer model on delay spread, showing poor predictive performance across all layers.}
\label{fig:nucpi_delay_spread_layerwise_test_losses}
\end{figure}

\begin{figure}[h]
\centering
\includegraphics[width=\columnwidth]{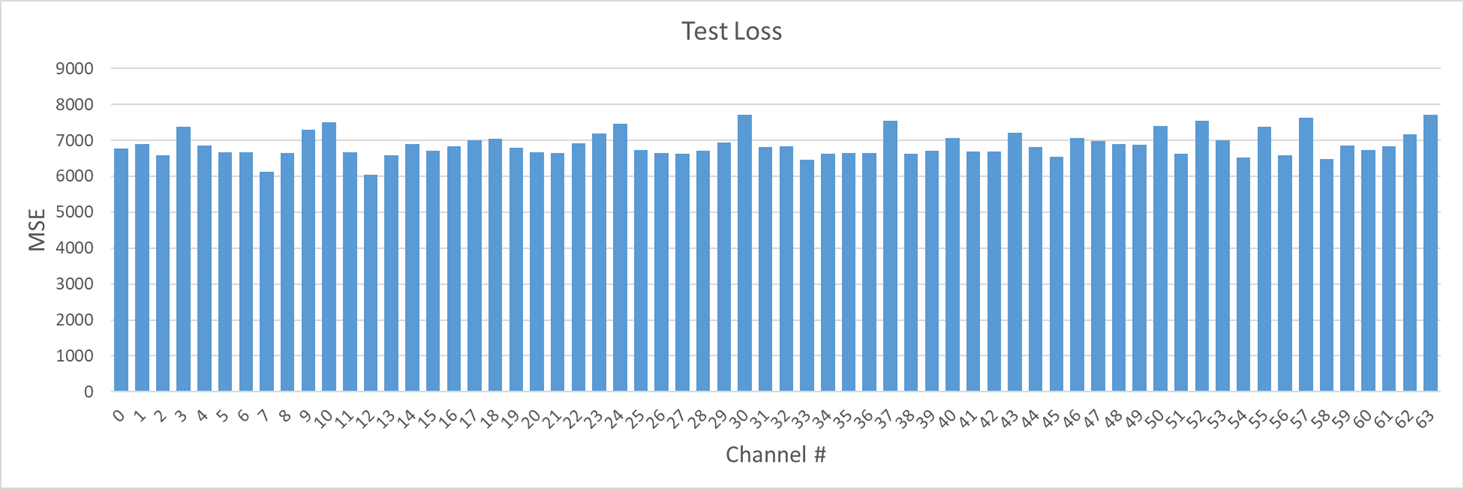}
\caption{Channelwise test losses for the explainer model on delay spread for layer B1-PRE, showing poor predictive performance across all channels.}
\label{fig:nucpi_delay_spread_channelwise_b1-pre_test_losses}
\end{figure}

\begin{figure}[h]
\centering
\includegraphics[width=\columnwidth]{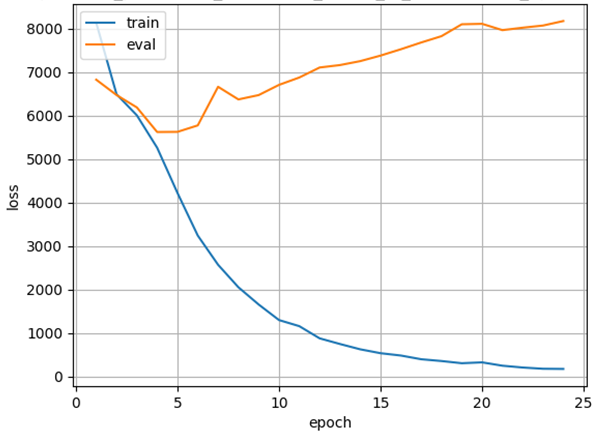}
\caption{Training curve on delay spread for layer B1-PRE, illustrating overfitting in the explainer model.}
\label{fig:nucpi_delay_spread_layerwise_b1-pre_training_curve}
\end{figure}

\begin{figure}[h]
\centering
\includegraphics[width=\columnwidth]{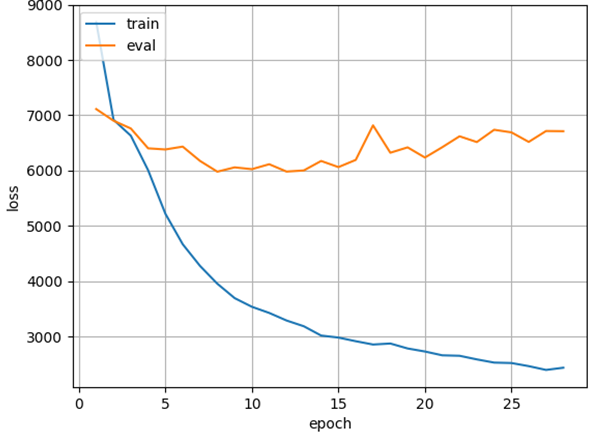}
\caption{Training curve on delay spread for layer OUT-POST, showing overfitting in the explainer model.}
\label{fig:nucpi_delay_spread_layerwise_out-post_training_curve}
\end{figure}

\begin{figure}[h]
\centering
\includegraphics[width=\columnwidth]{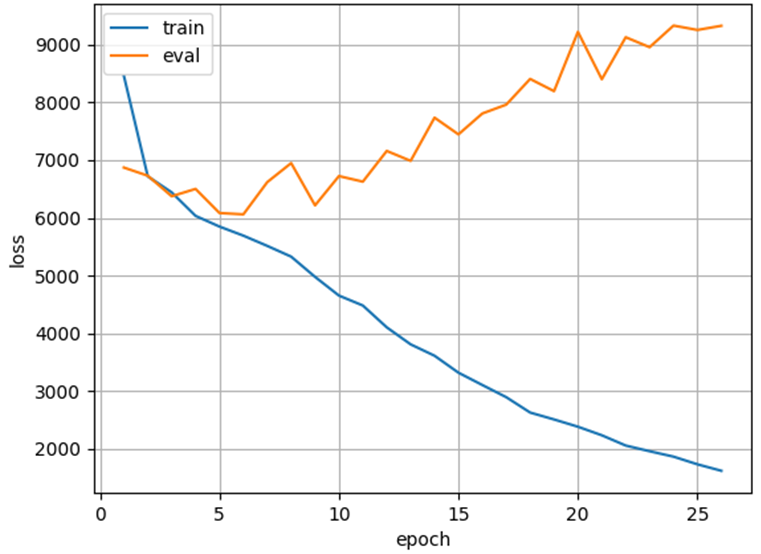}
\caption{Training curve on delay spread for channel 7 on layer B1-PRE, indicating overfitting in the explainer model.}
\label{fig:nucpi_delay_spread_channelwise_b1-pre_channel7_training_curve}
\end{figure}

\begin{figure}[h]
\centering
\includegraphics[width=\columnwidth]{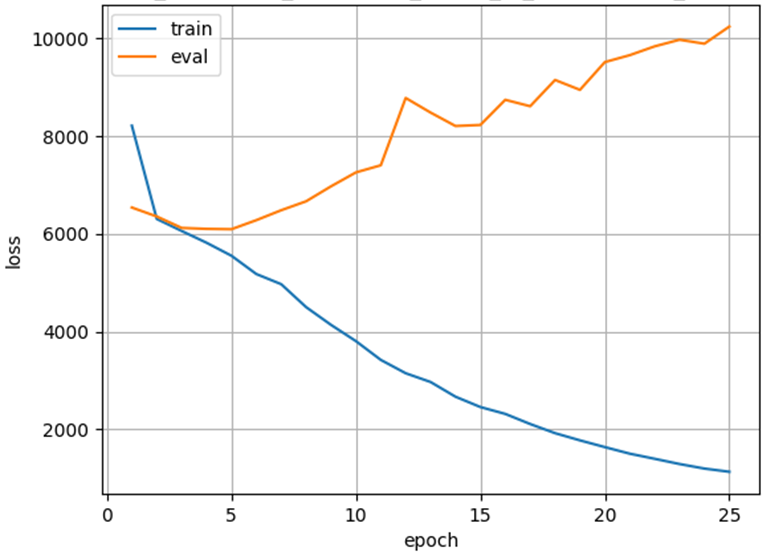}
\caption{Training curve on delay spread for channel 12 on layer B1-PRE, illustrating overfitting in the explainer model.}
\label{fig:nucpi_delay_spread_channelwise_b1-pre_channel12_training_curve}
\end{figure}

\subsection{Experiments on Radio Receiver}
\subsubsection{NUCPI Analysis of Delay Spread}
\label{appendix:nucpi_delay_spread}
We performed NUCPI analysis of delay spread using both layerwise and channelwise (for layer B1-PRE) perspectives. As shown in the layerwise (Figure~\ref{fig:nucpi_delay_spread_layerwise_test_losses}) and channelwise (Figure~\ref{fig:nucpi_delay_spread_channelwise_b1-pre_test_losses}) test losses of the explainer models, predictive performance is consistently poor across all units. Upon closer examination, it is evident that the explainer models are overfitting: while training loss decreases steadily, evaluation loss plateaus or increases, indicating limited generalization to unseen data \cite{goodfellow2016deep}. This is illustrated in training curves for both the layerwise (B1-PRE in Figure~\ref{fig:nucpi_delay_spread_layerwise_b1-pre_training_curve}; OUT-POST in Figure~\ref{fig:nucpi_delay_spread_layerwise_out-post_training_curve}) and channelwise (channels 7 and 12 on B1-PRE, in Figures~\ref{fig:nucpi_delay_spread_channelwise_b1-pre_channel7_training_curve} and \ref{fig:nucpi_delay_spread_channelwise_b1-pre_channel12_training_curve}, respectively) explainer models.

\begin{figure*}
\centering
\includegraphics[height=\textwidth, angle=-90]{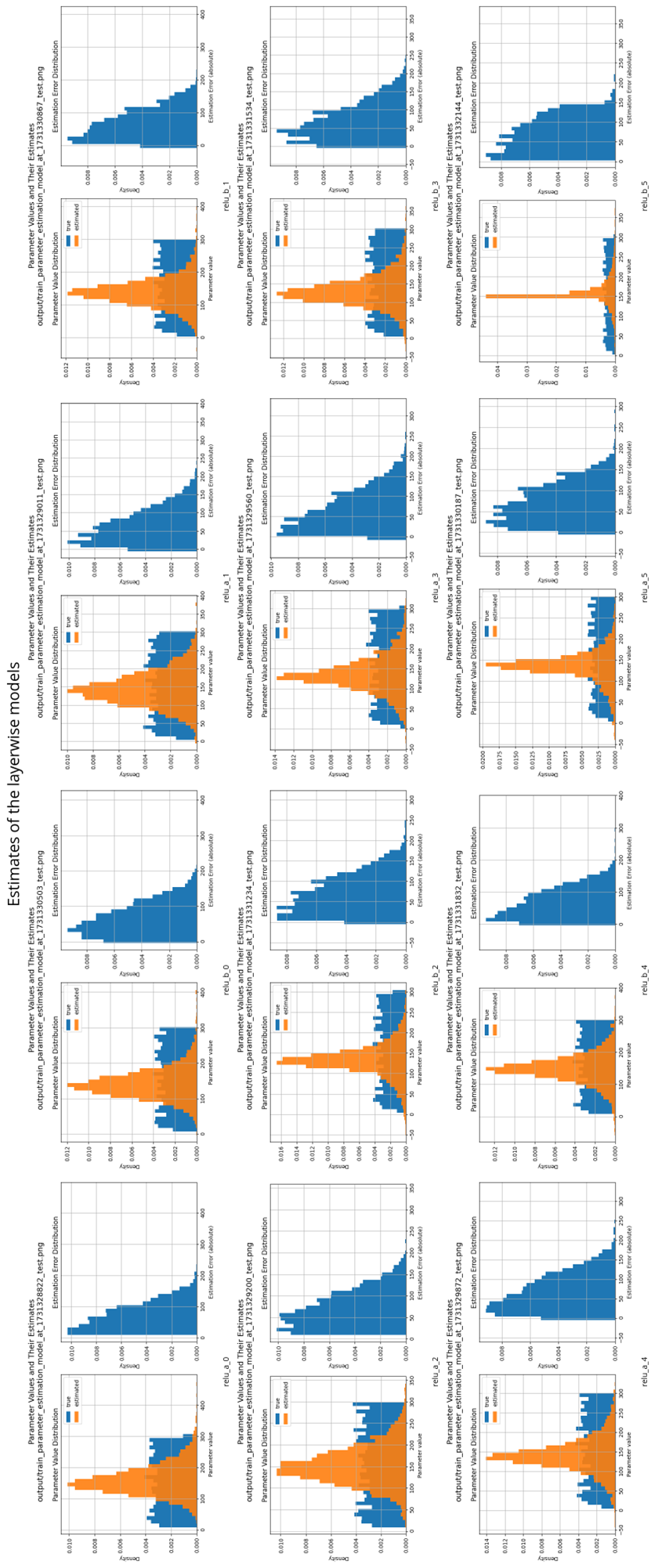}
\caption{Estimates of delay spread by layerwise explainer models, peaking around the mean of true delay spread parameters.}
\label{fig:nucpi_delay_spread_estimates_of_the_layerwise_models}
\end{figure*}

\begin{figure*}
\centering
\includegraphics[height=\textwidth, angle=-90]{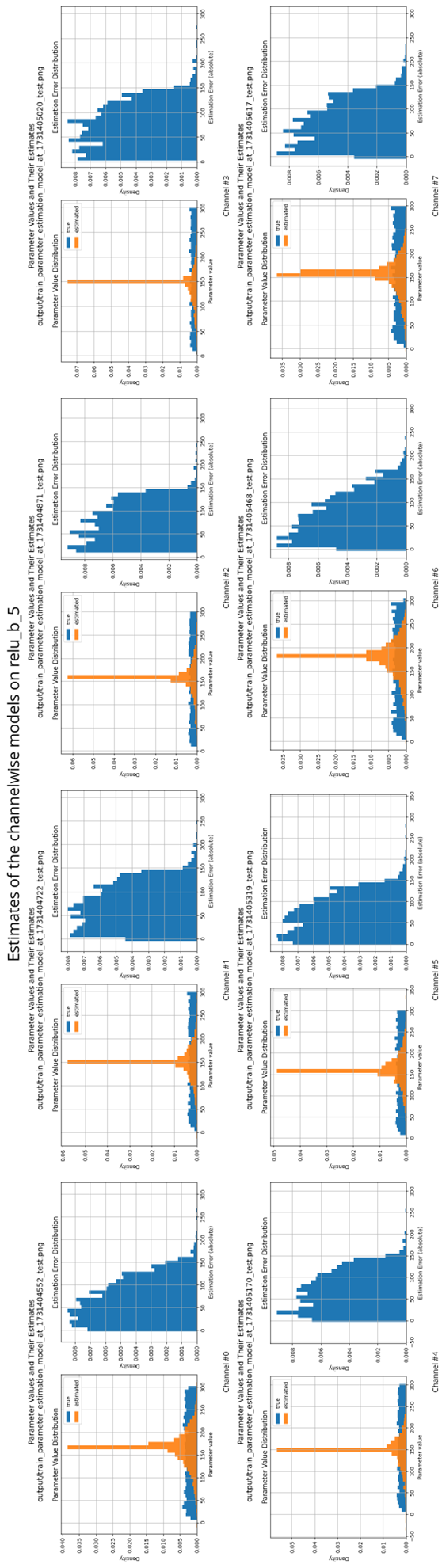}
\caption{Estimates of delay spread by channelwise explainer models on layer OUT-POST, peaking around the mean of true delay spread parameters.}
\label{fig:nucpi_delay_spread_estimates_of_the_channelwise_models_on_out-post}
\end{figure*}

\begin{figure*}
\centering
\includegraphics[width=0.70\textwidth]{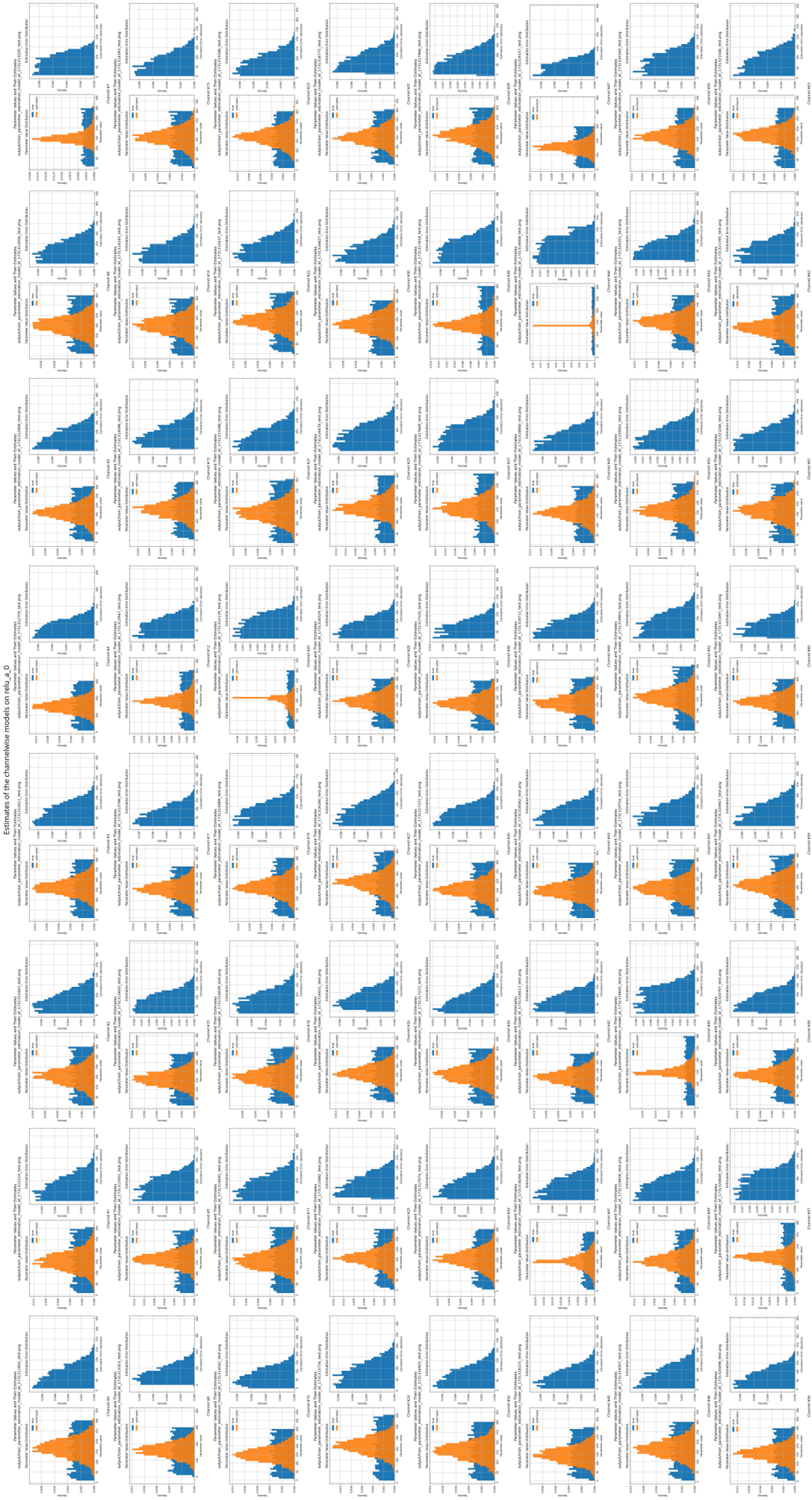}
\caption{Estimates of delay spread by channelwise explainer models on layer B1-PRE, peaking around the mean of true delay spread parameters.}
\label{fig:nucpi_delay_spread_estimates_of_the_channelwise_models_on_b1-pre}
\end{figure*}

The estimates produced by the explainer models (see Figure~\ref{fig:nucpi_delay_spread_estimates_of_the_layerwise_models} for layerwise, and Figures~\ref{fig:nucpi_delay_spread_estimates_of_the_channelwise_models_on_b1-pre} and~\ref{fig:nucpi_delay_spread_estimates_of_the_channelwise_models_on_out-post} for channelwise) show that the models tend to predict values near the mean of the delay spread distribution. This pattern occurs both in layerwise and channelwise models, suggesting that the models are effectively ignoring input variations and defaulting to the values around the global average. We conclude that the activations of the DeepRx model likely do not contain meaningful information about delay spread, at least not in a directly predictive form.

From a physical layer perspective, this is plausible given our simulation parameters (see Table~\ref{tab:simulation_parameters}). The bandwidth is $B = 192 \times 30\text{ kHz} = 5.76\text{ MHz}$, yielding a time resolution of $\Delta\tau = 1/B \approx 174\text{ ns}$, and a quantization uncertainty of approximately $\sigma \approx \Delta\tau/\sqrt{12} \approx \pm 50\text{ ns}$ (standard deviation of a uniform distribution~\cite[Appendix 15]{earl2021_appendix15}). As a result, multipath components closer than 174 ns apart (e.g., a 10 ns path) are indistinguishable, and even larger separations are affected by quantization noise. This limitation likely contributes to the lack of delay spread information in DeepRx’s activations. Interestingly, DeepRx still performs well under delay spread conditions, suggesting it may compensate for this effect without needing to explicitly quantify the delay spread (e.g., as RMS).

\subsubsection{NUCPI Analysis of User Equipment (UE) Speed}
\label{appendix:nucpi_ue_speed}
\begin{figure}
\centering
\includegraphics[width=\columnwidth]{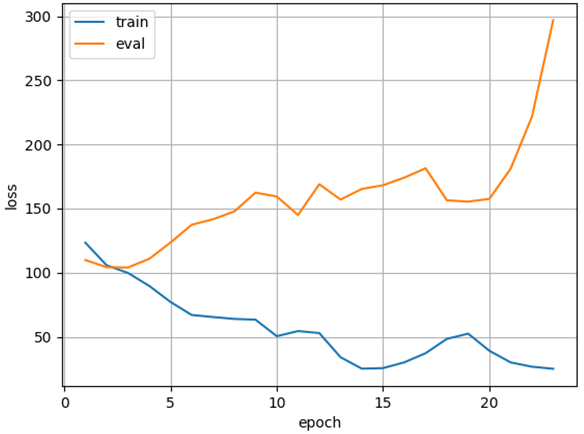}
\caption{Training curve on UE speed for layer B1-PRE, illustrating overfitting in the explainer model.}
\label{fig:nucpi_ue_speed_layerwise_b1-pre_training_curve}
\end{figure}

\begin{figure}
\centering
\includegraphics[width=\columnwidth]{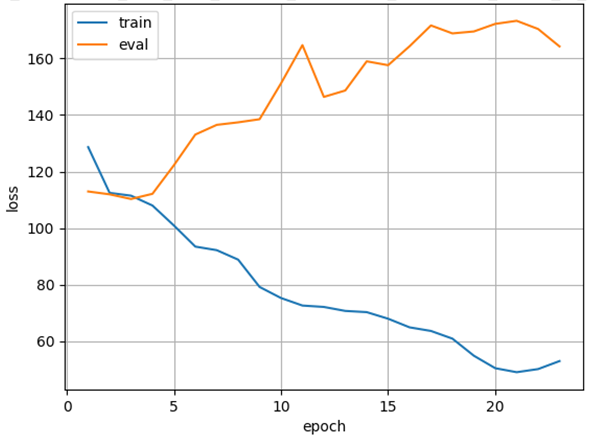}
\caption{Training curve on UE speed for layer OUT-POST, illustrating overfitting in the explainer model.}
\label{fig:nucpi_ue_speed_layerwise_out-post_training_curve}
\end{figure}

We also applied NUCPI analysis to UE speed, evaluating layerwise explainer performance for two layers. As with delay spread (see Appendix~\ref{appendix:nucpi_delay_spread}), the explainer models exhibit consistent overfitting: training losses decrease, but evaluation losses do not improve, as shown in Figures~\ref{fig:nucpi_ue_speed_layerwise_b1-pre_training_curve} and~\ref{fig:nucpi_ue_speed_layerwise_out-post_training_curve}. Based on this behavior, we conclude that DeepRx’s internal activations likely do not encode UE speed in a directly predictive way.

In traditional systems, UE speed---or Doppler spread more technically---is estimated by aggregating over tens of slots (\cite{yucek2005,zhao2009}). In contrast, DeepRx operates on a single slot without a memory of past slots. Despite this, DeepRx can still handle frequency offset effects caused by user movement (and other factors). A similar explanation as with delay spread may apply: the model may not need to explicitly estimate frequency offset or Doppler spread to effectively compensate for them.

%% file: supplemental/umap.tex
\section{Uniform Manifold Approximation and Projection (UMAP) Methodology for Manifold Analysis}
\label{supplemental:umap_embedding}
We use Uniform Manifold Approximation and Projection (UMAP)~\cite{mcinnes2018umap} to visualize the structure of high-dimensional activation features in two dimensions. UMAP is a nonlinear, stochastic dimensionality-reduction technique that aims to preserve local neighborhood relationships while maintaining coarse global structure. As a result, local geometry and exact point placement may vary across random initializations, and UMAP embeddings should be interpreted qualitatively rather than metrically. In this work, UMAP is used solely as a visualization tool to reveal large-scale organization in the learned representations; all quantitative results are computed in the original feature space. Repetition experiments assessing the robustness of the structures reported in the main paper (Section~\ref{sec:results_cluster_analysis}) are provided below.

\subsection{Repetition Experiments on Radio Receiver}
\label{supplemental:umap_repetition_radio}
\begin{figure}[t]
\centering
\includegraphics[trim={0 0 0 0cm},clip,width=\columnwidth]{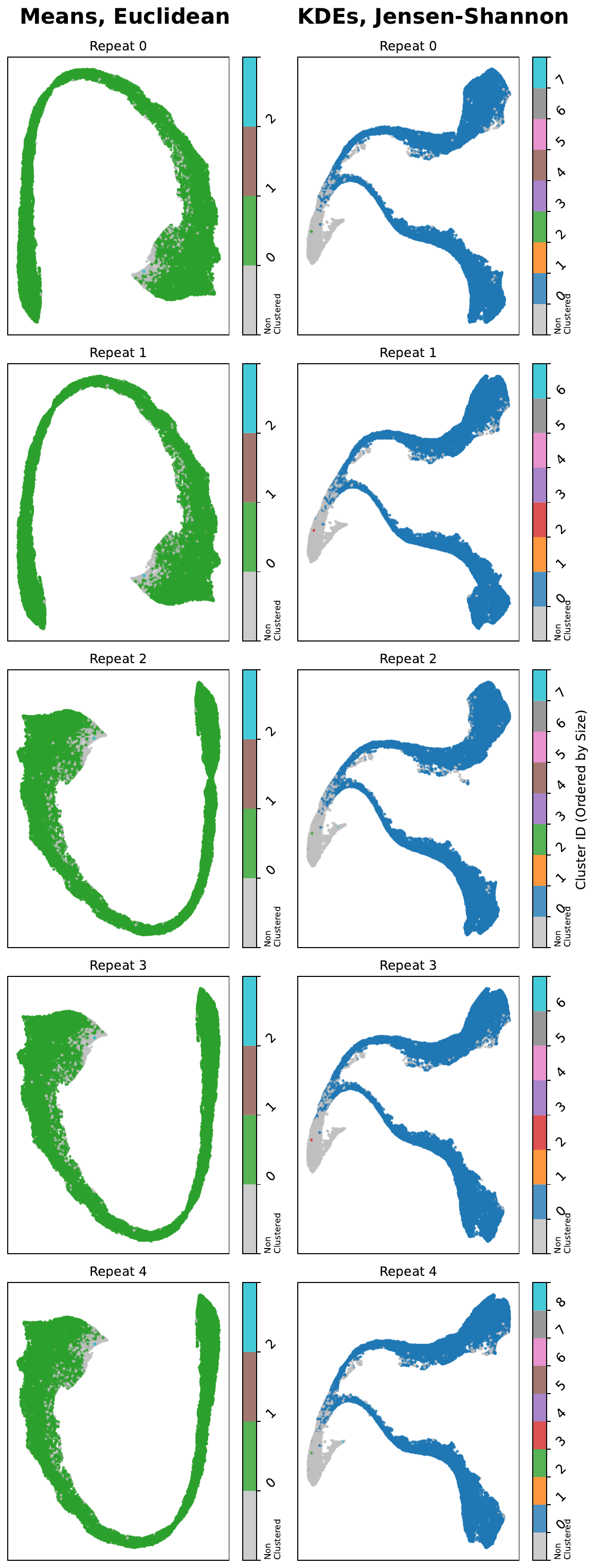}
\caption{UMAP embeddings of DeepRx activations from layer B1-PRE across five independent runs using the baseline (left) and proposed (right) methodology, \textbf{colored by cluster partitioning}. UMAP is used solely for visualization.}
\label{fig:data_structure_partitioning}
\end{figure}

\begin{figure}[t]
\centering
\includegraphics[trim={0 0 0 0cm},clip,width=\columnwidth]{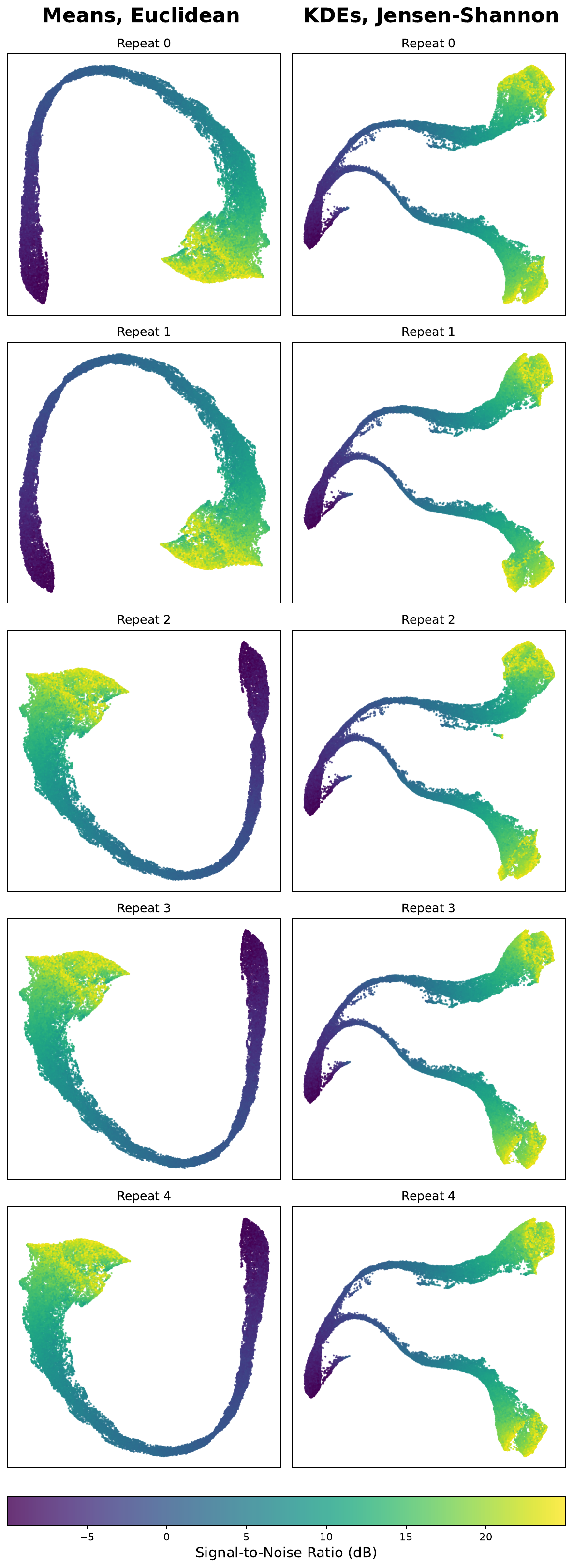}
\caption{UMAP embeddings of DeepRx activations from layer B1-PRE across five independent runs using the baseline (left) and proposed (right) methodology, \textbf{colored by SNR}. UMAP is used solely for visualization.}
\label{fig:data_structure_snr}
\end{figure}

\begin{figure}[t]
\centering
\includegraphics[trim={0 0 0 0cm},clip,width=\columnwidth]{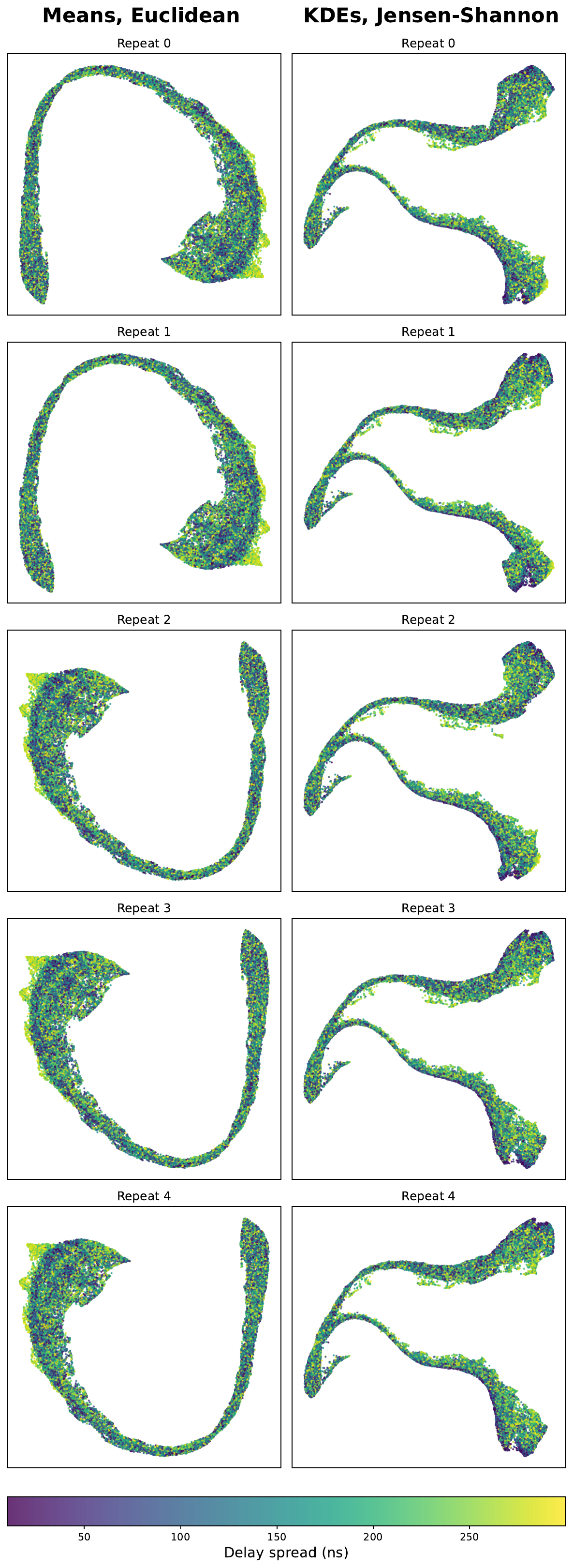}
\caption{UMAP embeddings of DeepRx activations from layer B1-PRE across five independent runs using the baseline (left) and proposed (right) methodology, \textbf{colored by delay spread}. UMAP is used solely for visualization.}
\label{fig:data_structure_delay_spread}
\end{figure}

\begin{figure}[t]
\centering
\includegraphics[trim={0 0 0 0cm},clip,width=\columnwidth]{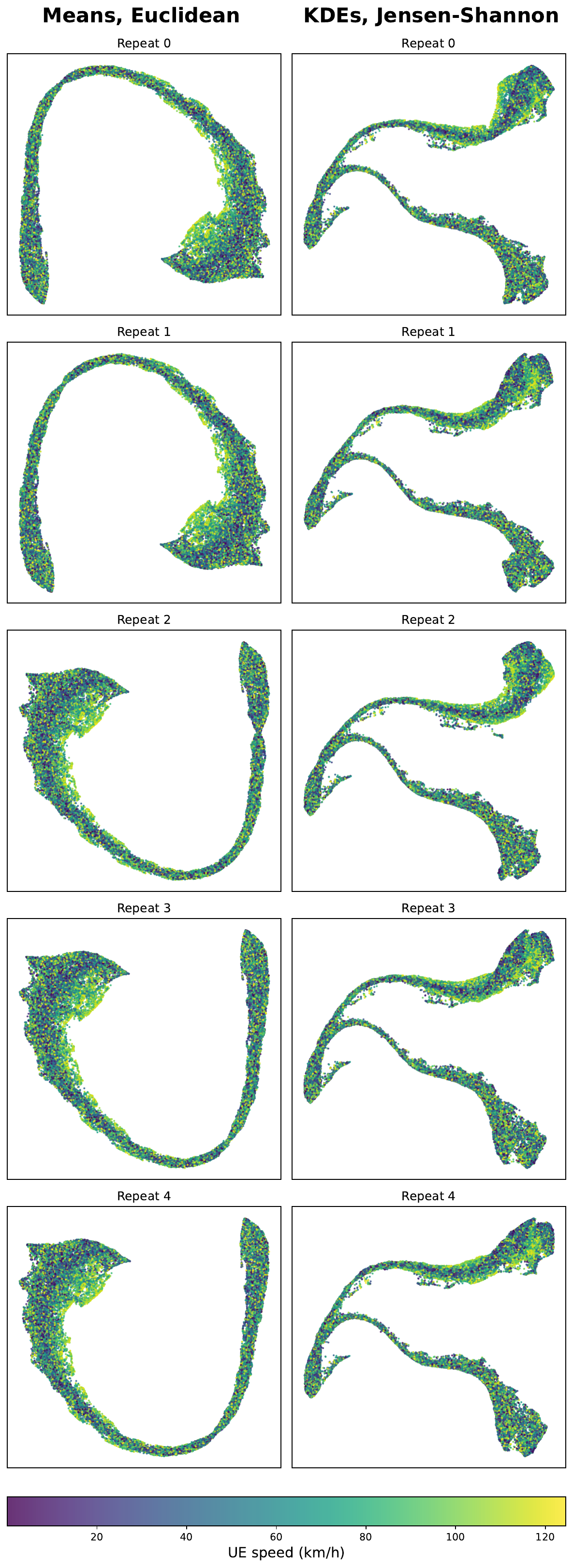}
\caption{UMAP embeddings of DeepRx activations from layer B1-PRE across five independent runs using the baseline (left) and proposed (right) methodology, \textbf{colored by UE speed}. UMAP is used solely for visualization.}
\label{fig:data_structure_ue_speed}
\end{figure}

\begin{figure}[t]
\centering
\includegraphics[trim={0 0 0 0cm},clip,width=\columnwidth]{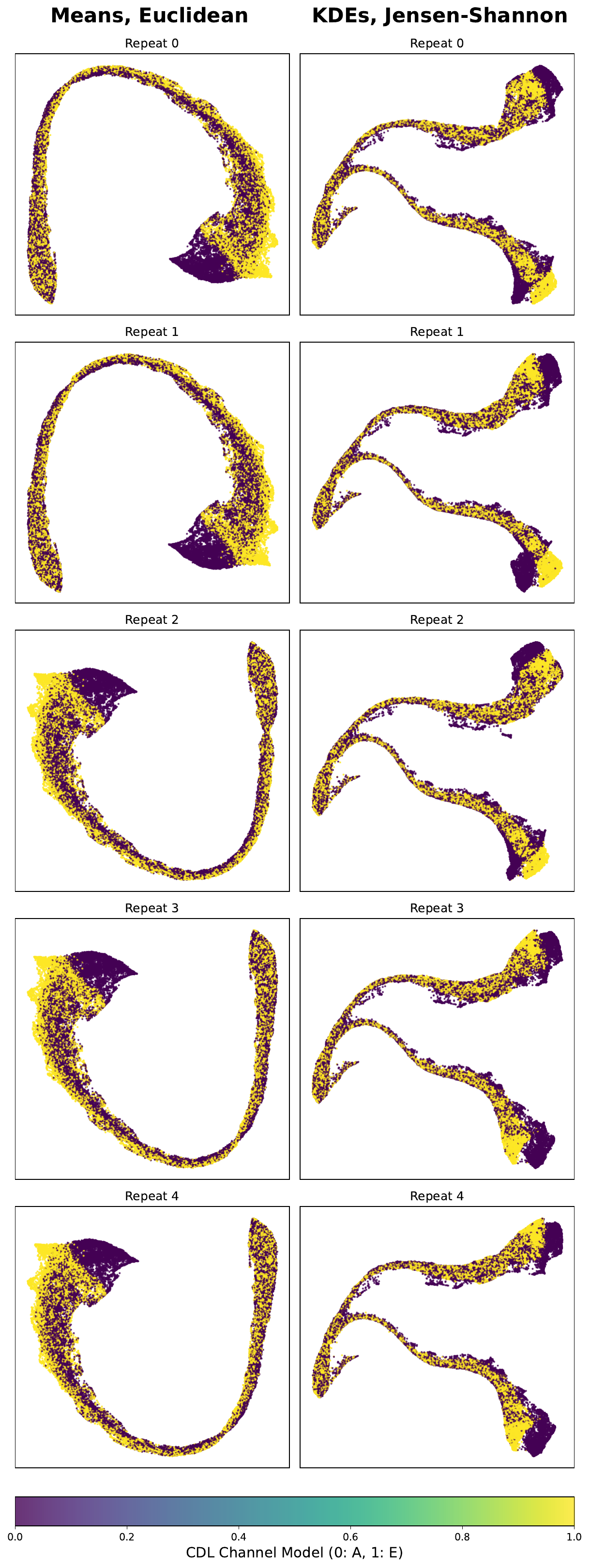}
\caption{UMAP embeddings of DeepRx activations from layer B1-PRE across five independent runs using the baseline (left) and proposed (right) methodology, \textbf{colored by channel model}. UMAP is used solely for visualization.}
\label{fig:data_structure_channel}
\end{figure}

As noted above, UMAP is a stochastic dimensionality-reduction technique, and its local geometry may vary across random initializations. To assess the robustness of the observed activation structure, we repeat the UMAP embedding five times using different random seeds for DeepRx activations from layer B1-PRE. Figures~\ref{fig:data_structure_partitioning}--\ref{fig:data_structure_channel} show the resulting embeddings for both the baseline and proposed methodologies, colored by different metadata attributes.

Across all repetitions, the global organization of the embeddings remains stable. In particular, samples consistently form a smooth, low-dimensional manifold primarily aligned with SNR (Fig.~\ref{fig:data_structure_snr}), while other channel attributes such as delay spread, UE speed, and channel model induce only secondary variations (Figs.~\ref{fig:data_structure_delay_spread}--\ref{fig:data_structure_channel}). Although local neighborhoods and fine-scale layout may differ between runs--as expected for UMAP--the dominant SNR-driven structure is preserved. This confirms that the observed manifold is not an artifact of a particular UMAP initialization but reflects a robust property of the learned receiver representations.

%% file: supplemental/conditional_knn.tex
\section{Conditional k-NN Methodology for Out-Of-Distribution Detection}
\label{supplemental:conditional_knn}
Let each training sample be represented by an activation vector $a_i \in \mathbb{R}^d$ and an associated SNR parameter $p_i \in \mathbb{R}$, and let the test sample and its parameter be
$(a_*, p_*)$. In a standard $k$-nearest-neighbour (k-NN) method, only distances in activation space are considered, typically
\begin{equation}
    d_i = \| a_i - a_* \|,
\end{equation}
and the $k$ smallest distances determine the neighborhood for test sample in question. However, the relationship between training and test samples may depend on the SNR level.  To capture this dependence,
we introduce a \emph{conditional distance} in which SNR differences influence the
effective neighbour selection.

\subsection{Hard Conditional Scheme}
\label{supplemental:conditional_knn_hard}
In the hard variant, SNR is used as a strict constraint. A training point is considered eligible only if its SNR is sufficiently close to the test SNR. Let $D > 0$ be the maximum allowed SNR deviation.  Define
\begin{equation}
    \Delta p_i = | p_i - p_* |.
\end{equation}
The effective distance is then
\begin{equation}
    d_i^{\text{eff}} =
    \begin{cases}
        d_i, & \text{if } \Delta p_i \le D, \\[4pt]
        \infty, & \text{otherwise}.
    \end{cases}
\end{equation}
Thus, neighbours are selected \emph{only} from activation-space distances of points whose SNR is within a hard threshold of $p_*$, and all other points are excluded.

\subsection{Soft Conditional Scheme}
\label{supplemental:conditional_knn_soft}
The hard cutoff can be replaced by a smooth penalty that grows with the SNR difference, allowing all points to contribute while favouring those with more similar SNR. Let $m(\Delta p_i)$ be a monotone increasing penalty function with $m(0) = 1$, and define the effective distance as
\begin{equation}
    d_i^{\text{eff}} = m(\Delta p_i)\, d_i.
\end{equation}

In these soft schemes, the $k$ neighbours are chosen as the training points with the smallest effective distances $d_i^{\text{eff}}$, enabling a continuous transition between strict conditioning and unrestricted k-NN. Two practical choices for $m(\cdot)$ are power-law and exponential penalty, described below in more detail.

\subsubsection{Power-law penalty.}
Let $D$ denote the characteristic SNR scale and $F > 1$ the penalty factor at $\Delta p_i = D$. With exponent $p \ge 1$,
\begin{equation}\label{eq:power_law_penalty}
    m(\Delta p_i)
= 1 + (F - 1)\left( \frac{\Delta p_i}{D} \right)^{p}.
\end{equation}
This yields linear ($p=1$) or stronger ($p>1$) growth of the penalty as the SNR difference increases.

\subsubsection{Exponential (Gaussian) penalty.}
Define a bandwidth $h > 0$ and let
\begin{equation}\label{eq:exponential_penalty}
    m(\Delta p_i) = \exp\!\left( \frac{1}{2}\left( \frac{\Delta p_i}{h} \right)^p \right).
\end{equation}
The bandwidth $h$ may be chosen so that $m(D)$ matches a desired penalty factor $F$, leading to
\begin{equation}
    h = \frac{D}{\sqrt[p]{2\ln F}}.
\end{equation}
In this formulation, points with larger SNR mismatch receive exponentially inflated effective distances and thus are gradually deprioritised in the neighbour ranking.

\begin{figure}[ht]
    \centering
    \includegraphics[width=\columnwidth]{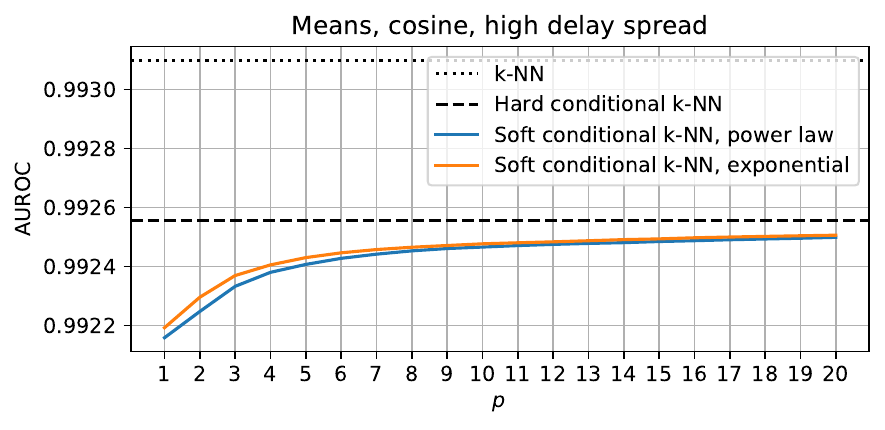}
    \caption{OOD detection performance of k-NN-based detectors under high delay spread on layer B1-PRE as a function of the penalty exponent $p$ in (\ref{eq:power_law_penalty}) and (\ref{eq:exponential_penalty}). Mean-activation features and the cosine distance are used.}
    \label{fig:figure_soft_knn_experiments_relu_a_0_means_cosine_delay_spread}
\end{figure}
\begin{figure}[ht]
    \centering
    \includegraphics[width=\columnwidth]{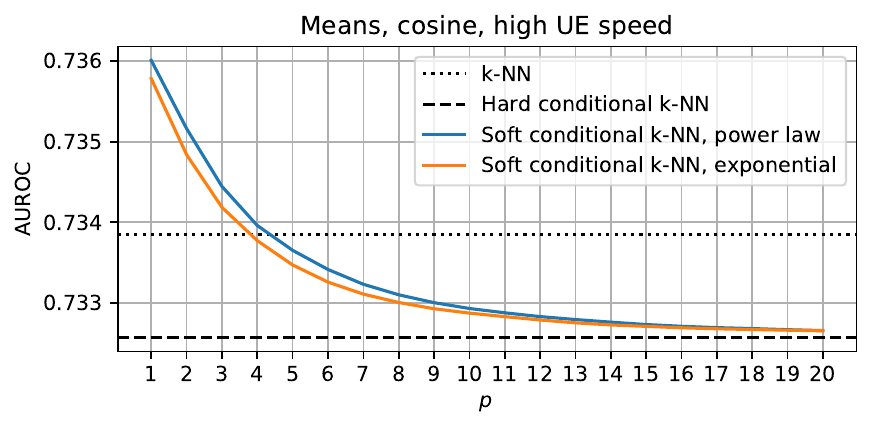}
    \caption{OOD detection performance of k-NN-based detectors under high UE speed on layer B1-PRE as a function of the penalty exponent $p$ in (\ref{eq:power_law_penalty}) and (\ref{eq:exponential_penalty}). Mean-activation features and the cosine distance are used.}
    \label{fig:figure_soft_knn_experiments_relu_a_0_means_cosine_ue_speed}
\end{figure}

\subsection{Parameter Selection}
\label{supplemental:conditional_knn_parameters}
The SNR tolerance $D$ is selected as the 99th percentile of absolute SNR differences observed in the ID training set, reflecting a typical operating range. The penalty factor is fixed to $F=2.0$ across experiments.

Conditional k-NN is evaluated using mean-activation features with cosine distance, as this combination yields the strongest k-NN baseline performance on average across the two OOD scenarios (Table~\ref{tab:single_model_baselines_on_b1-pre_average_over_scenarios}). We adopt the exponential penalty, as it degrades performance less than the power-law alternative in the high delay scenario, and set the exponent to $p=1$, which provides the largest gains in the high speed scenario.

Figures~\ref{fig:figure_soft_knn_experiments_relu_a_0_means_cosine_delay_spread} and~\ref{fig:figure_soft_knn_experiments_relu_a_0_means_cosine_ue_speed} illustrate the sensitivity of OOD detection performance to the penalty exponent under high delay spread and high UE speed, respectively.

%% file: supplemental/conformal_p_extension.tex
\section{Continuous Conformal P-Values Methodology for Out-Of-Distribution Detection}
\label{supplemental:extension_to_conformal_p}
We briefly recall the Benjamini--Hochberg (BH)-style conformal testing rule of Algorithm~1 in~\cite{magesh2023conformalpvalues}.
Let $\{\hat{Q}_{(i)}\}_{i=1}^K$ denote the ordered conformal p-values across $K$ detectors. The algorithm defines
\begin{equation}
    m = \max \left\{i : \hat{Q}_{(i)} \le \frac{\alpha\,i}{C(K)\,K} \right\}
\end{equation}
and declares a sample to be Out-Of-Distribution (OOD) if
\begin{equation}
    m \ge 1 .
\end{equation}
Equivalently, the sample is classified as OOD if there exists an index $i$ such that
\begin{equation}
    \hat{Q}_{(i)} \le \frac{\alpha\, i}{C(K)\,K}.
\end{equation}
Rearranging this condition yields
\begin{equation}
    \min_{i=1,\dots,K}
    \left(
        \frac{C(K)\,K}{i}\,\hat{Q}_{(i)}
    \right)
    \le \alpha .
\end{equation}

This observation allows us to convert the binary rejection rule of Algorithm~1 into a continuous score. We define the scalar
\begin{equation}
    S
    =
    \min_{i=1,\dots,K}
    \left(
        \frac{C(K)\,K}{i}\,\hat{Q}_{(i)}
    \right),
\end{equation}
which represents the smallest significance level $\alpha$ at which the sample would be declared OOD under the BH-style conformal test. While this continuous score is not introduced as a separate algorithm in prior work, it follows directly from the decision rule in~\cite{magesh2023conformalpvalues}.

The score $S$ provides a calibrated, continuous measure of out-of-distribution confidence and can be used to construct ROC or DET curves, enabling quantitative comparison and fusion with other OOD detectors.

%% file: supplemental/ood.tex
\section{Supplemental Experimental Results on Out-Of-Distribution Detection on Radio Receiver}
\label{supplemental:results}
This section provides detailed numerical results supporting the Out-Of-Distribution (OOD) detection experiments reported in the main paper (Section~\ref{sec:ood_experiments}). 
We organize results by analysis type--single-detector baselines, SNR-aware modeling, and detector fusion--while reporting performance separately for the three representative layers analyzed in the main text (B1-PRE, B3-POST, and OUT-POST).
All results are reported for both high delay and high speed OOD scenarios, as well as their average when appropriate.
Unless stated otherwise, results are averaged over five independent runs with different random seeds for DeepRx activations.

\subsection{Single-Detector Baselines}
\label{supplemental:results_single_model_baselines}
We first report performance of individual OOD detectors applied to a single layer without SNR conditioning or fusion.
Results for layers B1-PRE, B3-POST, and OUT-POST are reported separately in Tables~\ref{tab:single_model_baselines_on_b1-pre_high-delay}--\ref{tab:single_model_baselines_on_b6-post_average_over_scenarios}.

\subsection{Effect of SNR-Aware Modeling}
\label{supplemental:results_effect_of_snr}
We evaluate the impact of incorporating SNR information into OOD detection using two strategies:
(i) conditional k-NN (hard and soft conditioning), and
(ii) feature concatenation for OC-SVM and Mahalanobis detectors.
Results are reported for all three layers across both OOD scenarios in Tables~\ref{tab:snr_regimes_on_b1-pre_high-delay}--\ref{tab:snr_regimes_on_b6-post_high-speed}. Soft conditional k-NN yields the largest gains under high UE speed on B1-PRE, whereas hard conditioning is more robust to increased delay spread (Figs.~\ref{fig:figure_soft_knn_experiments_relu_a_0_means_cosine_delay_spread}--\ref{fig:figure_soft_knn_experiments_relu_a_0_means_cosine_ue_speed}).

\subsection{Within-Layer Fusion}
\label{supplemental:results_within_layer_fusion}
We evaluate the fusion of multiple detectors applied to the same layer using OC-SVM and conformal p-value aggregation.
All 27 combinations of the four activation-based detectors and raw SNR are evaluated per layer:
\begin{itemize}
    \item B1-PRE: Tables~\ref{tab:classifier_fusion_on_b1-pre_high-delay}--\ref{tab:single_model_within_layer_fusions_on_b1-pre_average_over_scenarios};
    \item B3-POST: Tables~\ref{tab:classifier_fusion_on_b3-post_high-delay}--\ref{tab:single_model_within_layer_fusions_on_b3-post_average_over_scenarios};
    \item OUT-POST: Tables~\ref{tab:classifier_fusion_on_b6-post_high-delay}--\ref{tab:single_model_within_layer_fusions_on_b6-post_average_over_scenarios};
\end{itemize}
and best-performing fusions compiled in Table~\ref{tab:best_fusions_within_layer_scenario_average}.

\subsection{Across-Layer Fusion}
\label{supplemental:results_across_layer_fusion}
Finally, we evaluate across-layer fusion, where detector outputs from multiple layers are combined to exploit complementary information across network depth. We consider fusing raw detector outputs as well as within-layer fused scores across B1-PRE, B3-POST, OUT-POST, and SNR. All 11 combinations of the scores are evaluated per detector:
\begin{itemize}
    \item HDBSCAN: Tables~\ref{tab:across_layer_fusion_on_hdbscan_high-delay}--\ref{tab:across_layer_fusion_on_hdbscan_average_over_scenarios};
    \item k-NN: Tables~\ref{tab:across_layer_fusion_on_knn_high-delay}--\ref{tab:across_layer_fusion_on_knn_average_over_scenarios};
    \item OC-SVM: Tables~\ref{tab:across_layer_fusion_on_ocsvm_high-delay}--\ref{tab:across_layer_fusion_on_ocsvm_average_over_scenarios};
    \item Mahalanobis: Tables~\ref{tab:across_layer_fusion_on_mahalanobis_high-delay}--\ref{tab:across_layer_fusion_on_mahalanobis_average_over_scenarios};
    \item Within-layer: Tables~\ref{tab:across_layer_fusion_on_within_layer_fusions_high-delay}--\ref{tab:across_layer_fusion_on_within_layer_fusions_average_over_scenarios};
\end{itemize}
and best-performing fusions compiled in Table~\ref{tab:best_fusions_across_layers_scenario_average}.

\clearpage
\begin{table*}
    \centering
    \input{supplemental/tables/ood/table_single_model_baselines_on_b1-pre_high-delay}
    \caption{OOD detection performance under high delay spread on layer B1-PRE. Mean over 5 independent repeats with random seeds. All standard deviations less than 0.055.}
    \label{tab:single_model_baselines_on_b1-pre_high-delay}
\end{table*}
\begin{table*}
    \centering
    \input{supplemental/tables/ood/table_single_model_baselines_on_b1-pre_high-speed}
    \caption{OOD detection performance under high UE speed on layer B1-PRE. Mean over 5 independent repeats with random seeds. All standard deviations less than 0.025.}
    \label{tab:single_model_baselines_on_b1-pre_high-speed}
\end{table*}
\begin{table*}
    \centering
    \input{supplemental/tables/ood/table_single_model_baselines_on_b3-post_high-delay}
    \caption{OOD detection performance under high delay spread on layer B3-POST. Mean over 5 independent repeats with random seeds. All standard deviations less than 0.035, except for 0.28 for HDBSCAN Energy + Euclidean.}
    \label{tab:single_model_baselines_on_b3-post_high-delay}
\end{table*}
\begin{table*}
    \centering
    \input{supplemental/tables/ood/table_single_model_baselines_on_b3-post_high-speed}
    \caption{OOD detection performance under high UE speed on layer B3-POST. Mean over 5 independent repeats with random seeds. All standard deviations less than 0.015, except for 0.19 for HDBSCAN Energy + Euclidean.}
    \label{tab:single_model_baselines_on_b3-post_high-speed}
\end{table*}
\begin{table*}
    \centering
    \input{supplemental/tables/ood/table_single_model_baselines_on_b3-post_average_over_scenarios}
    \caption{Average OOD detection performance over high delay spread and UE speed scenarios on layer B3-POST. Mean over 5 independent repeats with random seeds.}
    \label{tab:single_model_baselines_on_b3-post_average_over_scenarios}
\end{table*}
\begin{table*}
    \centering
    \input{supplemental/tables/ood/table_single_model_baselines_on_b6-post_high-delay}
    \caption{OOD detection performance under high delay spread on layer OUT-POST. Mean over 5 independent repeats with random seeds. All standard deviations less than 0.015.}
    \label{tab:single_model_baselines_on_b6-post_high-delay}
\end{table*}
\begin{table*}
    \centering
    \input{supplemental/tables/ood/table_single_model_baselines_on_b6-post_high-speed}
    \caption{OOD detection performance under high UE speed on layer OUT-POST. Mean over 5 independent repeats with random seeds. All standard deviations less than 0.015.}
    \label{tab:single_model_baselines_on_b6-post_high-speed}
\end{table*}
\begin{table*}
    \centering
    \input{supplemental/tables/ood/table_single_model_baselines_on_b6-post_average_over_scenarios}
    \caption{Average OOD detection performance over high delay spread and UE speed scenarios on layer OUT-POST. Mean over 5 independent repeats with random seeds.}
    \label{tab:single_model_baselines_on_b6-post_average_over_scenarios}
\end{table*}

\begin{landscape}
\begin{table}
    \centering
    \input{supplemental/tables/ood/table_snr_regimes_on_b1-pre_high-delay}
    \caption{OOD detection performance under high delay spread on layer B1-PRE. Mean over 5 independent repeats with random seeds. All standard deviations less than 0.015.}
    \label{tab:snr_regimes_on_b1-pre_high-delay}
\end{table}
\begin{table}
    \centering
    \input{supplemental/tables/ood/table_snr_regimes_on_b1-pre_high-speed}
    \caption{OOD detection performance under high UE speed on layer B1-PRE. Mean over 5 independent repeats with random seeds. All standard deviations less than 0.015.}
    \label{tab:snr_regimes_on_b1-pre_high-speed}
\end{table}
\begin{table}
    \centering
    \input{supplemental/tables/ood/table_snr_regimes_on_b3-post_high-delay}
    \caption{OOD detection performance under high delay spread on layer B3-POST. Mean over 5 independent repeats with random seeds. All standard deviations less than 0.015.}
    \label{tab:snr_regimes_on_b3-post_high-delay}
\end{table}
\begin{table}
    \centering
    \input{supplemental/tables/ood/table_snr_regimes_on_b3-post_high-speed}
    \caption{OOD detection performance under high UE speed on layer B3-POST. Mean over 5 independent repeats with random seeds. All standard deviations less than 0.015.}
    \label{tab:snr_regimes_on_b3-post_high-speed}
\end{table}
\begin{table}
    \centering
    \input{supplemental/tables/ood/table_snr_regimes_on_b6-post_high-delay}
    \caption{OOD detection performance under high delay spread on layer B6-POST. Mean over 5 independent repeats with random seeds. All standard deviations less than 0.015.}
    \label{tab:snr_regimes_on_b6-post_high-delay}
\end{table}
\begin{table}
    \centering
    \input{supplemental/tables/ood/table_snr_regimes_on_b6-post_high-speed}
    \caption{OOD detection performance under high UE speed on layer B6-POST. Mean over 5 independent repeats with random seeds. All standard deviations less than 0.015.}
    \label{tab:snr_regimes_on_b6-post_high-speed}
\end{table}
\end{landscape}

\begin{table*}
    \centering
    \input{supplemental/tables/ood/best_fusions_within_layer_scenario_average}
    \caption{Best-performing within-layer fusions by average AUROC across OOD scenarios. Results are averaged over high-delay and high-speed conditions. All the classifiers are applied to the mean features. The k-NN and HDBSCAN classifiers use the cosine distance and Euclidean distances, respectively.}
    \label{tab:best_fusions_within_layer_scenario_average}
\end{table*}
\begin{table*}
    \centering
    \input{supplemental/tables/ood/table_classifier_fusion_on_b1-pre_high-delay}
    \caption{OOD detection performance under high delay spread on layer B1-PRE. Mean over 5 independent repeats with random seeds. All standard deviations less than 0.005. A: mean, Euclidean, HDBSCAN; B: mean, cosine, k-NN (activations only); C: mean, OC-SVM (activations only); D: mean, Mahalanobis (activations only); E: SNR}
    \label{tab:classifier_fusion_on_b1-pre_high-delay}
\end{table*}
\begin{table*}
    \centering
    \input{supplemental/tables/ood/table_classifier_fusion_on_b1-pre_high-speed}
    \caption{OOD detection performance under high UE speed on layer B1-PRE. Mean over 5 independent repeats with random seeds. All standard deviations less than 0.015. A: mean, Euclidean, HDBSCAN; B: mean, cosine, k-NN (activations only); C: mean, OC-SVM (activations only); D: mean, Mahalanobis (activations only); E: SNR}
    \label{tab:classifier_fusion_on_b1-pre_high-speed}
\end{table*}
\begin{table*}
    \centering
    \input{supplemental/tables/ood/table_within_layer_fusions_on_b1-pre_average_over_scenarios}
    \caption{Average OOD detection performance over high delay spread and UE speed scenarios on within-layer fusion for layer B1-PRE. Mean over 5 independent repeats with random seeds. A: mean, Euclidean, HDBSCAN; B: mean, cosine, k-NN (activations only); C: mean, OC-SVM (activations only); D: mean, Mahalanobis (activations only); E: SNR}
    \label{tab:single_model_within_layer_fusions_on_b1-pre_average_over_scenarios}
\end{table*}
\begin{table*}
    \centering
    \input{supplemental/tables/ood/table_classifier_fusion_on_b3-post_high-delay}
    \caption{OOD detection performance under high delay spread on layer B3-POST. Mean over 5 independent repeats with random seeds. All standard deviations less than 0.035. A: mean, Euclidean, HDBSCAN; B: mean, cosine, k-NN (activations only); C: mean, OC-SVM (activations only); D: mean, Mahalanobis (activations only); E: SNR}
    \label{tab:classifier_fusion_on_b3-post_high-delay}
\end{table*}
\begin{table*}
    \centering
    \input{supplemental/tables/ood/table_classifier_fusion_on_b3-post_high-speed}
    \caption{OOD detection performance under high UE speed on layer B3-POST. Mean over 5 independent repeats with random seeds. All standard deviations less than 0.065. A: mean, Euclidean, HDBSCAN; B: mean, cosine, k-NN (activations only); C: mean, OC-SVM (activations only); D: mean, Mahalanobis (activations only); E: SNR}
    \label{tab:classifier_fusion_on_b3-post_high-speed}
\end{table*}
\begin{table*}
    \centering
    \input{supplemental/tables/ood/table_within_layer_fusions_on_b3-post_average_over_scenarios}
    \caption{Average OOD detection performance over high delay spread and UE speed scenarios on within-layer fusion for layer B3-POST. Mean over 5 independent repeats with random seeds. A: mean, Euclidean, HDBSCAN; B: mean, cosine, k-NN (activations only); C: mean, OC-SVM (activations only); D: mean, Mahalanobis (activations only); E: SNR}
    \label{tab:single_model_within_layer_fusions_on_b3-post_average_over_scenarios}
\end{table*}
\begin{table*}
    \centering
    \input{supplemental/tables/ood/table_classifier_fusion_on_b6-post_high-delay}
    \caption{OOD detection performance under high delay spread on layer OUT-POST. Mean over 5 independent repeats with random seeds. All standard deviations less than 0.015. A: mean, Euclidean, HDBSCAN; B: mean, cosine, k-NN (activations only); C: mean, OC-SVM (activations only); D: mean, Mahalanobis (activations only); E: SNR}
    \label{tab:classifier_fusion_on_b6-post_high-delay}
\end{table*}
\begin{table*}
    \centering
    \input{supplemental/tables/ood/table_classifier_fusion_on_b6-post_high-speed}
    \caption{OOD detection performance under high UE speed on layer OUT-POST. Mean over 5 independent repeats with random seeds. All standard deviations less than 0.015. A: mean, Euclidean, HDBSCAN; B: mean, cosine, k-NN (activations only); C: mean, OC-SVM (activations only); D: mean, Mahalanobis (activations only); E: SNR}
    \label{tab:classifier_fusion_on_b6-post_high-speed}
\end{table*}
\begin{table*}
    \centering
    \input{supplemental/tables/ood/table_within_layer_fusions_on_b6-post_average_over_scenarios}
    \caption{Average OOD detection performance over high delay spread and UE speed scenarios on within-layer fusion for layer OUT-POST. Mean over 5 independent repeats with random seeds. A: mean, Euclidean, HDBSCAN; B: mean, cosine, k-NN (activations only); C: mean, OC-SVM (activations only); D: mean, Mahalanobis (activations only); E: SNR}
    \label{tab:single_model_within_layer_fusions_on_b6-post_average_over_scenarios}
\end{table*}

\begin{table*}[t]
    \centering
    \input{supplemental/tables/ood/best_fusions_across_layers_scenario_average}
    \caption{Best-performing across-layer fusions by average AUROC across OOD scenarios. Results are averaged over high-delay and high-speed conditions.}
    \label{tab:best_fusions_across_layers_scenario_average}
\end{table*}
\begin{table*}
    \centering
    \input{supplemental/tables/ood/table_across_layer_fusion_on_hdbscan_high-delay}
    \caption{OOD detection performance under high delay spread using HDBSCAN classifiers with means and the Euclidean distance metric, fused across layers. A: B1-PRE, B: B3-POST, C: OUT-POST, D: SNR}
    \label{tab:across_layer_fusion_on_hdbscan_high-delay}
\end{table*}
\begin{table*}
    \centering
    \input{supplemental/tables/ood/table_across_layer_fusion_on_hdbscan_high-speed}
    \caption{OOD detection performance under high UE speed using HDBSCAN classifiers with means and the Euclidean distance metric, fused across layers. A: B1-PRE, B: B3-POST, C: OUT-POST, D: SNR}
    \label{tab:across_layer_fusion_on_hdbscan_high-speed}
\end{table*}
\begin{table*}
    \centering
    \input{supplemental/tables/ood/table_across_layer_fusion_on_hdbscan_average_over_scenarios}
    \caption{Average OOD detection performance over high delay spread and UE speed scenarios using HDBSCAN classifiers with means and the Euclidean distance metric, fused across layers. A: B1-PRE, B: B3-POST, C: OUT-POST, D: SNR}
    \label{tab:across_layer_fusion_on_hdbscan_average_over_scenarios}
\end{table*}
\begin{table*}
    \centering
    \input{supplemental/tables/ood/table_across_layer_fusion_on_knn_high-delay}
    \caption{OOD detection performance under high delay spread using k-NN classifiers with means and the cosine distance metric, fused across layers. A: B1-PRE, B: B3-POST, C: OUT-POST, D: SNR}
    \label{tab:across_layer_fusion_on_knn_high-delay}
\end{table*}
\begin{table*}
    \centering
    \input{supplemental/tables/ood/table_across_layer_fusion_on_knn_high-speed}
    \caption{OOD detection performance under high UE speed using k-NN classifiers with means and the cosine distance metric, fused across layers. A: B1-PRE, B: B3-POST, C: OUT-POST, D: SNR}
    \label{tab:across_layer_fusion_on_knn_high-speed}
\end{table*}
\begin{table*}
    \centering
    \input{supplemental/tables/ood/table_across_layer_fusion_on_knn_average_over_scenarios}
    \caption{Average OOD detection performance over high delay spread and UE speed scenarios using k-NN classifiers with means and the cosine distance metric, fused across layers. A: B1-PRE, B: B3-POST, C: OUT-POST, D: SNR}
    \label{tab:across_layer_fusion_on_knn_average_over_scenarios}
\end{table*}
\begin{table*}
    \centering
    \input{supplemental/tables/ood/table_across_layer_fusion_on_ocsvm_activations_high-delay}
    \caption{OOD detection performance under high delay spread using OC-SVM classifiers with means, fused across layers. A: B1-PRE, B: B3-POST, C: OUT-POST, D: SNR}
    \label{tab:across_layer_fusion_on_ocsvm_high-delay}
\end{table*}
\begin{table*}
    \centering
    \input{supplemental/tables/ood/table_across_layer_fusion_on_ocsvm_activations_high-speed}
    \caption{OOD detection performance under high UE speed using OC-SVM classifiers with means, fused across layers. A: B1-PRE, B: B3-POST, C: OUT-POST, D: SNR}
    \label{tab:across_layer_fusion_on_ocsvm_high-speed}
\end{table*}
\begin{table*}
    \centering
    \input{supplemental/tables/ood/table_across_layer_fusion_on_ocsvm_activations_average_over_scenarios}
    \caption{Average OOD detection performance over high delay spread and UE speed scenarios using OC-SVM classifiers with means, fused across layers. A: B1-PRE, B: B3-POST, C: OUT-POST, D: SNR}
    \label{tab:across_layer_fusion_on_ocsvm_average_over_scenarios}
\end{table*}
\begin{table*}
    \centering
    \input{supplemental/tables/ood/table_across_layer_fusion_on_mahalanobis_activations_high-delay}
    \caption{OOD detection performance under high delay spread using Mahalanobis classifiers with means, fused across layers. A: B1-PRE, B: B3-POST, C: OUT-POST, D: SNR}
    \label{tab:across_layer_fusion_on_mahalanobis_high-delay}
\end{table*}
\begin{table*}
    \centering
    \input{supplemental/tables/ood/table_across_layer_fusion_on_mahalanobis_activations_high-speed}
    \caption{OOD detection performance under high UE speed using Mahalanobis classifiers with means, fused across layers. A: B1-PRE, B: B3-POST, C: OUT-POST, D: SNR}
    \label{tab:across_layer_fusion_on_mahalanobis_high-speed}
\end{table*}
\begin{table*}
    \centering
    \input{supplemental/tables/ood/table_across_layer_fusion_on_mahalanobis_activations_average_over_scenarios}
    \caption{Average OOD detection performance over high delay spread and UE speed scenarios using Mahalanobis classifiers with means, fused across layers. A: B1-PRE, B: B3-POST, C: OUT-POST, D: SNR}
    \label{tab:across_layer_fusion_on_mahalanobis_average_over_scenarios}
\end{table*}
\begin{table*}
    \centering
    \input{supplemental/tables/ood/table_across_layer_fusion_on_within_layer_fusions_high-delay}
    \caption{OOD detection performance under high delay spread using within-layer fused classifiers, fused across layers. A: B1-PRE, B: B3-POST, C: OUT-POST, D: SNR}
    \label{tab:across_layer_fusion_on_within_layer_fusions_high-delay}
\end{table*}
\begin{table*}
    \centering
    \input{supplemental/tables/ood/table_across_layer_fusion_on_within_layer_fusions_high-speed}
    \caption{OOD detection performance under high UE speed using within-layer fused classifiers, fused across layers. A: B1-PRE, B: B3-POST, C: OUT-POST, D: SNR}
    \label{tab:across_layer_fusion_on_within_layer_fusions_high-speed}
\end{table*}
\begin{table*}
    \centering
    \input{supplemental/tables/ood/table_across_layer_fusion_on_within_layer_fusions_average_over_scenarios}
    \caption{Average OOD detection performance over high delay spread and UE speed scenarios using within-layer fused classifiers, fused across layers. A: B1-PRE, B: B3-POST, C: OUT-POST, D: SNR}
    \label{tab:across_layer_fusion_on_within_layer_fusions_average_over_scenarios}
\end{table*}

%% file: supplemental/tables/ood/table_single_model_baselines_on_b1-pre_high-delay.tex
\begin{tabular}{lllllllll}
\toprule
\multirow{2}{*}{Feature + Distance} & \multicolumn{2}{c}{HDBSCAN} & \multicolumn{2}{c}{k-NN} & \multicolumn{2}{c}{OC-SVM} & \multicolumn{2}{c}{Mahalanobis} \\
{} &            AUROC &            FPR95 &            AUROC &            FPR95 &                             AUROC &                             FPR95 &                             AUROC &                             FPR95 \\
\midrule
Mean + Euclidean     &  \textbf{0.9209} &  \textbf{1.0000} &           0.9741 &           0.1812 &  \multirow{2}{*}{\textbf{0.9394}} &  \multirow{2}{*}{\textbf{0.2084}} &  \multirow{2}{*}{\textbf{0.9965}} &  \multirow{2}{*}{\textbf{0.0200}} \\
Mean + Cosine        &           0.7360 &  \textbf{1.0000} &           0.9931 &  \textbf{0.0218} &                                   &                                   &                                   &                                   \\
KDE + Euclidean      &           0.9011 &  \textbf{1.0000} &           0.9930 &           0.0456 &           \multirow{3}{*}{0.8628} &           \multirow{3}{*}{0.4197} &           \multirow{3}{*}{0.9854} &           \multirow{3}{*}{0.0582} \\
KDE + Cosine         &           0.8920 &  \textbf{1.0000} &           0.9920 &           0.0526 &                                   &                                   &                                   &                                   \\
KDE + Jensen-Shannon &           0.8898 &  \textbf{1.0000} &  \textbf{0.9940} &           0.0374 &                                   &                                   &                                   &                                   \\
Energy + Euclidean   &           0.8586 &  \textbf{1.0000} &           0.9622 &           0.2538 &           \multirow{2}{*}{0.8861} &           \multirow{2}{*}{0.3286} &           \multirow{2}{*}{0.9905} &           \multirow{2}{*}{0.0458} \\
\bottomrule
\end{tabular}

%% file: supplemental/tables/ood/table_single_model_baselines_on_b1-pre_high-speed.tex
\begin{tabular}{lllllllll}
\toprule
\multirow{2}{*}{Feature + Distance} & \multicolumn{2}{c}{HDBSCAN} & \multicolumn{2}{c}{k-NN} & \multicolumn{2}{c}{OC-SVM} & \multicolumn{2}{c}{Mahalanobis} \\
{} &            AUROC &            FPR95 &            AUROC &            FPR95 &                             AUROC &                             FPR95 &                             AUROC &                             FPR95 \\
\midrule
Mean + Euclidean     &           0.6528 &  \textbf{1.0000} &           0.6670 &           0.9301 &           \multirow{2}{*}{0.5994} &           \multirow{2}{*}{0.9253} &           \multirow{2}{*}{0.7351} &           \multirow{2}{*}{0.8569} \\
Mean + Cosine        &           0.6285 &  \textbf{1.0000} &  \textbf{0.7339} &           0.8620 &                                   &                                   &                                   &                                   \\
KDE + Euclidean      &           0.6601 &  \textbf{1.0000} &           0.7178 &           0.8199 &  \multirow{3}{*}{\textbf{0.6143}} &  \multirow{3}{*}{\textbf{0.8947}} &  \multirow{3}{*}{\textbf{0.7357}} &  \multirow{3}{*}{\textbf{0.7864}} \\
KDE + Cosine         &  \textbf{0.6791} &  \textbf{1.0000} &           0.7164 &           0.8292 &                                   &                                   &                                   &                                   \\
KDE + Jensen-Shannon &           0.6715 &  \textbf{1.0000} &           0.7266 &  \textbf{0.8069} &                                   &                                   &                                   &                                   \\
Energy + Euclidean   &           0.6369 &  \textbf{1.0000} &           0.6691 &           0.9203 &           \multirow{2}{*}{0.6004} &           \multirow{2}{*}{0.9218} &           \multirow{2}{*}{0.7179} &           \multirow{2}{*}{0.8716} \\
\bottomrule
\end{tabular}

%% file: supplemental/tables/ood/table_single_model_baselines_on_b3-post_high-delay.tex
\begin{tabular}{lllllllll}
\toprule
\multirow{2}{*}{Feature + Distance} & \multicolumn{2}{c}{HDBSCAN} & \multicolumn{2}{c}{k-NN} & \multicolumn{2}{c}{OC-SVM} & \multicolumn{2}{c}{Mahalanobis} \\
{} &            AUROC &            FPR95 &            AUROC &            FPR95 &                             AUROC &                             FPR95 &                             AUROC &                             FPR95 \\
\midrule
Mean + Euclidean     &           0.8370 &           0.3543 &           0.9134 &           0.4588 &           \multirow{2}{*}{0.7372} &           \multirow{2}{*}{0.8597} &           \multirow{2}{*}{0.9410} &           \multirow{2}{*}{0.3327} \\
Mean + Cosine        &           0.8807 &  \textbf{0.2073} &           0.9174 &           0.6440 &                                   &                                   &                                   &                                   \\
KDE + Euclidean      &           0.8812 &           1.0000 &           0.9323 &           0.3689 &  \multirow{3}{*}{\textbf{0.8845}} &  \multirow{3}{*}{\textbf{0.5919}} &           \multirow{3}{*}{0.9389} &           \multirow{3}{*}{0.3576} \\
KDE + Cosine         &  \textbf{0.8896} &           0.3904 &           0.9395 &  \textbf{0.2397} &                                   &                                   &                                   &                                   \\
KDE + Jensen-Shannon &           0.8591 &           1.0000 &  \textbf{0.9501} &           0.3023 &                                   &                                   &                                   &                                   \\
Energy + Euclidean   &           0.7948 &           0.6573 &           0.9191 &           0.3872 &           \multirow{2}{*}{0.7569} &           \multirow{2}{*}{0.8255} &  \multirow{2}{*}{\textbf{0.9450}} &  \multirow{2}{*}{\textbf{0.3299}} \\
\bottomrule
\end{tabular}

%% file: supplemental/tables/ood/table_single_model_baselines_on_b3-post_high-speed.tex
\begin{tabular}{lllllllll}
\toprule
\multirow{2}{*}{Feature + Distance} & \multicolumn{2}{c}{HDBSCAN} & \multicolumn{2}{c}{k-NN} & \multicolumn{2}{c}{OC-SVM} & \multicolumn{2}{c}{Mahalanobis} \\
{} &            AUROC &            FPR95 &            AUROC &            FPR95 &                             AUROC &                             FPR95 &                             AUROC &                             FPR95 \\
\midrule
Mean + Euclidean     &           0.6955 &           0.5966 &           0.7317 &           0.8488 &  \multirow{2}{*}{\textbf{0.7387}} &           \multirow{2}{*}{0.8957} &  \multirow{2}{*}{\textbf{0.7472}} &           \multirow{2}{*}{0.8615} \\
Mean + Cosine        &  \textbf{0.7249} &  \textbf{0.5206} &           0.7522 &           0.9138 &                                   &                                   &                                   &                                   \\
KDE + Euclidean      &           0.6993 &           1.0000 &           0.7547 &           0.8202 &           \multirow{3}{*}{0.7344} &  \multirow{3}{*}{\textbf{0.8538}} &           \multirow{3}{*}{0.7385} &           \multirow{3}{*}{0.9170} \\
KDE + Cosine         &           0.7210 &           0.6114 &  \textbf{0.7820} &  \textbf{0.7403} &                                   &                                   &                                   &                                   \\
KDE + Jensen-Shannon &           0.7053 &           1.0000 &           0.7704 &           0.7839 &                                   &                                   &                                   &                                   \\
Energy + Euclidean   &           0.6864 &           0.7691 &           0.7590 &           0.8194 &           \multirow{2}{*}{0.7279} &           \multirow{2}{*}{0.9121} &           \multirow{2}{*}{0.7411} &  \multirow{2}{*}{\textbf{0.8559}} \\
\bottomrule
\end{tabular}

%% file: supplemental/tables/ood/table_single_model_baselines_on_b3-post_average_over_scenarios.tex
\begin{tabular}{lllllllll}
\toprule
\multirow{2}{*}{Feature + Distance} & \multicolumn{2}{c}{HDBSCAN} & \multicolumn{2}{c}{k-NN} & \multicolumn{2}{c}{OC-SVM} & \multicolumn{2}{c}{Mahalanobis} \\
{} &            AUROC &            FPR95 &            AUROC &            FPR95 &                             AUROC &                             FPR95 &                             AUROC &                             FPR95 \\
\midrule
Mean + Euclidean     &           0.7662 &           0.4754 &           0.8225 &           0.6538 &           \multirow{2}{*}{0.7379} &           \multirow{2}{*}{0.8777} &  \multirow{2}{*}{\textbf{0.8441}} &           \multirow{2}{*}{0.5971} \\
Mean + Cosine        &           0.8028 &  \textbf{0.3640} &           0.8348 &           0.7789 &                                   &                                   &                                   &                                   \\
KDE + Euclidean      &           0.7902 &           1.0000 &           0.8435 &           0.5945 &  \multirow{3}{*}{\textbf{0.8094}} &  \multirow{3}{*}{\textbf{0.7229}} &           \multirow{3}{*}{0.8387} &           \multirow{3}{*}{0.6373} \\
KDE + Cosine         &  \textbf{0.8053} &           0.5009 &  \textbf{0.8608} &  \textbf{0.4900} &                                   &                                   &                                   &                                   \\
KDE + Jensen-Shannon &           0.7822 &           1.0000 &           0.8603 &           0.5431 &                                   &                                   &                                   &                                   \\
Energy + Euclidean   &           0.7406 &           0.7132 &           0.8391 &           0.6033 &           \multirow{2}{*}{0.7424} &           \multirow{2}{*}{0.8688} &           \multirow{2}{*}{0.8431} &  \multirow{2}{*}{\textbf{0.5929}} \\
\bottomrule
\end{tabular}

%% file: supplemental/tables/ood/table_single_model_baselines_on_b6-post_high-delay.tex
\begin{tabular}{lllllllll}
\toprule
\multirow{2}{*}{Feature + Distance} & \multicolumn{2}{c}{HDBSCAN} & \multicolumn{2}{c}{k-NN} & \multicolumn{2}{c}{OC-SVM} & \multicolumn{2}{c}{Mahalanobis} \\
{} &            AUROC &            FPR95 &            AUROC &            FPR95 &                             AUROC &                             FPR95 &                             AUROC &                             FPR95 \\
\midrule
Mean + Euclidean     &           0.5208 &  \textbf{1.0000} &           0.5769 &           0.9495 &           \multirow{2}{*}{0.5509} &           \multirow{2}{*}{0.9486} &           \multirow{2}{*}{0.5677} &  \multirow{2}{*}{\textbf{0.9359}} \\
Mean + Cosine        &  \textbf{0.6745} &  \textbf{1.0000} &  \textbf{0.6929} &  \textbf{0.8847} &                                   &                                   &                                   &                                   \\
KDE + Euclidean      &           0.5692 &  \textbf{1.0000} &           0.6397 &           0.9485 &           \multirow{3}{*}{0.4978} &           \multirow{3}{*}{0.9746} &  \multirow{3}{*}{\textbf{0.6051}} &           \multirow{3}{*}{0.9499} \\
KDE + Cosine         &           0.5574 &  \textbf{1.0000} &           0.6584 &           0.9484 &                                   &                                   &                                   &                                   \\
KDE + Jensen-Shannon &           0.5741 &  \textbf{1.0000} &           0.6698 &           0.9466 &                                   &                                   &                                   &                                   \\
Energy + Euclidean   &           0.5195 &  \textbf{1.0000} &           0.5816 &           0.9495 &  \multirow{2}{*}{\textbf{0.5564}} &  \multirow{2}{*}{\textbf{0.9455}} &           \multirow{2}{*}{0.5697} &           \multirow{2}{*}{0.9481} \\
\bottomrule
\end{tabular}

%% file: supplemental/tables/ood/table_single_model_baselines_on_b6-post_high-speed.tex
\begin{tabular}{lllllllll}
\toprule
\multirow{2}{*}{Feature + Distance} & \multicolumn{2}{c}{HDBSCAN} & \multicolumn{2}{c}{k-NN} & \multicolumn{2}{c}{OC-SVM} & \multicolumn{2}{c}{Mahalanobis} \\
{} &            AUROC &            FPR95 &            AUROC &            FPR95 &                             AUROC &                             FPR95 &                             AUROC &                             FPR95 \\
\midrule
Mean + Euclidean     &           0.5096 &  \textbf{1.0000} &           0.5303 &           0.9565 &           \multirow{2}{*}{0.4458} &           \multirow{2}{*}{0.9919} &           \multirow{2}{*}{0.5204} &           \multirow{2}{*}{0.9465} \\
Mean + Cosine        &  \textbf{0.6372} &  \textbf{1.0000} &  \textbf{0.6620} &  \textbf{0.9235} &                                   &                                   &                                   &                                   \\
KDE + Euclidean      &           0.5470 &  \textbf{1.0000} &           0.6039 &           0.9536 &  \multirow{3}{*}{\textbf{0.4606}} &  \multirow{3}{*}{\textbf{0.9813}} &  \multirow{3}{*}{\textbf{0.5719}} &           \multirow{3}{*}{0.9505} \\
KDE + Cosine         &           0.5418 &  \textbf{1.0000} &           0.6182 &           0.9546 &                                   &                                   &                                   &                                   \\
KDE + Jensen-Shannon &           0.5499 &  \textbf{1.0000} &           0.6187 &           0.9516 &                                   &                                   &                                   &                                   \\
Energy + Euclidean   &           0.5086 &  \textbf{1.0000} &           0.5279 &           0.9570 &           \multirow{2}{*}{0.4446} &           \multirow{2}{*}{0.9927} &           \multirow{2}{*}{0.5147} &  \multirow{2}{*}{\textbf{0.9458}} \\
\bottomrule
\end{tabular}

%% file: supplemental/tables/ood/table_single_model_baselines_on_b6-post_average_over_scenarios.tex
\begin{tabular}{lllllllll}
\toprule
\multirow{2}{*}{Feature + Distance} & \multicolumn{2}{c}{HDBSCAN} & \multicolumn{2}{c}{k-NN} & \multicolumn{2}{c}{OC-SVM} & \multicolumn{2}{c}{Mahalanobis} \\
{} &            AUROC &            FPR95 &            AUROC &            FPR95 &                             AUROC &                             FPR95 &                             AUROC &                             FPR95 \\
\midrule
Mean + Euclidean     &           0.5152 &  \textbf{1.0000} &           0.5536 &           0.9530 &           \multirow{2}{*}{0.4983} &           \multirow{2}{*}{0.9703} &           \multirow{2}{*}{0.5441} &  \multirow{2}{*}{\textbf{0.9412}} \\
Mean + Cosine        &  \textbf{0.6558} &  \textbf{1.0000} &  \textbf{0.6775} &  \textbf{0.9041} &                                   &                                   &                                   &                                   \\
KDE + Euclidean      &           0.5581 &  \textbf{1.0000} &           0.6218 &           0.9510 &           \multirow{3}{*}{0.4792} &           \multirow{3}{*}{0.9779} &  \multirow{3}{*}{\textbf{0.5885}} &           \multirow{3}{*}{0.9502} \\
KDE + Cosine         &           0.5496 &  \textbf{1.0000} &           0.6383 &           0.9515 &                                   &                                   &                                   &                                   \\
KDE + Jensen-Shannon &           0.5620 &  \textbf{1.0000} &           0.6443 &           0.9491 &                                   &                                   &                                   &                                   \\
Energy + Euclidean   &           0.5140 &  \textbf{1.0000} &           0.5547 &           0.9532 &  \multirow{2}{*}{\textbf{0.5005}} &  \multirow{2}{*}{\textbf{0.9691}} &           \multirow{2}{*}{0.5422} &           \multirow{2}{*}{0.9470} \\
\bottomrule
\end{tabular}

%% file: supplemental/tables/ood/table_snr_regimes_on_b1-pre_high-delay.tex
\begin{tabular}{lllllllllllll}
\toprule
\multirow{2}{*}{SNR regime} & \multicolumn{4}{c}{k-NN} & \multicolumn{4}{c}{OC-SVM} & \multicolumn{4}{c}{Mahalanobis} \\
{} &            AUROC &    $\Delta$AUROC &            FPR95 &    $\Delta$FPR95 &            AUROC &    $\Delta$AUROC &            FPR95 &    $\Delta$FPR95 &            AUROC &    $\Delta$AUROC &            FPR95 &    $\Delta$FPR95 \\
\midrule
Activations only            &  \textbf{0.9931} &  \textbf{0.0000} &  \textbf{0.0218} &  \textbf{0.0000} &  \textbf{0.9394} &  \textbf{0.0000} &  \textbf{0.2084} &  \textbf{0.0000} &  \textbf{0.9965} &  \textbf{0.0000} &  \textbf{0.0200} &  \textbf{0.0000} \\
SNR only                    &           0.7326 &          -0.2605 &           0.7897 &           0.7679 &           0.5000 &          -0.4394 &           1.0000 &           0.7916 &           0.5000 &          -0.4965 &           0.9589 &           0.9389 \\
Both combined (hard/concat) &           0.9926 &          -0.0005 &           0.0240 &           0.0021 &           0.9392 &          -0.0002 &           0.2096 &           0.0012 &           0.9696 &          -0.0269 &           0.1484 &           0.1285 \\
Both combined (soft)        &           0.9922 &          -0.0009 &           0.0278 &           0.0060 &                  &                  &                  &                  &                  &                  &                  &                  \\
\bottomrule
\end{tabular}

%% file: supplemental/tables/ood/table_snr_regimes_on_b1-pre_high-speed.tex
\begin{tabular}{lllllllllllll}
\toprule
\multirow{2}{*}{SNR regime} & \multicolumn{4}{c}{k-NN} & \multicolumn{4}{c}{OC-SVM} & \multicolumn{4}{c}{Mahalanobis} \\
{} &            AUROC &    $\Delta$AUROC &            FPR95 &    $\Delta$FPR95 &            AUROC &    $\Delta$AUROC &            FPR95 &    $\Delta$FPR95 &            AUROC &    $\Delta$AUROC &            FPR95 &    $\Delta$FPR95 \\
\midrule
Activations only            &           0.7339 &          -0.0019 &  \textbf{0.8620} &  \textbf{0.0000} &           0.5994 &          -0.0000 &           0.9253 &           0.0001 &  \textbf{0.7351} &  \textbf{0.0000} &  \textbf{0.8569} &  \textbf{0.0000} \\
SNR only                    &           0.6128 &          -0.1230 &           0.9144 &           0.0524 &           0.5000 &          -0.0994 &           1.0000 &           0.0748 &           0.5000 &          -0.2351 &           0.9589 &           0.1020 \\
Both combined (hard/concat) &           0.7326 &          -0.0032 &           0.8660 &           0.0040 &  \textbf{0.5994} &  \textbf{0.0000} &  \textbf{0.9252} &  \textbf{0.0000} &           0.6527 &          -0.0823 &           0.8987 &           0.0418 \\
Both combined (soft)        &  \textbf{0.7358} &  \textbf{0.0000} &           0.8651 &           0.0031 &                  &                  &                  &                  &                  &                  &                  &                  \\
\bottomrule
\end{tabular}

%% file: supplemental/tables/ood/table_snr_regimes_on_b3-post_high-delay.tex
\begin{tabular}{lllllllllllll}
\toprule
\multirow{2}{*}{SNR regime} & \multicolumn{4}{c}{k-NN} & \multicolumn{4}{c}{OC-SVM} & \multicolumn{4}{c}{Mahalanobis} \\
{} &            AUROC &    $\Delta$AUROC &            FPR95 &    $\Delta$FPR95 &            AUROC &    $\Delta$AUROC &            FPR95 &    $\Delta$FPR95 &            AUROC &    $\Delta$AUROC &            FPR95 &    $\Delta$FPR95 \\
\midrule
Activations only            &           0.9174 &          -0.0042 &           0.6440 &           0.0433 &           0.7372 &          -0.0006 &           0.8597 &           0.0060 &  \textbf{0.9410} &  \textbf{0.0000} &  \textbf{0.3327} &  \textbf{0.0000} \\
SNR only                    &           0.6758 &          -0.2458 &           0.8233 &           0.2226 &           0.5000 &          -0.2378 &           1.0000 &           0.1463 &           0.5000 &          -0.4410 &           0.9589 &           0.6261 \\
Both combined (hard/concat) &  \textbf{0.9216} &  \textbf{0.0000} &  \textbf{0.6007} &  \textbf{0.0000} &  \textbf{0.7378} &  \textbf{0.0000} &  \textbf{0.8537} &  \textbf{0.0000} &           0.9010 &          -0.0400 &           0.3724 &           0.0396 \\
\bottomrule
\end{tabular}

%% file: supplemental/tables/ood/table_snr_regimes_on_b3-post_high-speed.tex
\begin{tabular}{lllllllllllll}
\toprule
\multirow{2}{*}{SNR regime} & \multicolumn{4}{c}{k-NN} & \multicolumn{4}{c}{OC-SVM} & \multicolumn{4}{c}{Mahalanobis} \\
{} &            AUROC &    $\Delta$AUROC &            FPR95 &    $\Delta$FPR95 &            AUROC &    $\Delta$AUROC &            FPR95 &    $\Delta$FPR95 &            AUROC &    $\Delta$AUROC &            FPR95 &    $\Delta$FPR95 \\
\midrule
Activations only            &           0.7522 &          -0.0089 &           0.9138 &           0.0193 &           0.7387 &          -0.0000 &  \textbf{0.8957} &  \textbf{0.0000} &           0.7472 &          -0.0004 &           0.8615 &           0.0463 \\
SNR only                    &           0.6797 &          -0.0814 &  \textbf{0.8944} &  \textbf{0.0000} &           0.5000 &          -0.2387 &           1.0000 &           0.1043 &           0.5000 &          -0.2475 &           0.9589 &           0.1437 \\
Both combined (hard/concat) &  \textbf{0.7611} &  \textbf{0.0000} &           0.9015 &           0.0071 &  \textbf{0.7387} &  \textbf{0.0000} &           0.8979 &           0.0022 &  \textbf{0.7475} &  \textbf{0.0000} &  \textbf{0.8152} &  \textbf{0.0000} \\
\bottomrule
\end{tabular}

%% file: supplemental/tables/ood/table_snr_regimes_on_b6-post_high-delay.tex
\begin{tabular}{lllllllllllll}
\toprule
\multirow{2}{*}{SNR regime} & \multicolumn{4}{c}{k-NN} & \multicolumn{4}{c}{OC-SVM} & \multicolumn{4}{c}{Mahalanobis} \\
{} &            AUROC &    $\Delta$AUROC &            FPR95 &    $\Delta$FPR95 &            AUROC &    $\Delta$AUROC &            FPR95 &    $\Delta$FPR95 &            AUROC &    $\Delta$AUROC &            FPR95 &    $\Delta$FPR95 \\
\midrule
Activations only            &  \textbf{0.6929} &  \textbf{0.0000} &  \textbf{0.8847} &  \textbf{0.0000} &           0.5509 &          -0.0121 &           0.9486 &           0.0362 &           0.5677 &          -0.0313 &           0.9359 &           0.0501 \\
SNR only                    &           0.5373 &          -0.1556 &           0.9309 &           0.0462 &           0.5000 &          -0.0630 &           1.0000 &           0.0876 &           0.5000 &          -0.0990 &           0.9589 &           0.0731 \\
Both combined (hard/concat) &           0.6905 &          -0.0024 &           0.8952 &           0.0105 &  \textbf{0.5630} &  \textbf{0.0000} &  \textbf{0.9124} &  \textbf{0.0000} &  \textbf{0.5990} &  \textbf{0.0000} &  \textbf{0.8858} &  \textbf{0.0000} \\
\bottomrule
\end{tabular}

%% file: supplemental/tables/ood/table_snr_regimes_on_b6-post_high-speed.tex
\begin{tabular}{lllllllllllll}
\toprule
\multirow{2}{*}{SNR regime} & \multicolumn{4}{c}{k-NN} & \multicolumn{4}{c}{OC-SVM} & \multicolumn{4}{c}{Mahalanobis} \\
{} &            AUROC &    $\Delta$AUROC &            FPR95 &    $\Delta$FPR95 &            AUROC &    $\Delta$AUROC &            FPR95 &    $\Delta$FPR95 &            AUROC &    $\Delta$AUROC &            FPR95 &    $\Delta$FPR95 \\
\midrule
Activations only            &           0.6620 &          -0.0235 &           0.9235 &           0.0227 &           0.4458 &          -0.0880 &           0.9919 &           0.0183 &           0.5204 &          -0.0407 &           0.9465 &           0.0233 \\
SNR only                    &           0.6488 &          -0.0367 &  \textbf{0.9008} &  \textbf{0.0000} &           0.5000 &          -0.0338 &           1.0000 &           0.0263 &           0.5000 &          -0.0611 &           0.9589 &           0.0357 \\
Both combined (hard/concat) &  \textbf{0.6855} &  \textbf{0.0000} &           0.9238 &           0.0230 &  \textbf{0.5338} &  \textbf{0.0000} &  \textbf{0.9737} &  \textbf{0.0000} &  \textbf{0.5611} &  \textbf{0.0000} &  \textbf{0.9232} &  \textbf{0.0000} \\
\bottomrule
\end{tabular}

%% file: supplemental/tables/ood/best_fusions_within_layer_scenario_average.tex
\begin{tabular}{lll}
\toprule
Layer & OC-SVM fusion & BH-style fusion \\
\midrule
B1-PRE (Table~\ref{tab:single_model_within_layer_fusions_on_b1-pre_average_over_scenarios}) & k-NN, OC-SVM, SNR & k-NN, Mahalanobis \\
B3-POST (Table~\ref{tab:single_model_within_layer_fusions_on_b3-post_average_over_scenarios}) & k-NN, OC-SVM, Mahalanobis, SNR & k-NN, OC-SVM, Mahalanobis, SNR \\
OUT-POST (Table~\ref{tab:single_model_within_layer_fusions_on_b6-post_average_over_scenarios}) & HDBSCAN, Mahalanobis, SNR & HDBSCAN, k-NN \\
\bottomrule
\end{tabular}

%% file: supplemental/tables/ood/table_classifier_fusion_on_b1-pre_high-delay.tex
\begin{tabular}{lllllllll}
\toprule
\multirow{2}{*}{\shortstack{Baseline\\classifiers fused}} & \multicolumn{4}{c}{One-Class SVM} & \multicolumn{4}{c}{Conformal P-value based approach} \\
{} &            AUROC &    $\Delta$AUROC &            FPR95 &    $\Delta$FPR95 &                            AUROC &    $\Delta$AUROC &            FPR95 &    $\Delta$FPR95 \\
\midrule
\{A, B, C, D, E\} &           0.9961 &          -0.0004 &           0.0122 &           0.0022 &                           0.9952 &          -0.0005 &           0.0265 &           0.0065 \\
\{B, C, D, E\}    &           0.9966 &          -0.0000 &           0.0188 &           0.0088 &                           0.9954 &          -0.0004 &           0.0265 &           0.0065 \\
\{A, C, D, E\}    &           0.9945 &          -0.0020 &           0.0233 &           0.0133 &                           0.9937 &          -0.0020 &           0.0348 &           0.0148 \\
\{A, B, D, E\}    &           0.9954 &          -0.0012 &  \textbf{0.0100} &  \textbf{0.0000} &                           0.9944 &          -0.0013 &           0.0237 &           0.0037 \\
\{A, B, C, E\}    &           0.9952 &          -0.0013 &           0.0167 &           0.0067 &                           0.9941 &          -0.0016 &           0.0304 &           0.0104 \\
\{A, B, C, D\}    &           0.9959 &          -0.0006 &           0.0117 &           0.0017 &                           0.9956 &          -0.0002 &           0.0249 &           0.0049 \\
\bottomrule
\end{tabular}

%% file: supplemental/tables/ood/table_classifier_fusion_on_b1-pre_high-speed.tex
\begin{tabular}{lllllllll}
\toprule
\multirow{2}{*}{\shortstack{Baseline\\classifiers fused}} & \multicolumn{4}{c}{One-Class SVM} & \multicolumn{4}{c}{Conformal P-value based approach} \\
{} &            AUROC &    $\Delta$AUROC &            FPR95 &    $\Delta$FPR95 &                            AUROC &    $\Delta$AUROC &            FPR95 &    $\Delta$FPR95 \\
\midrule
\{A, B, C, D, E\} &           0.7471 &          -0.0079 &           0.8334 &           0.0150 &                           0.7288 &          -0.0121 &           0.8729 &           0.0166 \\
\{B, C, D, E\}    &           0.7525 &          -0.0025 &           0.8305 &           0.0121 &                           0.7330 &          -0.0079 &           0.8803 &           0.0240 \\
\{A, C, D, E\}    &           0.7159 &          -0.0391 &           0.8995 &           0.0811 &                           0.7101 &          -0.0309 &           1.0000 &           0.1437 \\
\{A, B, D, E\}    &           0.7440 &          -0.0110 &           0.8628 &           0.0444 &                           0.7259 &          -0.0150 &           0.8958 &           0.0395 \\
\{A, B, C, E\}    &           0.7503 &          -0.0047 &           0.8196 &           0.0013 &                           0.7241 &          -0.0168 &           0.8891 &           0.0328 \\
\{A, B, C, D\}    &           0.7298 &          -0.0252 &           0.8664 &           0.0480 &                           0.7309 &          -0.0101 &           0.8680 &           0.0117 \\
\bottomrule
\end{tabular}

%% file: supplemental/tables/ood/table_within_layer_fusions_on_b1-pre_average_over_scenarios.tex
\begin{tabular}{lllllllll}
\toprule
\multirow{2}{*}{\shortstack{Baseline\\classifiers fused}} & \multicolumn{4}{c}{One-Class SVM} & \multicolumn{4}{c}{Conformal P-value based approach} \\
{} &            AUROC &    $\Delta$AUROC &            FPR95 &    $\Delta$FPR95 &                            AUROC &    $\Delta$AUROC &            FPR95 &    $\Delta$FPR95 \\
\midrule
\{A, B, C, D, E\} &           0.8716 &          -0.0039 &           0.4228 &           0.0046 &                           0.8620 &          -0.0061 &           0.4497 &           0.0111 \\
\{B, C, D, E\}    &           0.8745 &          -0.0010 &           0.4246 &           0.0065 &                           0.8642 &          -0.0039 &           0.4534 &           0.0147 \\
\{A, C, D, E\}    &           0.8552 &          -0.0202 &           0.4614 &           0.0432 &                           0.8519 &          -0.0162 &           0.5174 &           0.0787 \\
\{A, B, D, E\}    &           0.8697 &          -0.0058 &           0.4364 &           0.0182 &                           0.8602 &          -0.0079 &           0.4597 &           0.0211 \\
\{A, B, C, E\}    &           0.8728 &          -0.0027 &  \textbf{0.4182} &  \textbf{0.0000} &                           0.8591 &          -0.0090 &           0.4597 &           0.0211 \\
\{A, B, C, D\}    &           0.8629 &          -0.0126 &           0.4390 &           0.0209 &                           0.8632 &          -0.0049 &           0.4465 &           0.0078 \\
\bottomrule
\end{tabular}

%% file: supplemental/tables/ood/table_classifier_fusion_on_b3-post_high-delay.tex
\begin{tabular}{lllllllll}
\toprule
\multirow{2}{*}{\shortstack{Baseline\\classifiers fused}} & \multicolumn{4}{c}{One-Class SVM} & \multicolumn{4}{c}{Conformal P-value based approach} \\
{} &            AUROC &    $\Delta$AUROC &            FPR95 &    $\Delta$FPR95 &                            AUROC &    $\Delta$AUROC &            FPR95 &    $\Delta$FPR95 \\
\midrule
\{A, B, C, D, E\} &           0.9232 &          -0.0104 &           0.5057 &           0.1210 &                           0.9296 &          -0.0043 &           0.4651 &           0.1302 \\
\{B, C, D, E\}    &           0.9263 &          -0.0073 &           0.4924 &           0.1077 &                           0.9307 &          -0.0032 &           0.4651 &           0.1302 \\
\{A, C, D, E\}    &           0.9191 &          -0.0146 &           0.4592 &           0.0744 &                           0.9188 &          -0.0150 &           0.4028 &           0.0679 \\
\{A, B, D, E\}    &           0.9162 &          -0.0175 &           0.6367 &           0.2520 &                           0.9263 &          -0.0076 &           0.5065 &           0.1716 \\
\{A, B, C, E\}    &           0.9210 &          -0.0127 &           0.5433 &           0.1586 &                           0.9238 &          -0.0101 &           0.5273 &           0.1924 \\
\{A, B, C, D\}    &           0.9234 &          -0.0103 &           0.5086 &           0.1239 &                           0.9303 &          -0.0036 &           0.4643 &           0.1294 \\
\bottomrule
\end{tabular}

%% file: supplemental/tables/ood/table_classifier_fusion_on_b3-post_high-speed.tex
\begin{tabular}{lllllllll}
\toprule
\multirow{2}{*}{\shortstack{Baseline\\classifiers fused}} & \multicolumn{4}{c}{One-Class SVM} & \multicolumn{4}{c}{Conformal P-value based approach} \\
{} &            AUROC &    $\Delta$AUROC &            FPR95 &    $\Delta$FPR95 &                            AUROC &    $\Delta$AUROC &            FPR95 &    $\Delta$FPR95 \\
\midrule
\{A, B, C, D, E\} &           0.7707 &          -0.0050 &           0.9332 &           0.0615 &                           0.7746 &          -0.0012 &           1.0000 &           0.1836 \\
\{B, C, D, E\}    &  \textbf{0.7757} &  \textbf{0.0000} &           0.9241 &           0.0525 &                  \textbf{0.7757} &  \textbf{0.0000} &           0.8279 &           0.0115 \\
\{A, C, D, E\}    &           0.7550 &          -0.0207 &           0.8815 &           0.0099 &                           0.7591 &          -0.0167 &           1.0000 &           0.1836 \\
\{A, B, D, E\}    &           0.7627 &          -0.0130 &           0.9623 &           0.0906 &                           0.7679 &          -0.0079 &           1.0000 &           0.1836 \\
\{A, B, C, E\}    &           0.7673 &          -0.0084 &           0.9210 &           0.0494 &                           0.7738 &          -0.0019 &           0.8627 &           0.0463 \\
\{A, B, C, D\}    &           0.7612 &          -0.0145 &           0.9168 &           0.0452 &                           0.7662 &          -0.0096 &           1.0000 &           0.1836 \\
\bottomrule
\end{tabular}

%% file: supplemental/tables/ood/table_within_layer_fusions_on_b3-post_average_over_scenarios.tex
\begin{tabular}{lllllllll}
\toprule
\multirow{2}{*}{\shortstack{Baseline\\classifiers fused}} & \multicolumn{4}{c}{One-Class SVM} & \multicolumn{4}{c}{Conformal P-value based approach} \\
{} &            AUROC &    $\Delta$AUROC &            FPR95 &    $\Delta$FPR95 &                            AUROC &    $\Delta$AUROC &            FPR95 &    $\Delta$FPR95 \\
\midrule
\{A, B, C, D, E\} &           0.8470 &          -0.0041 &           0.7194 &           0.0874 &                           0.8521 &          -0.0011 &           0.7325 &           0.1220 \\
\{B, C, D, E\}    &  \textbf{0.8510} &  \textbf{0.0000} &           0.7083 &           0.0762 &                  \textbf{0.8532} &  \textbf{0.0000} &           0.6465 &           0.0359 \\
\{A, C, D, E\}    &           0.8370 &          -0.0140 &           0.6703 &           0.0382 &                           0.8390 &          -0.0143 &           0.7014 &           0.0908 \\
\{A, B, D, E\}    &           0.8394 &          -0.0116 &           0.7995 &           0.1674 &                           0.8471 &          -0.0061 &           0.7532 &           0.1427 \\
\{A, B, C, E\}    &           0.8441 &          -0.0069 &           0.7322 &           0.1001 &                           0.8488 &          -0.0044 &           0.6950 &           0.0844 \\
\{A, B, C, D\}    &           0.8423 &          -0.0087 &           0.7127 &           0.0806 &                           0.8482 &          -0.0050 &           0.7322 &           0.1216 \\
\bottomrule
\end{tabular}

%% file: supplemental/tables/ood/table_classifier_fusion_on_b6-post_high-delay.tex
\begin{tabular}{lllllllll}
\toprule
\multirow{2}{*}{\shortstack{Baseline\\classifiers fused}} & \multicolumn{4}{c}{One-Class SVM} & \multicolumn{4}{c}{Conformal P-value based approach} \\
{} &            AUROC &    $\Delta$AUROC &            FPR95 &    $\Delta$FPR95 &                            AUROC &    $\Delta$AUROC &            FPR95 &    $\Delta$FPR95 \\
\midrule
\{A, B, C, D, E\} &           0.7268 &          -0.0485 &           0.8285 &           0.0794 &                           0.6425 &          -0.0514 &           0.9231 &           0.0436 \\
\{B, C, D, E\}    &           0.7229 &          -0.0524 &           0.8776 &           0.1286 &                           0.6421 &          -0.0518 &           0.9299 &           0.0503 \\
\{A, C, D, E\}    &           0.7346 &          -0.0407 &           0.8440 &           0.0949 &                           0.6364 &          -0.0574 &  \textbf{0.8795} &  \textbf{0.0000} \\
\{A, B, D, E\}    &           0.7578 &          -0.0175 &  \textbf{0.7491} &  \textbf{0.0000} &                           0.6583 &          -0.0355 &           0.9244 &           0.0449 \\
\{A, B, C, E\}    &           0.6160 &          -0.1593 &           0.9805 &           0.2314 &                           0.6315 &          -0.0624 &           0.9213 &           0.0418 \\
\{A, B, C, D\}    &           0.6441 &          -0.1312 &           0.8400 &           0.0910 &                           0.6194 &          -0.0745 &           0.9319 &           0.0523 \\
\bottomrule
\end{tabular}

%% file: supplemental/tables/ood/table_classifier_fusion_on_b6-post_high-speed.tex
\begin{tabular}{lllllllll}
\toprule
\multirow{2}{*}{\shortstack{Baseline\\classifiers fused}} & \multicolumn{4}{c}{One-Class SVM} & \multicolumn{4}{c}{Conformal P-value based approach} \\
{} &            AUROC &    $\Delta$AUROC &            FPR95 &    $\Delta$FPR95 &                            AUROC &    $\Delta$AUROC &            FPR95 &    $\Delta$FPR95 \\
\midrule
\{A, B, C, D, E\} &           0.6593 &          -0.0449 &           0.8953 &           0.0214 &                           0.6006 &          -0.0607 &           0.9239 &           0.0233 \\
\{B, C, D, E\}    &           0.6591 &          -0.0451 &           0.9224 &           0.0486 &                           0.6011 &          -0.0603 &           0.9197 &           0.0190 \\
\{A, C, D, E\}    &           0.6657 &          -0.0386 &           0.8884 &           0.0145 &                           0.5714 &          -0.0900 &           0.9070 &           0.0064 \\
\{A, B, D, E\}    &           0.6920 &          -0.0123 &  \textbf{0.8739} &  \textbf{0.0000} &                           0.6066 &          -0.0547 &           0.9294 &           0.0287 \\
\{A, B, C, E\}    &           0.6106 &          -0.0936 &           0.9777 &           0.1038 &                           0.6090 &          -0.0524 &           0.9182 &           0.0176 \\
\{A, B, C, D\}    &           0.5538 &          -0.1504 &           0.9286 &           0.0547 &                           0.5834 &          -0.0779 &           0.9272 &           0.0266 \\
\bottomrule
\end{tabular}

%% file: supplemental/tables/ood/table_within_layer_fusions_on_b6-post_average_over_scenarios.tex
\begin{tabular}{lllllllll}
\toprule
\multirow{2}{*}{\shortstack{Baseline\\classifiers fused}} & \multicolumn{4}{c}{One-Class SVM} & \multicolumn{4}{c}{Conformal P-value based approach} \\
{} &            AUROC &    $\Delta$AUROC &            FPR95 &    $\Delta$FPR95 &                            AUROC &    $\Delta$AUROC &            FPR95 &    $\Delta$FPR95 \\
\midrule
\{A, B, C, D, E\} &           0.6931 &          -0.0467 &           0.8619 &           0.0504 &                           0.6215 &          -0.0561 &           0.9235 &           0.0303 \\
\{B, C, D, E\}    &           0.6910 &          -0.0488 &           0.9000 &           0.0886 &                           0.6216 &          -0.0560 &           0.9248 &           0.0315 \\
\{A, C, D, E\}    &           0.7001 &          -0.0396 &           0.8662 &           0.0547 &                           0.6039 &          -0.0737 &  \textbf{0.8933} &  \textbf{0.0000} \\
\{A, B, D, E\}    &           0.7249 &          -0.0149 &  \textbf{0.8115} &  \textbf{0.0000} &                           0.6325 &          -0.0451 &           0.9269 &           0.0336 \\
\{A, B, C, E\}    &           0.6133 &          -0.1265 &           0.9791 &           0.1676 &                           0.6202 &          -0.0574 &           0.9198 &           0.0265 \\
\{A, B, C, D\}    &           0.5989 &          -0.1408 &           0.8843 &           0.0728 &                           0.6014 &          -0.0762 &           0.9295 &           0.0362 \\
\bottomrule
\end{tabular}

%% file: supplemental/tables/ood/best_fusions_across_layers_scenario_average.tex
\begin{tabular}{lll}
\toprule
Baseline method & OC-SVM fusion & BH-style fusion \\
\midrule
HDBSCAN (Table~\ref{tab:across_layer_fusion_on_hdbscan_average_over_scenarios}) & B1-PRE, B3-POST, OUT-POST, SNR & B1-PRE, B3-POST \\
k-NN (Table~\ref{tab:across_layer_fusion_on_knn_average_over_scenarios}) & B1-PRE, B3-POST, OUT-POST, SNR & B1-PRE, B3-POST, OUT-POST, SNR \\
OC-SVM (Table~\ref{tab:across_layer_fusion_on_ocsvm_average_over_scenarios}) & B1-PRE, B3-POST, SNR & B1-PRE, B3-POST \\
Mahalanobis (Table~\ref{tab:across_layer_fusion_on_mahalanobis_average_over_scenarios}) & B1-PRE, B3-POST, OUT-POST, SNR & B1-PRE, B3-POST \\
Within-layer (Table~\ref{tab:across_layer_fusion_on_within_layer_fusions_average_over_scenarios}) & B1-PRE, B3-POST, SNR & B1-PRE, B3-POST, OUT-POST \\
\bottomrule
\end{tabular}

%% file: supplemental/tables/ood/table_across_layer_fusion_on_hdbscan_high-delay.tex
\begin{tabular}{lllllllll}
\toprule
\multirow{2}{*}{Layers fused} & \multicolumn{4}{c}{One-Class SVM} & \multicolumn{4}{c}{Conformal P-value based approach} \\
{} &            AUROC &    $\Delta$AUROC &            FPR95 &    $\Delta$FPR95 &                            AUROC &    $\Delta$AUROC &            FPR95 &    $\Delta$FPR95 \\
\midrule
\{A, B, C, D\} &  \textbf{0.9536} &  \textbf{0.0000} &           0.1483 &           0.0203 &                           0.9160 &          -0.0141 &           0.2385 &           0.0430 \\
\{B, C, D\}    &           0.8647 &          -0.0889 &           0.7688 &           0.6408 &                           0.8381 &          -0.0921 &           1.0000 &           0.8045 \\
\{A, C, D\}    &           0.9493 &          -0.0043 &  \textbf{0.1280} &  \textbf{0.0000} &                           0.8873 &          -0.0428 &           0.2230 &           0.0275 \\
\{A, B, D\}    &           0.9459 &          -0.0077 &           0.2003 &           0.0723 &                           0.9161 &          -0.0141 &           0.2385 &           0.0430 \\
\{A, B, C\}    &           0.9413 &          -0.0123 &           0.1731 &           0.0451 &                  \textbf{0.9301} &  \textbf{0.0000} &           0.2148 &           0.0193 \\
\{C, D\}       &           0.5222 &          -0.4314 &           0.9458 &           0.8179 &                           0.5000 &          -0.4301 &           0.9140 &           0.7185 \\
\bottomrule
\end{tabular}

%% file: supplemental/tables/ood/table_across_layer_fusion_on_hdbscan_high-speed.tex
\begin{tabular}{lllllllll}
\toprule
\multirow{2}{*}{Layers fused} & \multicolumn{4}{c}{One-Class SVM} & \multicolumn{4}{c}{Conformal P-value based approach} \\
{} &            AUROC &    $\Delta$AUROC &            FPR95 &    $\Delta$FPR95 &                            AUROC &    $\Delta$AUROC &            FPR95 &    $\Delta$FPR95 \\
\midrule
\{A, B, C, D\} &           0.7198 &          -0.0121 &           0.9051 &           0.0015 &                           0.7009 &          -0.0088 &           1.0000 &           0.0838 \\
\{B, C, D\}    &  \textbf{0.7319} &  \textbf{0.0000} &  \textbf{0.9037} &  \textbf{0.0000} &                           0.7091 &          -0.0006 &           1.0000 &           0.0838 \\
\{A, C, D\}    &           0.6851 &          -0.0469 &           0.9175 &           0.0139 &                           0.6559 &          -0.0538 &           1.0000 &           0.0838 \\
\{A, B, D\}    &           0.6986 &          -0.0334 &           0.9096 &           0.0059 &                           0.7035 &          -0.0062 &           1.0000 &           0.0838 \\
\{A, B, C\}    &           0.7022 &          -0.0298 &           0.9073 &           0.0036 &                           0.6923 &          -0.0173 &           1.0000 &           0.0838 \\
\{C, D\}       &           0.5291 &          -0.2028 &           0.9448 &           0.0411 &                           0.4970 &          -0.2126 &  \textbf{0.9162} &  \textbf{0.0000} \\
\bottomrule
\end{tabular}

%% file: supplemental/tables/ood/table_across_layer_fusion_on_hdbscan_average_over_scenarios.tex
\begin{tabular}{lllllllll}
\toprule
\multirow{2}{*}{Layers fused} & \multicolumn{4}{c}{One-Class SVM} & \multicolumn{4}{c}{Conformal P-value based approach} \\
{} &            AUROC &    $\Delta$AUROC &            FPR95 &    $\Delta$FPR95 &                            AUROC &    $\Delta$AUROC &            FPR95 &    $\Delta$FPR95 \\
\midrule
\{A, B, C, D\} &  \textbf{0.8367} &  \textbf{0.0000} &           0.5267 &           0.0039 &                           0.8085 &          -0.0028 &           0.6193 &           0.0215 \\
\{B, C, D\}    &           0.7983 &          -0.0384 &           0.8362 &           0.3135 &                           0.7736 &          -0.0376 &           1.0000 &           0.4023 \\
\{A, C, D\}    &           0.8172 &          -0.0195 &  \textbf{0.5228} &  \textbf{0.0000} &                           0.7716 &          -0.0396 &           0.6115 &           0.0137 \\
\{A, B, D\}    &           0.8222 &          -0.0145 &           0.5549 &           0.0322 &                           0.8098 &          -0.0015 &           0.6193 &           0.0215 \\
\{A, B, C\}    &           0.8217 &          -0.0150 &           0.5402 &           0.0174 &                  \textbf{0.8112} &  \textbf{0.0000} &           0.6074 &           0.0097 \\
\{C, D\}       &           0.5257 &          -0.3111 &           0.9453 &           0.4226 &                           0.4985 &          -0.3127 &           0.9151 &           0.3173 \\
\bottomrule
\end{tabular}

%% file: supplemental/tables/ood/table_across_layer_fusion_on_knn_high-delay.tex
\begin{tabular}{lllllllll}
\toprule
\multirow{2}{*}{Layers fused} & \multicolumn{4}{c}{One-Class SVM} & \multicolumn{4}{c}{Conformal P-value based approach} \\
{} &            AUROC &    $\Delta$AUROC &            FPR95 &    $\Delta$FPR95 &                            AUROC &    $\Delta$AUROC &            FPR95 &    $\Delta$FPR95 \\
\midrule
\{A, B, C, D\} &           0.9926 &          -0.0023 &           0.0312 &           0.0123 &                           0.9924 &          -0.0024 &           0.0354 &           0.0116 \\
\{B, C, D\}    &           0.9399 &          -0.0549 &           0.4035 &           0.3846 &                           0.9460 &          -0.0488 &           0.3417 &           0.3179 \\
\{A, C, D\}    &           0.9949 &          -0.0000 &  \textbf{0.0189} &  \textbf{0.0000} &                           0.9944 &          -0.0005 &           0.0263 &           0.0025 \\
\{A, B, D\}    &           0.9902 &          -0.0047 &           0.0375 &           0.0186 &                           0.9893 &          -0.0056 &           0.0365 &           0.0127 \\
\{A, B, C\}    &           0.9921 &          -0.0027 &           0.0306 &           0.0117 &                           0.9929 &          -0.0020 &           0.0327 &           0.0089 \\
\{C, D\}       &           0.5737 &          -0.4212 &           0.9759 &           0.9570 &                           0.6304 &          -0.3645 &           0.9067 &           0.8829 \\
\bottomrule
\end{tabular}

%% file: supplemental/tables/ood/table_across_layer_fusion_on_knn_high-speed.tex
\begin{tabular}{lllllllll}
\toprule
\multirow{2}{*}{Layers fused} & \multicolumn{4}{c}{One-Class SVM} & \multicolumn{4}{c}{Conformal P-value based approach} \\
{} &            AUROC &    $\Delta$AUROC &            FPR95 &    $\Delta$FPR95 &                            AUROC &    $\Delta$AUROC &            FPR95 &    $\Delta$FPR95 \\
\midrule
\{A, B, C, D\} &  \textbf{0.8054} &  \textbf{0.0000} &           0.7762 &           0.0020 &                           0.7881 &          -0.0094 &           0.8212 &           0.0541 \\
\{B, C, D\}    &           0.7791 &          -0.0262 &           0.9074 &           0.1332 &                           0.7792 &          -0.0182 &           0.8349 &           0.0678 \\
\{A, C, D\}    &           0.7512 &          -0.0542 &           0.8319 &           0.0577 &                           0.7285 &          -0.0690 &           0.9388 &           0.1717 \\
\{A, B, D\}    &           0.8005 &          -0.0048 &           0.8032 &           0.0291 &                           0.7717 &          -0.0258 &           0.8807 &           0.1136 \\
\{A, B, C\}    &           0.7937 &          -0.0117 &  \textbf{0.7742} &  \textbf{0.0000} &                  \textbf{0.7975} &  \textbf{0.0000} &  \textbf{0.7671} &  \textbf{0.0000} \\
\{C, D\}       &           0.6091 &          -0.1962 &           0.9577 &           0.1835 &                           0.6132 &          -0.1842 &           0.9440 &           0.1769 \\
\bottomrule
\end{tabular}

%% file: supplemental/tables/ood/table_across_layer_fusion_on_knn_average_over_scenarios.tex
\begin{tabular}{lllllllll}
\toprule
\multirow{2}{*}{Layers fused} & \multicolumn{4}{c}{One-Class SVM} & \multicolumn{4}{c}{Conformal P-value based approach} \\
{} &            AUROC &    $\Delta$AUROC &            FPR95 &    $\Delta$FPR95 &                            AUROC &    $\Delta$AUROC &            FPR95 &    $\Delta$FPR95 \\
\midrule
\{A, B, C, D\} &  \textbf{0.8990} &  \textbf{0.0000} &           0.4037 &           0.0013 &                           0.8903 &          -0.0049 &           0.4283 &           0.0284 \\
\{B, C, D\}    &           0.8595 &          -0.0394 &           0.6555 &           0.2531 &                           0.8626 &          -0.0325 &           0.5883 &           0.1884 \\
\{A, C, D\}    &           0.8730 &          -0.0259 &           0.4254 &           0.0230 &                           0.8614 &          -0.0338 &           0.4825 &           0.0827 \\
\{A, B, D\}    &           0.8954 &          -0.0036 &           0.4204 &           0.0180 &                           0.8805 &          -0.0147 &           0.4586 &           0.0587 \\
\{A, B, C\}    &           0.8929 &          -0.0061 &  \textbf{0.4024} &  \textbf{0.0000} &                  \textbf{0.8952} &  \textbf{0.0000} &  \textbf{0.3999} &  \textbf{0.0000} \\
\{C, D\}       &           0.5914 &          -0.3075 &           0.9668 &           0.5644 &                           0.6218 &          -0.2734 &           0.9253 &           0.5255 \\
\bottomrule
\end{tabular}

%% file: supplemental/tables/ood/table_across_layer_fusion_on_ocsvm_activations_high-delay.tex
\begin{tabular}{lllllllll}
\toprule
\multirow{2}{*}{Layers fused} & \multicolumn{4}{c}{One-Class SVM} & \multicolumn{4}{c}{Conformal P-value based approach} \\
{} &            AUROC &    $\Delta$AUROC &            FPR95 &    $\Delta$FPR95 &                            AUROC &    $\Delta$AUROC &            FPR95 &    $\Delta$FPR95 \\
\midrule
\{A, B, C, D\} &           0.8936 &          -0.0428 &           0.3077 &           0.0855 &                           0.8791 &          -0.0402 &           0.3957 &           0.1114 \\
\{B, C, D\}    &           0.7000 &          -0.2364 &           0.9575 &           0.7352 &                           0.6842 &          -0.2351 &           0.8673 &           0.5830 \\
\{A, C, D\}    &           0.9036 &          -0.0328 &           0.2790 &           0.0567 &                           0.8954 &          -0.0239 &           0.3425 &           0.0582 \\
\{A, B, D\}    &           0.9198 &          -0.0166 &           0.2698 &           0.0475 &                           0.8997 &          -0.0195 &           0.3380 &           0.0537 \\
\{A, B, C\}    &           0.8910 &          -0.0453 &           0.3035 &           0.0813 &                           0.8850 &          -0.0342 &           0.3716 &           0.0873 \\
\{C, D\}       &           0.6209 &          -0.3154 &           0.9186 &           0.6963 &                           0.6277 &          -0.2915 &           0.9459 &           0.6616 \\
\bottomrule
\end{tabular}

%% file: supplemental/tables/ood/table_across_layer_fusion_on_ocsvm_activations_high-speed.tex
%

%% file: supplemental/tables/ood/table_across_layer_fusion_on_ocsvm_activations_average_over_scenarios.tex
%

%% file: supplemental/tables/ood/table_across_layer_fusion_on_mahalanobis_activations_high-delay.tex
%

%% file: supplemental/tables/ood/table_across_layer_fusion_on_mahalanobis_activations_high-speed.tex
%

%% file: supplemental/tables/ood/table_across_layer_fusion_on_mahalanobis_activations_average_over_scenarios.tex
%

%% file: supplemental/tables/ood/table_across_layer_fusion_on_within_layer_fusions_high-delay.tex
%

%% file: supplemental/tables/ood/table_across_layer_fusion_on_within_layer_fusions_high-speed.tex
%

%% file: supplemental/tables/ood/table_across_layer_fusion_on_within_layer_fusions_average_over_scenarios.tex
%